\documentclass[journal]{IEEEtran}
\usepackage[citecolor=blue, colorlinks]{hyperref}
\usepackage{graphicx}
\usepackage{float}
\usepackage{subfigure}
\usepackage{epsfig}
\usepackage{epstopdf}
\usepackage{indentfirst}
\usepackage{amsfonts}
\usepackage{amsmath,amssymb}
\usepackage{algorithm}
\usepackage{algorithmic}
\usepackage{verbatim}
\usepackage{multirow}
\usepackage{array}
\usepackage{color}
\usepackage{cite}
\usepackage{makecell}
\usepackage{balance}
\usepackage{hyperref}
\usepackage[table]{xcolor}
\usepackage{fancyhdr}
\usepackage{hyperref}
\usepackage{soul}
\usepackage[normalem]{ulem}
\usepackage{fancyhdr}

\usepackage{bbding}
\usepackage{amssymb}
\usepackage{pifont}
\usepackage{soul}
\usepackage[figuresright]{rotating}

\ifCLASSINFOpdf
\else
\fi

\begin{document}

\title{\textcolor{black}{EGRC-Net: Embedding-induced Graph Refinement Clustering Network}}

\author{
    Zhihao~Peng,
    Hui~Liu,
	Yuheng~Jia, ~\IEEEmembership{Member,~IEEE}, 
	Junhui~Hou, ~\IEEEmembership{Senior Member,~IEEE}

\thanks{This work was supported by the Hong
Kong University Grants Committee through the Institutional Development
Scheme Research Infrastructure under Grant UGC/IDS11/19. Corresponding author: \textit{Hui Liu}.}
\thanks{Z. Peng and J. Hou are with the Department of Computer Science, City University of Hong Kong, Kowloon, Hong Kong 999077 (E-mail: zhihapeng3-c@my.cityu.edu.hk; jh.hou@cityu.edu.hk).}
\thanks{H. Liu is with the School of Computing \& Information Sciences, Caritas Institute of Higher Education, Hong Kong (E-mail: hliu99-c@my.cityu.edu.hk).}
\thanks{Y. Jia is with the School of Computer Science and Engineering, Southeast University, Nanjing 210096, China, and also with Key Laboratory of Computer Network and Information Integration (Southeast University), Ministry of Education, China (E-mail: yhjia@seu.edu.cn).}
}

\markboth{
}{\MakeLowercase{\textit{et al.}}}

\maketitle

\begin{abstract}
Existing graph clustering networks heavily rely on a predefined yet fixed graph, which can lead to failures when the initial graph fails to accurately capture the data topology structure of the embedding space. In order to address this issue, we propose a novel clustering network called Embedding-Induced Graph Refinement Clustering Network (EGRC-Net), which effectively utilizes the learned embedding to adaptively refine the initial graph and enhance the clustering performance. To begin, we leverage both semantic and topological information by employing a vanilla auto-encoder and a graph convolution network, respectively, to learn a latent feature representation. Subsequently, we utilize the local geometric structure within the feature embedding space to construct an adjacency matrix for the graph. This adjacency matrix is dynamically fused with the initial one using our proposed fusion architecture. To train the network in an unsupervised manner, we minimize the Jeffreys divergence between multiple derived distributions. Additionally, we introduce an improved approximate personalized propagation of neural predictions to replace the standard graph convolution network, enabling EGRC-Net to scale effectively. Through extensive experiments conducted on nine widely-used benchmark datasets, we demonstrate that our proposed methods consistently outperform several state-of-the-art approaches. Notably, EGRC-Net achieves an improvement of more than 11.99\% in Adjusted Rand Index (ARI) over the best baseline on the DBLP dataset. Furthermore, our scalable approach exhibits a 10.73\% gain in ARI while reducing memory usage by 33.73\% and decreasing running time by 19.71\%. The code for EGRC-Net will be made publicly available at \url{https://github.com/ZhihaoPENG-CityU/EGRC-Net}.

\end{abstract}


\IEEEpeerreviewmaketitle

\section{Introduction}
\textcolor{black}{
Clustering is a typical yet challenging machine learning topic with a series of real-world applications, including object detection \cite{zhan2018multiview,li2020deep,peng2022adaptive,jia2023semi,xia2023graph}, social network analysis \cite{ghosh2021learning,wang2021decorrelated,gong2022attributed,zhang2022attributed}, and face recognition \cite{miklautz2020deep,peng2021maximum,jia2021clustering,yu2022sail}. Clustering aims to partition data into different groups based on their intrinsic patterns, such that similar samples are clustered together while dissimilar samples are separated from each other. Due to the powerful embedding learning capability of the convolutional neural network, deep embedding clustering has been widely studied in recent decades. For example, Hinton \textit{et al}. \cite{hinton2006reducing} proposed a deep auto-encoder network (DAE) to focus on the node attribute information for clustering assignments. Xie \textit{et al.} \cite{xie2016unsupervised} jointly learned feature representations and clustering assignments to present the deep embedded clustering network (DEC). Guo \textit{et al.} \cite{guo2017improved} improved DEC via a reconstruction loss.}

\begin{figure}[!t]
	\includegraphics [width=0.98\columnwidth]{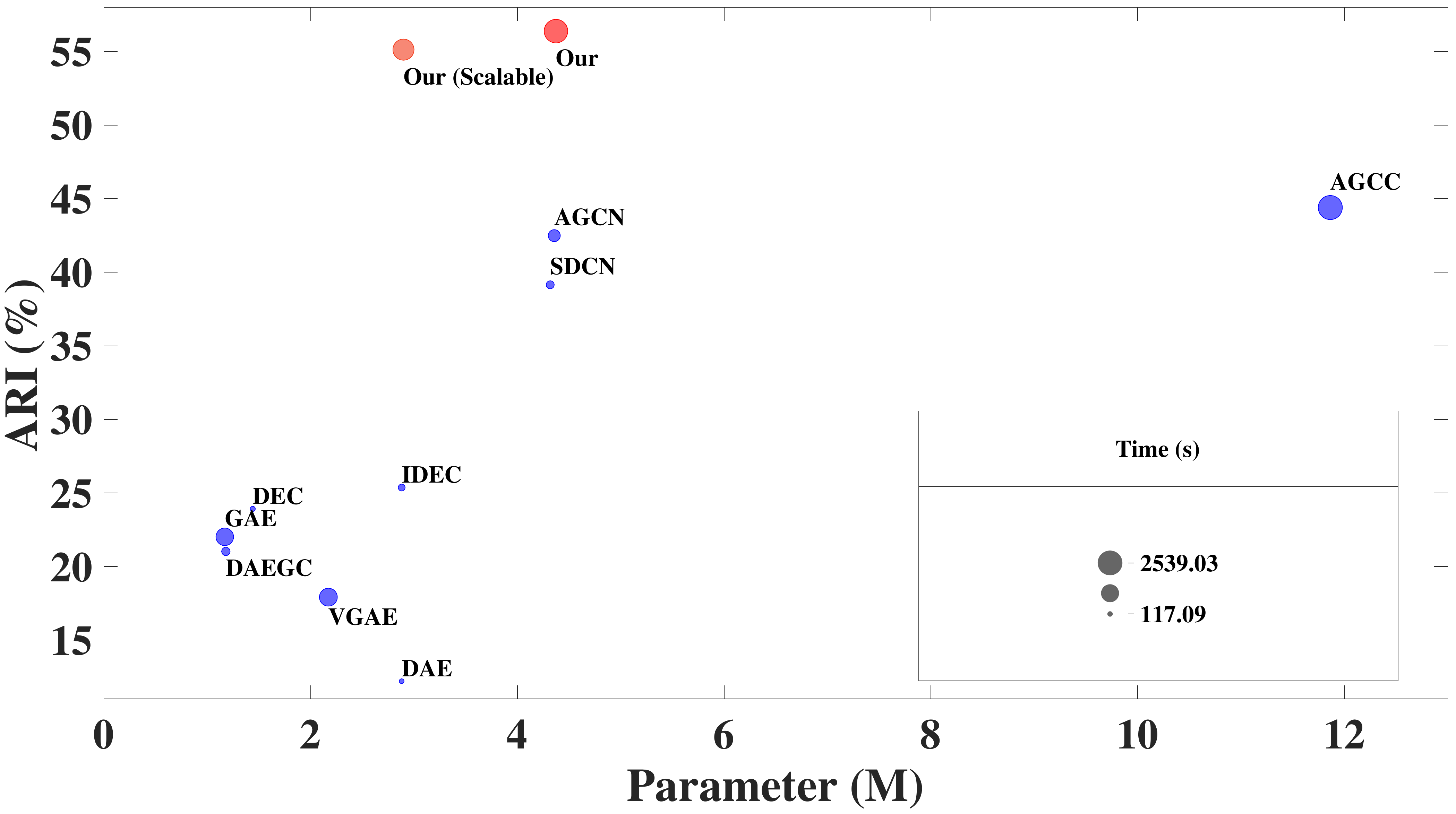}
    \caption{Comparisons of the adjusted rand index (ARI), parameter numbers (in millions), and running time (in seconds) of different methods on DBLP, where we trained the proposed methods, three DAE-based clustering methods (DAE \cite{hinton2006reducing}, DEC \cite{xie2016unsupervised}, and IDEC \cite{guo2017improved}), three GCN-based clustering methods (GAE \cite{kipf2016variational}, VGAE \cite{kipf2016variational}, and DAEGC\cite{wang2019attributed}), and three DAE and GCN combination-based clustering methods (SDCN \cite{bo2020structural}, AGCN \cite{peng2021attention}, and AGCC \cite{he2022parallelly}) with 200 epochs and repeated experiments ten times. The diameter of the bubble is proportional to the running time, and we make it easier to see all the bubbles by changing the range of diameters to be between 4 and 20 points.}
	\label{fig: metric_para_time}
\end{figure}

\textcolor{black}{Although these DAE-based networks have shown impressive embedding learning capability, they neglect the topology structure information of the input, limiting the clustering performance. To this end, the graph convolution network (GCN) is utilized to conduct the deep embedding clustering by propagating spatial relations with the representations of its neighbors, namely the graph clustering network. For example, Kipf and Welling \cite{kipf2016variational} proposed the graph auto-encoder to learn the graph structure feature. Pan \textit{et al.} \cite{pan2018adversarially} injected an adversarial regularizer into the GCN framework. Zhang \textit{et al.} \cite{zhang2022embedding} used a relaxed K-means \cite{macqueen1967some} to improve the representation learning capability of the GCN-based clustering method. Recently, numerous works have combined DAE and GCN to simultaneously consider node attribute and topology structure information, achieving significant improvement in clustering performance. For instance, Bo \textit{et al.} \cite{bo2020structural} integrated the DAE and GCN features via the structural deep clustering network. Peng \textit{et al.} \cite{peng2021attention} employed two feature fusion modules to merge the DAE and GCN features. 
He \textit{et al.} \cite{he2022parallelly} conducted the feature fusion and graph reconstruction layer by layer. 
\textit{Nevertheless, all these GCN-based graph clustering networks adopt a predefined graph as a fixed input and may fail to correctly distinguish the samples if the initial graph cannot truly and precisely reflect their topology structures on the embedding space, \textcolor{black}{which is referred to as the fixed graph issue}.} Thus, it is expected that the clustering performance can be boosted by performing graph refinement based on the guidance from the learned embedding representation.}

In this paper, we propose a novel embedding-induced graph refinement clustering network (\textbf{EGRC-Net}) framework to adaptively use the learned embedding for evolving the initial graph, aiming to boost the embedding learning capability of the graph clustering network. Specifically, we first use the vanilla DAE and GCN modules to conduct embedding learning. Subsequently, we explore the local geometric structure information on the embedding space to construct an adjacency matrix. Then, we utilize multilayer perceptron layers and a series of normalization terms to dynamically fuse the constructed graph with the initial one, seeking to preserve the intrinsic semantic structure information. Finally, we minimize the Jeffreys divergence of multiple derived distributions in an unsupervised manner to jointly conduct embedding learning and graph refinement. 

Furthermore, we advance EGRC-Net in a scalable manner to handle the practical large-graph problem \cite{ying2018graph,zhang2022embedding,park2022cgc}. Specifically, the weak scalability of EGRC-Net ascribes to the utilization of GCN, since the explicit message-passing of GCN leads to an expensive neighborhood expansion \cite{bojchevski2020scaling}. Thus, we utilize the approximate personalized propagation of neural predictions (APPNP) \cite{gasteigerpredict} instead of multi-layer graph convolution to complete the transmission and aggregation of node information on the graph. However, it is still required to predefine a heuristic threshold that determines the teleport probability. Besides, the predefined value also limits its availability since, in real-world clustering tasks, there has little guideline to set a suitable threshold. Thus, we empirically initialize the threshold using a random generator to sample from a uniform distribution on the interval $[0,1)$ and learn that variable during the network training. To this end, we can achieve information propagation by using the personalized PageRank with a learned threshold value without relying on a predefined heuristic value, namely improved APPNP (IAPPNP). In this way, we obtain the \textbf{scalable EGRC-Net}.

\textcolor{black}{
In summary, the contributions of our work are as follows:
\begin{itemize}
    \item We handle the fixed graph issue by dynamically fusing an embedding-induced graph with the initial one, aiming to preserve the intrinsic semantic structure information. 
    \item We design a unified learning framework by minimizing the Jeffreys divergence of multiple derived distributions to jointly conduct embedding learning and graph refinement, mutually benefiting both components. 
    \item We investigate the scalability of the proposed method by utilizing an improved approximate personalized propagation of neural predictions instead of multi-layer graph convolution to complete the transmission and aggregation of node information on the graph.
    \item Extensive experiments on nine commonly used benchmark datasets demonstrate that the proposed methods consistently outperform several state-of-the-art approaches. In particular, as shown in Figure \ref{fig: metric_para_time}, EGRC-Net improves the ARI by more than 11.99\% over the best baseline on DBLP, and the scalable one has a 10.73\% gain in ARI while saving 33.73\% in memory and 19.71\% in running time.
\end{itemize}
}

{The rest of this paper is organized as follows. We briefly review the related works in Section \ref{sec: rw} and introduce the proposed network in Section \ref{sec: pm}. Afterward, we give the experimental results and analyses in Section \ref{sec: eprm} and conclude this paper in Section \ref{sec: con}.}

\textcolor{black}{Throughout this paper, matrices are denoted by bold upper case letters, vectors by bold lower case letters, and scalars by italic lower case letters, respectively. Given a matrix $\mathbf{B}\in\mathbb{R}^{m\times n}$, $ \left\|\mathbf{B}\right\|_F$ denotes the Frobenius norm of $\mathbf{B}$, i.e., $\left\|\mathbf{B}\right\|_F=\sqrt{\sum_{i=1}^{m}\sum_{j=1}^{n}{\left| b_{i,j} \right|}^2}$. \textcolor{black}{Let $\mathbf{X}\in\mathbb{R}^{n\times d}$ be the raw data, $\mathbf{A}\in\mathbb{R}^{n\times n}$ the adjacency matrix of the graph, we summarize the main notations in Table \ref{tab: notation}.}}

\begin{table}[htb!]
\caption{Main notations and descriptions.}
\label{tab: notation}
\centering
\resizebox{0.88\columnwidth}{!}{
\begin{tabular}{cl|l}
    \hline\hline
    \multicolumn{2}{c|}{Notations} & Descriptions \\ 
    \hline\hline
    $\mathbf{X}$&$\in\mathbb{R}^{\emph{n}\times \emph{d}}$                        & The raw matrix
    \\
    $\hat{\mathbf{X}}$&$\in\mathbb{R}^{\emph{n}\times \emph{d}}$                  & The reconstructed matrix
    \\
    $\mathbf{A}$&$\in\mathbb{R}^{\emph{n}\times \emph{n}}$                        & The adjacency matrix of the predefined graph
    \\
    $\mathbf{D}$&$\in\mathbb{R}^{\emph{n}\times \emph{n}}$                        & The degree matrix
    \\
    $\mathbf{Z}_{a}$&$\in\mathbb{R}^{\emph{n}\times \kappa}$                      & The learned embedding
    \\ 
    $\mathbf{A}_{z}$&$\in\mathbb{R}^{\emph{n}\times \emph{n}}$                    & The constructed adjacency matrix from $\mathbf{Z}_{a}$        
    \\
    $\mathbf{A}_{f}$&$\in\mathbb{R}^{\emph{n}\times \emph{n}}$                    & The fused adjacency matrix
    \\
    $\mathbf{H}$&$\in\mathbb{R}^{\emph{n}\times {\emph{d}_\emph{l}}}$             & The feature extracted from DAE
    \\
    $\mathbf{Q}$&$\in\mathbb{R}^{\emph{n}\times \kappa}$                          & The distribution obtained from $\mathbf{H}$
    \\
    $\mathbf{P}$&$\in\mathbb{R}^{\emph{n}\times \kappa}$                          & The distribution obtained from $\mathbf{Z}_{a}$
    \\
    \hline
    $\emph{n}$ &                         & The number of samples                                 \\
    $\kappa$   &                         & The number of clusters                                \\
    $\emph{l}$ &                         & The number of encoder/decoder layers                 \\
    $\emph{d}$ &                         & The dimension of $\mathbf{X}$                         \\
    $\emph{d}_\emph{l}$ &                & The dimension of $\mathbf{H}$                         \\
    $\rho$     &                         & The predefined threshold \\
    $\Theta$   &                         & The learned threshold \\
    $\tau$     &                         & The number of power iteration steps \\
    \hline
    $\cdot \| \cdot $  &                 & The concatenation operation                           \\
    $\left\|\cdot\right\|_F$  &          & The Frobenius norm                                    \\
    \hline\hline  
\end{tabular}
}
\end{table}

\section{Related Work}\label{sec: rw} 

\subsection{DAE-based Clustering Methods}
Due to the powerful expression ability of the convolutional neural network (CNN), many deep embedding clustering methods have been proposed and achieved impressive performance in practical applications \cite{xie2016unsupervised,guo2017improved,miklautz2020deep,naumov2021objective,lin2022mixture,niu2022spice}. \textcolor{black}{For example, Hinton \textit{et al}. \cite{hinton2006reducing} adopted a series of CNNs and a reconstruction loss to build a vanilla DAE, which can be formulated as
\begin{equation}
\begin{aligned}
&\min_{\mathbf{H}_{i}} \left\| \mathbf{X} - \hat{\mathbf{X}} \right\|^2_F \quad \rm{s.t.} \quad \mathbf{H}_{0}=\mathbf{X}, \quad \hat{\mathbf{H}}_\emph{l}=\hat{\mathbf{X}},
\label{eq: DAE}
\end{aligned}
\end{equation}
where $\hat{\mathbf{X}}\in\mathbb{R}^{n\times d}$, $\mathbf{H}_{i} = \phi ( \mathbf{W}_{i}^{e}\mathbf{H}_{i-1}+ \mathbf{b}_{i}^{e})\in\mathbb{R}^{{\it n} \times {{\it d}_i}}$, and $\hat{\mathbf{H}}_{i} = \phi ( \mathbf{W}_{i}^{d}\hat{\mathbf{H}}_{i-1}+ \mathbf{b}_{i}^{d})\in\mathbb{R}^{{\it n} \times {\hat{{\it d}}_i}}$ 
indicate the reconstructed data, the $i$-{th} encoder and decoder outputs, respectively. $\mathbf{W}_{i}^{e}$ and $\mathbf{b}_{i}^{e}$, $\mathbf{W}_{i}^{d}$ and $\mathbf{b}_{i}^{d}$, $\phi ( \cdot )$, and $\left\|\cdot\right\|_F$ indicate the network weight and bias of the $i$-{th} encoder layer, those of the $i$-{th} decoder layer, the ReLU activation function, and the Frobenius norm, respectively. }
Xie \textit{et al}. \cite{xie2016unsupervised} designed DEC to improve the DAE framework by learning the embedding and clustering assignments in a joint optimization fashion. Guo \textit{et al}. \cite{guo2017improved} introduced a reconstruction loss into the DAE framework to encourage local structure preservation, namely improved deep embedded clustering (IDEC). However, DAE is designed to deal with the grid-wise features of Euclidean space, failing to handle the complex topological data, i.e., non-Euclidean data, in real-world cases.

\subsection{GCN-based Clustering Methods}
\textcolor{black}{
GCN integrates the vertex and its neighbors to determine a graph convolution in a spatial-based feature fusion manner, i.e.,
\begin{equation}
\begin{aligned}
& \mathbf{Z}_{i} = \rm{LReLU}(\mathbf{D}^{-\frac{1}{2}}(\mathbf{A}+\mathbf{I})\mathbf{D}^{-\frac{1}{2}}\mathbf{Z}_{i-1}^{'}\mathbf{W}_i),
\label{eq: GCN}
\end{aligned}
\end{equation}
where $\mathbf{Z}_{i-1}^{'}\in\mathbb{R}^{n\times d_{i-1}}$, $\mathbf{Z}_{i}\in\mathbb{R}^{n\times d_{i}}$, $\rm{LReLU}$,  $\mathbf{D}\in\mathbb{R}^{n\times n}$, and $\mathbf{W}_i$ indicate the $i$-{th} GCN layer input, the $i$-{th} GCN layer output, the Leaky ReLU activation function \cite{maas2013rectifier}, the degree matrix, and the learnable weight matrix, respectively. $\mathbf{Z}_{0}^{'} = \mathbf{X}$. }In the recent decade, GCN has rapidly developed in deep graph clustering, crediting from its efficiency, simplicity, and generality \cite{kipf2016variational,pan2021multi,xu2021multi,zhang2022non,chen2022bag,zhang2022graph,wang2023correntropy,zhang2023towards}. For instance, Kipf and Welling \cite{kipf2016variational} built the graph auto-encoder (GAE) and variational graph auto-encoder (VGAE) based on the variational architecture to learn the graph structure feature via a reconstruction constraint for graph structure preservation. Pan \textit{et al}. \cite{pan2018adversarially} presented the adversarially regularized graph auto-encoder (ARGA) to enhance the learning ability of the network by injecting an adversarial regularizer into the GAE framework. Wang \textit{et al}. \cite{wang2019attributed} improved GAE by introducing the graph attention network model \cite{velivckovic2018graph} into its network architecture to encode the topological structure and node contents, called deep attentional embedded graph clustering (DAEGC). 
\textcolor{black}{
Salha \textit{et al}. \cite{salha2021fastgae} designed a general framework (FastGAE) to scale graph AE and VAE to large graphs.
Mrabah \textit{et al}. \cite{mrabah2022rethinking} presented a series of rethinking graph auto-encoder models (e.g., R-GMM-VGAE (RG-VGAE)) to consider the feature randomness and feature drift.
Salha \textit{et al}. \cite{salha2022modularity} designed modularity-aware graph autoencoders (MA-GAE) to consider both the initial graph structure and modularity-based prior communities.
}
Zhang \textit{et al}. \cite{zhang2022embedding} developed an embedding graph auto-encoder (EGAE) to use a relaxed K-means \cite{macqueen1967some} to alternatively update GAE by gradient descent and perform clustering on inner-product distance space.

\begin{figure}[!t]
    \centering
    \includegraphics [width=0.98\columnwidth]{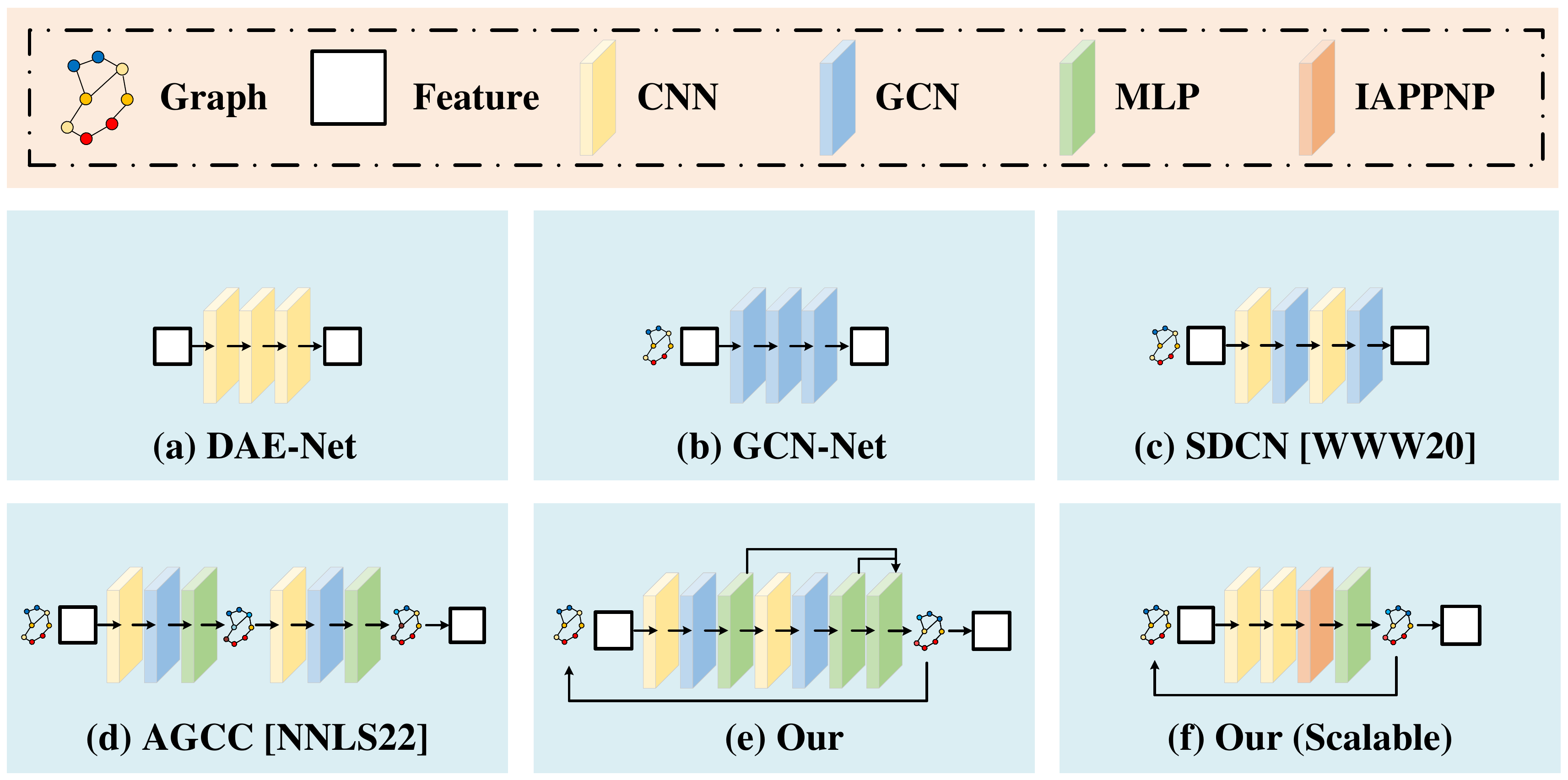}
    \caption{Illustration of different network architectures for deep embedding clustering. (a) DAE-based clustering methods, (b) GCN-based clustering methods, (c) SDCN \cite{bo2020structural}, (d) AGCC \cite{he2022parallelly}, (e) Our EGRC-Net and (e) Our scalable EGRC-Net. Colored cubes represent different network modules, where our proposed methods conduct the network training in an unsupervised manner to jointly conduct embedding learning and graph refinement, mutually benefiting both components.}
    \label{fig: Diff_ours}
\end{figure}

\subsection{DAE and GCN Combination-based Methods}
Recently, numerous works have combined DAE and GCN to achieve clustering \cite{bo2020structural,peng2021attention,he2022parallelly,GNNBook2022,peng2022deep}. For example, Bo \textit{et al}. \cite{bo2020structural} integrated the node attribute and topology structure information based on the DEC framework to design the structural deep clustering network (SDCN). Peng \textit{et al}. \cite{peng2021attention} utilized two attention-based feature fusion modules to merge the DAE and GCN features, namely attention-driven graph clustering network (AGCN).
He \textit{et al}. \cite{he2022parallelly} conducted the DAE and GCN feature fusion and the graph construction in each layer to improve the representation ability. 
Although these approaches have obtained remarkable improvements, they heavily rely on a predefined graph and may fail if the fixed graph cannot well reflect the intrinsic semantic structures on the embedding space. To this end, we propose a novel graph clustering network framework EGRC-Net that adaptively uses the learned embedding to evolve the initial graph.

\subsection{Approximate Personalized Propagation of Neural Predictions (APPNPs)} 
\textcolor{black}{Personalized propagation of neural predictions (PPNP) is constructed by using the propagation scheme based on the relationship between GCN and personalized PageRank \cite{bojchevski2020scaling,gasteigerpredict,spinelli2020adaptive,hou2021automated,choi2022personalized}. \textcolor{black}{Let $\mathbf{E}_{0}=\emph{f}(\mathbf{X}\mathbf{W}_0)$ be the mapped embedding with the learned weight $\mathbf{W}_0$, $\mathbf{I}\in\mathbb{R}^{n\times n}$ the identity matrix, $\rho$ the predefined heuristic threshold, then the output of PPNP is $\mathbf{E} = \emph{Softmax}\left(\rho\left(\mathbf{I}-\left(1-\rho\right)\mathbf{A}\right)^{-1}\mathbf{E}_{0}\right)$.} In practice, to avoid the matrix inversion, Gasteiger \textit{et al}. \cite{gasteigerpredict} designed APPNP to approximate PPNP in a power iteration manner, i.e., 
\begin{equation}
\begin{aligned}
& \mathbf{E}_{\tau} = \emph{Softmax}\left( \left(1-\rho\right)\mathbf{A}\mathbf{E}_{\tau-1}+\rho \mathbf{E}_{0}\right),
\label{eq: APPNP}
\end{aligned}
\end{equation}
where $\tau$ is the number of power iteration steps. \textcolor{black}{Notably, for GCN, the $\emph{l}$-{th} feature $\mathbf{Z}_{l}$ could be formulated as $(...\rm{LReLU}(\mathbf{D}^{-\frac{1}{2}}(\mathbf{A}+\mathbf{I})\mathbf{D}^{-\frac{1}{2}}\mathbf{X}\mathbf{W}_0)...\mathbf{W}_
{\emph{l}-1})$, where network parameters increase as the number of network layers increases. Differently, APPNP separates the mapping and propagation phases and utilizes the personalized PageRank to conduct messaging propagation, resulting in fewer parameters (only $\mathbf{W}_0$) and less training time.} Although APPNP-based clustering approaches have significantly improved scalability, most of them have to pre-specify a heuristic threshold to determine the teleport probability, and few models focus on adaptively learning a suitable threshold to promote the representative capability in clustering tasks. Thus, we propose a simple yet effective strategy to initialize the threshold and learn it during the network training, namely IAPPNP. In this way, we utilize the IAPPNP modules to develop the scalable EGRC-Net. \textcolor{black}{The framework comparison of typical deep embedding clustering approaches and our methods is illustrated in Figure \ref{fig: Diff_ours}.}
}

\begin{figure}[!t]
    \centering
    \includegraphics [width=0.6965\columnwidth]{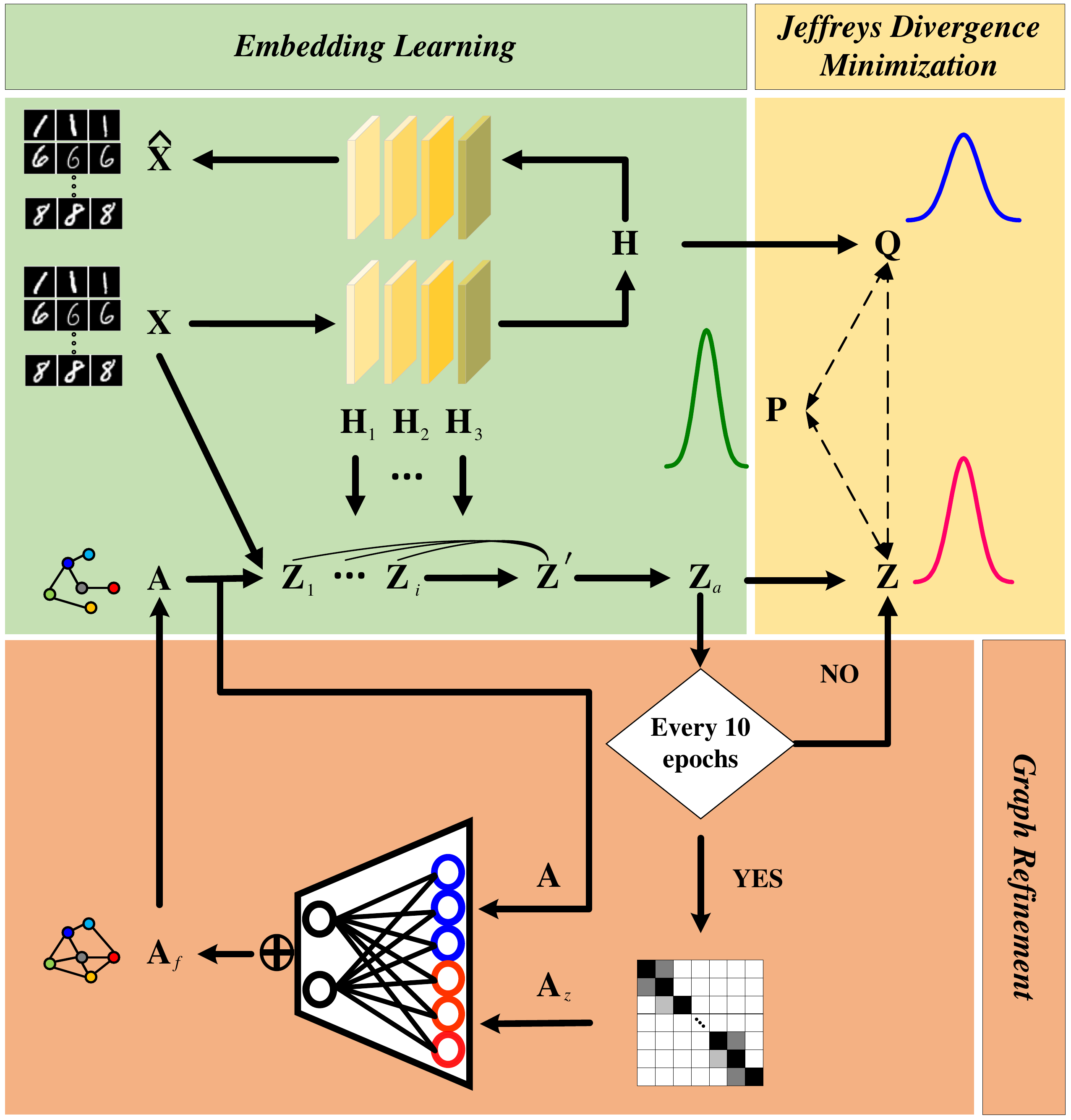}
    \caption{\textcolor{black}{The overall flowchart of the proposed method, namely embedding-induced graph refinement clustering network (EGRC-Net), where our proposed method conducts the network training in an unsupervised manner to jointly conduct embedding learning and graph refinement, mutually benefiting both components.}}
    \label{fig: EGRC-Framework}
\end{figure}

\section{Proposed Methods}\label{sec: pm}  
\textcolor{black}{
In this section, we first present a novel embedding-induced graph refinement clustering network (EGRC-Net) framework, as shown in Figure \ref{fig: EGRC-Framework}. It mainly consists of the embedding learning module (Section \ref{sec: emb_learn}) and the graph refinement architecture (Section \ref{sec: graph_ref}). Then, we employ the Jeffreys divergence minimization among multiple derived distributions to train the network in a unified optimization manner (Section \ref{sec: clu_opt}), capable of dynamically and jointly conducting embedding learning and graph refinement to mutually benefit both components. Moreover, we further designed an improved APPNP to replace the vanilla GCN for embedding learning, developing a simple yet effective scalable variant of EGRC-Net (Section \ref{sec: scalable}). Finally, the computational complexity of EGRC-Net and the scalable one is provided in Section \ref{sec: comp}. }

\subsection{Embedding Learning}\label{sec: emb_learn}
\textcolor{black}{
For embedding learning, we first utilize the vanilla DAE and GCN modules to extract the node attribute information and the topology structure information. Specifically, we use DAE to extract the $i$-{th} DAE feature representation $\mathbf{H}_{i}$ by Eq. (\ref{eq: DAE}) and the $i$-{th} GCN feature representation $\mathbf{Z}_{i}$ by Eq. (\ref{eq: GCN}). Then, we exploit a multilayer perceptron (MLP) layer with a normalization operation to learn the weight vectors of $\mathbf{H}_\emph{i}$ and $\mathbf{Z}_\emph{i}$ for subsequent information fusion, i.e.,
\begin{equation}
\begin{aligned}
[\mathbf{m}_{i,1},\mathbf{m}_{i,2}]=\ell_{2}\left( \emph{Softmax} \left(\rm{LReLU}\left(\left[\mathbf{H}_i\|\mathbf{Z}_{i}\right]\mathbf{W}_{i}^\mathbf{F}\right)\right) \right),
\label{eq: HZ}
\end{aligned} 
\end{equation}
where $\mathbf{m}_{i,1}$ and $\mathbf{m}_{i,2}$ measure the importance of $\mathbf{H}_\emph{i}$ and $\mathbf{Z}_\emph{i}$, $\mathbf{W}_{i}^\mathbf{F}\in\mathbb{R}^{{2d_i}\times 2}$, $\cdot \| \cdot$, and $\ell_2$ indicate a learnable weight matrix, the concatenation operation, and the $\ell_2$ normalization, respectively. 
Afterward, based on the learned weight vectors, we can dynamically merge $\mathbf{H}_{i}$ and $\mathbf{Z}_{i}$, i.e., 
\begin{equation}
\begin{aligned}
& \mathbf{Z}_{i}^{'}=\left(\mathbf{m}_{i,1}\mathbf{1}_i \right) \odot \mathbf{H}_i+\left( \mathbf{m}_{i,2}\mathbf{1}_i \right) \odot \mathbf{Z}_{i},
\label{eq: HZ}
\end{aligned} 
\end{equation}
where $\odot$ and $\mathbf{1}_{i}\in\mathbb{R}^{1\times d_{i}}$ indicate the Hadamard product of matrices and the all ones vector, respectively.
Moreover, we aggregate the outputs of different GCN layers for fully utilizing the off-the-shelf information, where the weighted fusion process can be formulated as 
\begin{equation}
\begin{aligned}
& \mathbf{Z}^{'} =[\left(\mathbf{u}_{1}\mathbf{1}_{1}\right)\odot\mathbf{Z}_1\|\cdots\|
\left(\mathbf{u}_{l+1}\mathbf{1}_{l+1}\right)\odot\mathbf{Z}_{l+1}],
\label{eq: multiscale}
\end{aligned}
\end{equation}
where we utilize the same weight learning strategy with a learnable weight matrix $\mathbf{W}^\mathbf{S}$ to learn the corresponding weights of multi-scale features from different GCN layers. The formulation is as follows, $\mathbf{U}=\ell_{2}(\emph{Softmax}(\rm{LReLU}( [\mathbf{Z}_1\|\cdots\|\mathbf{Z}_{l+1}]\mathbf{W}^\mathbf{S})))$, where $\mathbf{u}_{i}$ is the $\emph{i}$-th element of $\mathbf{U}$ and $l=4$ in the experiments. Notably, we use a weighted concatenation form to conduct feature fusion because the outputs of different GCN layers have different dimensions.} In addition, we use the Laplacian smoothing technique, a learnable weight matrix $\mathbf{W}^\mathbf{Z}$, and the Softmax function to learn the final embedding, i.e., 
\begin{equation}
\begin{aligned}
& \mathbf{Z}_{a}= \emph{Softmax}(\mathbf{D}^{-\frac{1}{2}}(\mathbf{A}+\mathbf{I})\mathbf{D}^{-\frac{1}{2}}\mathbf{Z}^{'}\mathbf{W}^\mathbf{Z}),
\label{eq: Z}
\end{aligned}
\end{equation}
where $\mathbf{Z}_{a}\in \mathbb{R}^{n\times \kappa}$ indicates the learned embedding with $\kappa$ being the cluster number. 

\subsection{Graph Refinement} \label{sec: graph_ref}
\textcolor{black}{
The adjacency matrix $\mathbf{A}$ is crucial in GCN-based graph clustering. However, earlier works heavily relied on the adjacency matrix of a predefined graph and may fail if that fixed graph cannot well reflect the intrinsic semantic structure of the embedding space. Such a phenomenon inspires us to propose a novel graph clustering network that adaptively refines the graph rather than relies on the fixed one. 
}

{For graph refinement, we construct a local neighborhood graph to approximate the manifold construction in the embedding space, preserving the local geometry of neighborhoods. Specifically, we first employ the extracted feature representation $\mathbf{Z}_{\emph{a}}$ and its similarity matrix $\mathbf{S}_{\emph{z}}=\left(\mathbf{Z}_{\emph{a}} \mathbf{Z}_{\emph{a}}^\mathsf{T}\right) / \left(\left\|\mathbf{Z}_{\emph{a}}\right\|_F \left\|\mathbf{Z}_{\emph{a}}^\mathsf{T}\right\|_F\right)$ to construct a graph $\mathbf{G}$ with its element
\begin{equation}
\begin{aligned}
& \emph{g}_{i,\emph{j}} = \left\{\begin{array}{ll}
\emph{s}_{i,\emph{j}} & \text { if } \emph{s}_{i,\emph{j}} == \max \left( \mathbf{S}_{i} \right), \\
0 & \text { otherwise,}
\end{array}\right. \\ 
& \rm{s.t.} \quad
\mathbf{S} = \mathbf{S}_{\emph{z}}-\rm{diag}\left(\mathbf{S}_{\emph{z}}\right),
\label{eq: cosine_Z}
\end{aligned}
\end{equation}
where $\emph{s}_{i,\emph{j}}$ is the $(\emph{i}, \emph{j})$-th element of $\mathbf{S}$, $\max \left( \mathbf{S}_{i} \right)$ is the maximum value of $\mathbf{S}$ in the $\emph{i}$-th row, $\rm{diag}\left(\mathbf{S}_{\emph{z}}\right)$ is a diagonal matrix containing the elements of the column vector of the main diagonal elements of $\mathbf{S}_{\emph{z}}$. 
Here, we use cosine similarity because it is not affected by the absolute magnitude of the node vector, which is important to indicate the similarity of the feature representation.
Such a graph construction way exclusively focuses on the nearest neighbor of samples, which is expected to be the most reliable information. Then, since a graph neural network can propagate information more reliably when self-loops are considered \cite{topping2022understanding}, we apply the self-loop operation to normalize $\mathbf{G}$, and then symmetrize it, i.e.,
\begin{equation}
\begin{aligned}
& \emph{g}_{i,\emph{j}} = \emph{g}_{j,\emph{i}} = \left\{\begin{array}{ll}
\max \{ \emph{g}_{\emph{i},\emph{j} }, \emph{g}_{j,\emph{i}} \} & \text { if } \emph{i} \neq \emph{j}, \\
1 & \text { if } \emph{i} == \emph{j},
\end{array}\right.
\label{eq: norm1_G}
\end{aligned}
\end{equation}
where $\rm{max} \{ \emph{g}_{\emph{i},\emph{j} }, \emph{g}_{\emph{j},\emph{i}} \}$ denotes the maximum value among $\emph{g}_{\emph{i},\emph{j} }$ and $\emph{g}_{\emph{j},\emph{i}}$, and we can obtain the adjacency matrix $\mathbf{A}_{z}=\mathbf{D}_{z}^{-1}\mathbf{G}$ with its degree matrix $\mathbf{D}_{z}$. Then, we build a multilayer perceptron layer parametrized by a weight matrix $\mathbf{W}^\mathbf{A}\in\mathbb{R}^{2n \times 2}$ to capture the relationship among the adjacency matrix $\mathbf{A}$ of the initial graph and the constructed adjacency matrix $\mathbf{A}_{z}$ with a normalization operation, which can be formulated as 
\begin{equation}
\begin{aligned}
\mathbf{V}=[\mathbf{v}_{z}\| \mathbf{v}]=
\ell_{2}\left( \emph{Softmax} \left(\rm{LReLU}\left(\left[\mathbf{A}_{z}\|\mathbf{A}\right]\mathbf{W}^\mathbf{A}\right)\right) \right),
\label{eq: GR-weight}
\end{aligned}
\end{equation}
where $\mathbf{V}\in\mathbb{R}^{\emph{n}\times 2}$ is the weight coefficient matrix with entries being greater than $0$, $\mathbf{v}_{z}$ and $\mathbf{v}$ are the weight vectors for measuring the importance of $\mathbf{A}_{z}$ and $\mathbf{A}$, respectively. Thus, we can fuse $\mathbf{A}_{z}$ and $\mathbf{A}$ in an adaptive fusion manner as
\begin{equation}
\begin{aligned}
& \mathbf{A}_{f}= \left( \mathbf{v}_{z}\mathbf{1} \right) \odot \mathbf{A}_\emph{z} + \left( \mathbf{v}\mathbf{1} \right) \odot \mathbf{A},
\label{eq: ga}
\end{aligned}
\end{equation}
where $\mathbf{1}\in\mathbb{R}^{1\times \emph{n}}$ denotes the vector of all ones. Finally, we iteratively perform graph refinement to boost the adjacency matrix of the graph, promoting graph embedding learning. \textcolor{black}{In each iteration, only the nearest neighbor connection information is added to the graph fusion process. In addition, as the number of iterations increases, more information is incorporated into the adjacency matrix. The adjacency matrix is updated every $\emph{i}_{\it p}$ epoch, with $\emph{i}_{\it p}=10$ being used empirically. More experiments and analyses are given in Section \ref{sec: hyper}. 2.}

\subsection{Clustering Optimization}\label{sec: clu_opt}
\textcolor{black}{
Since clustering is an unsupervised task, we have designed a practical end-to-end optimization method to train the network in a unified optimization manner. We start by calculating the similarity between an embedding $\mathbf{h}_{i}$ and its centroid $\boldsymbol{\mu}_j$ using the Student's t-distribution function \cite{helmert1876genauigkeit,student1908probable}. This yields a similarity that can be represented as a probability distribution $\mathbf{Q}$ and we can formulate its ($\emph{i}, \emph{j}$)-th element as
\begin{equation}
\begin{aligned}
&q_{i,{\it j}} = \frac{(1+\|\mathbf{h}_{i}-\boldsymbol{\mu}_{{\it j}}\|^2/\alpha)^{-\frac{\alpha+1}{2}}}{ \sum_{{\it j}^{'}} (1+\|\mathbf{h}_{i}-\boldsymbol{\mu}_{{\it j}^{'}}\|^2/\alpha)^{-\frac{\alpha+1}{2}} },
\label{eq: KL-q}
\end{aligned}
\end{equation}
where $\mathbf{H}=\mathbf{H}_{l}=[\mathbf{h}_1,\cdots,\mathbf{h}_{n}]^\mathsf{T}$, $\boldsymbol{\mu}_{\it j}$ is obtained with K-means \cite{macqueen1967some} based on the feature $\mathbf{h}_{\it j}$, and $\alpha$ is set to $1$. 
Then, we derive an auxiliary distribution from the element of $\mathbf{Z}_{a}$ (i.e., $z_{i,{\it j}}$) following \cite{xie2016unsupervised}, i.e.,
\begin{equation}
\begin{aligned}
&p_{i,{\it j}}=\frac{ z_{i,{\it j}}^{2}/\sum_{i} z_{i,{\it j}} } {\sum_{\emph{j}^{'}} z_{i,\emph{j}^{'}}^{2}/\sum_{i} z_{i,{\it j}^{'}}},
\label{eq: KL-p}
\end{aligned}
\end{equation}
where $1\geq p_{i,{\it j}}\geq 0$ is the element of the auxiliary distribution $\mathbf{P}$. Finally, to achieve the consistent alignment of multiple derived distributions, we minimize the Jeffreys divergence between $\mathbf{Q}$, $\mathbf{Z}_{a}$, and $\mathbf{P}$. By combining with Eq. (\ref{eq: DAE}), the loss function of our method can be written as
\begin{equation}
\begin{aligned}
& \min_{\mathbf{Z}_{a}}
\left\| \mathbf{X} - \hat{\mathbf{X}} \right\|^2_F \\
& + \lambda_1 \sum_\emph{i}^\emph{n}\sum_\emph{j}^\kappa
\left(
{\emph{p}_{\emph{i},\emph{j}}log{\frac{\emph{p}_{\emph{i},\emph{j}}}{\emph{z}_{i,\emph{j}}}}}
+
{\emph{z}_{\emph{i},\emph{j}}log{\frac{\emph{z}_{\emph{i},\emph{j}}}{\emph{p}_{i,\emph{j}}}}}
\right) \\
& + \lambda_2 \sum_\emph{i}^\emph{n}\sum_\emph{j}^\kappa
\left( 
\emph{p}_{\emph{i},\emph{j}}log{\frac{\emph{p}_{\emph{i},\emph{j}}}{\emph{q}_{i,\emph{j}}}}
+
\emph{q}_{\emph{i},\emph{j}}log{\frac{\emph{q}_{\emph{i},\emph{j}}}{\emph{p}_{i,\emph{j}}}}
\right) \\
& + \lambda_3 \sum_\emph{i}^\emph{n}\sum_\emph{j}^\kappa
\left( \emph{z}_{\emph{i},\emph{j}}log{\frac{\emph{z}_{\emph{i},\emph{j}}}{\emph{q}_{i,\emph{j}}}}
+
\emph{q}_{\emph{i},\emph{j}}log{\frac{\emph{q}_{\emph{i},\emph{j}}}{\emph{z}_{i,\emph{j}}}}
\right),
\label{eq: DGAC_loss}
\end{aligned}
\end{equation}
where $\lambda_1, \lambda_2, \lambda_3>0$ are the trade-off parameters.
When the network training is well accomplished, $\mathbf{Z}_{a}$ can directly indicate the clustering assignments as
\begin{equation}
\begin{aligned}
& y_{i}=\mathop{\arg\max}_{{\it j}} {z}_{i,{\it j}} \\ 
& \rm{s.t.} \quad {\it j}=1,\cdots,\kappa,
\label{eq: clustering_result}
\end{aligned}
\end{equation}
where $y_{i}$ is the inferred prediction with respect to (w.r.t.) the data $\textbf{x}_{i}$.  The training process of our method EGRC-Net is summarized in Alg. \ref{alg1}.
}

\subsection{Scalable EGRC-Net} \label{sec: scalable}
\textcolor{black}{
We further develop EGRC-Net in a scalable manner to explore its scalability to handle the practical large-graph problem \cite{ying2018graph,zhang2022embedding,park2022cgc}. It is well known that the scalability of GCN-based clustering methods has been limited due to the utilization of GCN since its explicit message-passing leads to an expensive neighborhood expansion \cite{bojchevski2020scaling}. Thus, we early utilize the approximate personalized propagation of neural predictions (APPNP) \cite{gasteigerpredict} instead of multi-layer graph convolution to complete the transmission and aggregation of node information on the graph. However, we observe that APPNP requires pre-specifying a heuristic threshold that determines the teleport probability. Besides, its value limits its availability in real-world applications since there is a little guideline to set a suitable threshold in clustering tasks. To this end, we empirically use a random number generator to sample from a uniform distribution on the interval $[0,1)$. Then, we set it as trainable and calculate its gradient with the backpropagation. Afterward, we can achieve the information propagation by using the personalized PageRank with a learned teleport probability value rather than relying on predefined heuristic thresholds. By imposing the designed IAPPNP module, the graph embedding learning of the scalable EGRC-Net formulation can be written as
\begin{equation}
\begin{aligned}
& \mathbf{Z}_{a}=\mathbf{E}_{\tau} \quad \rm{s.t.} \quad \iota=1,...,\tau, \\ 
& \mathbf{E}_{\iota} = \emph{Softmax}\left( \left(1-\Theta\right)\mathbf{E}_{0}+\Theta\mathbf{A}\mathbf{E}_{\iota-1}\right),
\label{eq: z-APPNP}
\end{aligned}
\end{equation}
where $\mathbf{Z}_{a}$ is the final output of IAPPNP, $\Theta$ is the learned threshold, $\mathbf{E}_{0}$ is the linear mapping of $\mathbf{X}$ via MLP. We uniformly set $\tau=1$ here, and more specific analyses are given in Section \ref{sec: ana_scalable}.
}

\begin{algorithm}[tb]
	\caption{Training process of the proposed EGRC-Net}
	\label{alg1}
	\textbf{Input}: Input data $\mathbf{X}$; Adjacency matrix $\mathbf{A}$; Parameters $\lambda_1,\lambda_2, \lambda_3$; Cluster number $\kappa$;\\
	\raggedright\textbf{Output}: Clustering assignments $\mathbf{y}$;\\
	\begin{algorithmic}[1]
    	\STATE Let $i_{\it Iter}=1$, $i_{\it all}=200$, $\emph{i}_{\it p}=10$;
    	\STATE Initialize the parameters of DAE;
		\WHILE{$i_{\it Iter} < i_{\it all} $}
		\STATE Calculate the matrix $\mathbf{H}$ by Eq. (\ref{eq: DAE});
		\STATE Calculate the matrix $\mathbf{Z}_{a}$ by Eq. (\ref{eq: Z});
		\IF{$i_{\it Iter}\%\emph{i}_{\it p}==0$} 
            \STATE Update the adjacency matrix $\mathbf{A}$ by Eq. (\ref{eq: ga});
            \ENDIF
		\STATE Calculate the distribution $\mathbf{Q}$ by Eq. (\ref{eq: KL-q});
		\STATE Calculate the distribution $\mathbf{P}$ by Eq. (\ref{eq: KL-p});
		\STATE Calculate the overall loss function by Eq. (\ref{eq: DGAC_loss});
		\STATE Initialize the parameter gradient of the network to zero; perform the backpropagation; update all parameters of the network;
		\STATE $i_{\it Iter} = i_{\it Iter} + 1$;
		\ENDWHILE \\
		\STATE Obtain the final clustering assignments $\mathbf{y}$ by Eq. (\ref{eq: clustering_result});
	\end{algorithmic}
\end{algorithm}

\subsection{Computational Complexity} \label{sec: comp}
\textcolor{black}{
For DAE, the computational complexity is $\mathcal{O}(n\sum_{i=2}^{l}d_{i-1}d_{i})$; for GCN, the computational complexity is $\mathcal{O}(|\mathcal{E}|\sum_{i=2}^{l}d_{i-1}d_{i})$ corresponding to \cite{pan2019learning}, where $|\mathcal{E}|$ is the number of edges; for IAPPNP, the computational complexity is $\mathcal{O}(|\mathcal{E}|d\kappa)$; for Eq. (\ref{eq: KL-q}), the computational complexity is $\mathcal{O}(n\kappa+n\log n)$ corresponding to \cite{xie2016unsupervised}; for the MLP modules, the computational complexity is $\mathcal{O}(\sum_{i=1}^{l-1}(d_{i})+(\sum_{i=1}^{l+1} d_{i})(l+1))$; for the graph refinement module, the computational complexity is $n$. 
Thus, the total computational complexity of EGRC-Net in one iteration is $\mathcal{O}(n\sum_{i=2}^{l}d_{i-1}d_{i} + |\mathcal{E}|\sum_{i=2}^{l}d_{i-1}d_{i} + n\kappa+n\log n + \sum_{i=1}^{l+1}d_{i} + l\sum_{i=1}^{l+1} d_{i} + n)$, while that of the scalable one is $\mathcal{O}(n\sum_{i=2}^{l}d_{i-1}d_{i} + |\mathcal{E}|d\kappa + n\kappa+n\log n + \sum_{i=1}^{l+1}d_{i} + l\sum_{i=1}^{l+1} d_{i} + n)$. 
}

\section{Experiments}\label{sec: eprm}
\textcolor{black}{
In this section, we first present the used benchmark datasets (Section \ref{sec: data}), the compared methods (Section \ref{sec: methods}), the evaluation metrics (Section \ref{sec: metrics}), and the network implementation details (Section \ref{sec: details}). 
Then, we conduct quantitative experiments and analyses (Section \ref{sec: results}) to evaluate the effectiveness of our method EGRC-Net. Afterward, we further investigate EGRC-Net by conducting parameters analyses (Section \ref{sec: hyper}), model stability analysis (Section \ref{sec: conver}), ablation studies (Section \ref{sec: AS}), and visual comparisons in Section \ref{sec: visua}. 
Furthermore, we design the experiments w.r.t. the different strategies of the teleport probability value and the number of propagation steps to investigate the effectiveness of the scalable EGRC-Net (Section \ref{sec: ana_scalable}). Finally, we compare our methods and the state-of-the-art approaches w.r.t. running time and network parameters (Section \ref{sec: time-para}).
}

\begin{table}[!t]
\centering
\caption{Summary of the used datasets.}
\label{tab: datasets}
\resizebox{0.88\columnwidth}{!}{
\begin{tabular}{c|c|c|c|c|c}
\hline\hline
Dataset     & Type   & Samples & Dimension       & Classes & Edges \\ 
\hline\hline
USPS        & Image  & 9298    & 256             & 10      & 27894       \\
STL10       & Image  & 13000   & 512             & 10      & 39000       \\
ImageNet-10 & Image  & 13000   & 512             & 10      & 39000       \\
HHAR        & Record & 10299   & 561             & 6       & 30897       \\
REUT        & Text   & 10000   & 2000            & 4       & 30000      \\
\hline
ACM         & Graph  & 3025    & 1870            & 3       & 13128      \\
CITE        & Graph  & 3327    & 3703            & 6       & 4552      \\ 
DBLP        & Graph  & 4057    & 334             & 4       & 3528       \\ 
PubMed      & Graph  & 19717   & 500             & 3      & 44324\\ 
\hline\hline
\end{tabular}
}
\end{table}

\subsection{Datasets} \label{sec: data}
 We conducted the comparisons on nine commonly used datasets, including USPS \cite{le1990handwritten}, STL10 \cite{coates2011analysis}, ImageNet-10 \cite{deng2009imagenet}, HHAR \cite{stisen2015smart}), REUT \cite{lewis2004rcv1}, ACM$\footnote{http://dl.acm.org}$, CITE$\footnote{http://CiteSeerx.ist.psu.edu/}$, DBLP$\footnote{https://dblp.uni-trier.de}$, and PubMed \cite{sen2008collective}, which are summarized in Table \ref{tab: datasets}. Particularly, We adopt ResNet34 \cite{he2016deep} to preprocess STL10 and ImageNet-10, where its output dimension of the fully-connected classification layer is set to 512.
 The corresponding details are provided as follows. 
\begin{itemize}
    \item	\textcolor{black}{\textbf{USPS}. The USPS dataset is the earliest version of USPS, which has ten classes, labeled `0' to `9'. It comprises 9,298 handwritten digit images, all of which have been standardized to a uniform size $16 \times 16$.}
    \item	\textcolor{black}{\textbf{STL10}. The STL10 dataset is an image dataset derived from ImageNet and popularly used to evaluate algorithms of unsupervised feature learning. It contains 13,000 labeled images from 10 object classes (such as birds, cats, trucks).}
    \item	\textcolor{black}{\textbf{ImageNet-10}. The ImageNet-10 dataset is a small-scale subset of the ImageNet database, containing 13,000 labeled images from 10 object classes. Although significantly smaller, it retains the structure and diversity of the original ImageNet dataset.} 
    \item	\textbf{HHAR}. The HHAR dataset encompasses 10,299 sensor readings collected from smartphones and smartwatches. These readings have been divided into six categories that correspond to different human activities, including biking, sitting, standing, walking, stair up, and stair down.
    \item	\textcolor{black}{\textbf{REUT}. The REUT dataset comprises a compilation of English news articles, organized into categorical labels. Due to scalability limitations of certain algorithms with the full Reuters (REUT) dataset (685,071 samples), we employ a widely used smaller version (10,000 samples) of the REUT dataset for comparison purposes \cite{xie2016unsupervised}.}
    \item 	\textbf{ACM}. The ACM dataset is a network of scholarly papers sourced from the ACM digital library, where two papers are linked with an edge if they share a common author. The features of the papers were derived from keywords associated with the KDD, SIGMOD, SIGCOMM, and MobiCOMM conferences, and were categorized into three classes based on their research area: database, wireless communication, and data mining.
    \item	\textbf{CITE}. The CITE dataset is a network of citations comprised of sparse bag-of-words feature vectors for each document and a collection of citation links between documents. The labels for the documents are organized into six distinct research areas: agents, artificial intelligence, database, information retrieval, machine language, and human-computer interaction.
    \item	\textbf{DBLP}. The DBLP dataset is a network of authors sourced from the DBLP computer science bibliography, where two authors are linked with an edge if they have a co-author relationship. The author's features are represented as elements in a bag-of-words, constructed from keywords, and the authors are divided into four research areas: database, data mining, machine learning, and information retrieval. The authors are labeled based on the conference they have submitted to.
    \item	\textbf{PubMed}. The PubMed dataset comprises 19,717 scientific publications related to diabetes, classified into one of three categories, sourced from the PubMed database. The citation network within the dataset includes 44,324 links. Each publication is represented by a Term Frequency/Inverse Document Frequency weighted word vector, derived from a dictionary of 500 unique words.
\end{itemize}

\subsection{Compared Methods} \label{sec: methods}
\textcolor{black}{
The compared methods include K-means \cite{macqueen1967some}, three DAE-based clustering methods (DAE \cite{hinton2006reducing}, DEC \cite{xie2016unsupervised}, and IDEC \cite{guo2017improved}), eight GCN-based clustering methods ({GAE} \cite{kipf2016variational}, {VGAE} \cite{kipf2016variational}, {ARGA} \cite{pan2018adversarially}, {DAEGC} \cite{wang2019attributed}, FastGAE \cite{salha2021fastgae}, RG-VGAE \cite{mrabah2022rethinking}, MA-GAE \cite{salha2022modularity}, and {EGAE} \cite{zhang2022embedding}), as well as three DAE and GCN combination-based clustering methods ({SDCN} \cite{bo2020structural}, {AGCN} \cite{peng2021attention}, and {AGCC} \cite{he2022parallelly}).
}

\begin{table*}[]
\centering
\caption{\textcolor{black}{Experimental comparisons. We highlighted the best and second-best results with \textbf{bold} and \underline{underline}, respectively. ‘OOM’ denotes the out-of-memory issue.}}
\label{tab: all_results}
\resizebox{1.38\columnwidth}{!}{
\begin{tabular}{c|c|cccccccc}
\hline\hline    
\multirow{2}{*}{Datasets}   & \multirow{2}{*}{Metrics} &  K-means  \cite{macqueen1967some}  & DAE \cite{hinton2006reducing}         & DEC \cite{xie2016unsupervised}         & IDEC \cite{guo2017improved}        & GAE \cite{kipf2016variational}         & VGAE \cite{kipf2016variational}        & ARGA \cite{pan2018adversarially}        & DAEGC \cite{wang2019attributed}                \\
                            &                          &              & [Science06]           & [ICML16]     & [AAAI17]     & [NIPS16]     & [NIPS16]              & [IJCAI18]             & [AAAI19]              \\
\hline\hline
\multirow{3}{*}{USPS}       & ARI                      & 54.55$\pm$0.06 & 58.83$\pm$0.05          & 63.70$\pm$0.27 & 67.86$\pm$0.12 & 50.30$\pm$0.55 & 40.96$\pm$0.59          & 51.10$\pm$0.60          & 63.33$\pm$0.34          \\
                            & ACC                      & 66.82$\pm$0.04 & 71.04$\pm$0.03          & 73.31$\pm$0.17 & 76.22$\pm$0.12 & 63.10$\pm$0.33 & 56.19$\pm$0.72          & 66.80$\pm$0.70          & 73.55$\pm$0.40          \\
                            & NMI                      & 62.63$\pm$0.05 & 67.53$\pm$0.03          & 70.58$\pm$0.25 & 75.56$\pm$0.06 & 60.69$\pm$0.58 & 51.08$\pm$0.37          & 61.60$\pm$0.30          & 71.12$\pm$0.24          \\
\hline\multirow{3}{*}{STL10}      & ARI                      & 67.24$\pm$0.12 & 69.13$\pm$0.17          & 08.08$\pm$1.13 & 36.76$\pm$4.51 & 70.6$\pm$1.38  & 66.95$\pm$2.93          & 45.30$\pm$6.02          & 84.03$\pm$0.85          \\
                            & ACC                      & 83.53$\pm$0.07 & 84.36$\pm$0.11          & 28.34$\pm$1.10 & 53.95$\pm$2.93 & 82.82$\pm$0.51 & 79.65$\pm$1.97          & 68.28$\pm$4.08          & 92.22$\pm$0.42          \\
                            & NMI                      & 75.43$\pm$0.03 & 76.18$\pm$0.11          & 13.81$\pm$1.53 & 50.78$\pm$4.53 & 82.43$\pm$0.76 & 79.25$\pm$1.88          & 58.35$\pm$5.11          & 84.46$\pm$0.64          \\
\hline\multirow{3}{*}{ImageNet-10} & ARI                      & 69.03$\pm$0.05 & 69.73$\pm$0.62          & 03.00$\pm$0.49 & 03.24$\pm$1.22 & 56.78$\pm$4.98 & 72.74$\pm$1.47          & 25.48$\pm$5.08          & 11.00$\pm$1.21          \\
                            & ACC                      & 76.70$\pm$0.02 & 77.36$\pm$1.33          & 19.61$\pm$0.79 & 18.36$\pm$1.21 & 73.66$\pm$3.22 & 83.10$\pm$0.51          & 47.67$\pm$6.81          & 30.68$\pm$1.85          \\
                            & NMI                      & 75.89$\pm$0.04 & 76.44$\pm$0.36          & 05.63$\pm$0.90 & 05.38$\pm$1.75 & 66.00$\pm$2.83 & \underline{81.50$\pm$2.22} & 34.86$\pm$4.21          & 17.71$\pm$1.38          \\
\hline\multirow{3}{*}{HHAR}       & ARI                      & 46.09$\pm$0.02 & 60.36$\pm$0.88          & 61.25$\pm$0.51 & 62.83$\pm$0.45 & 42.63$\pm$1.63 & 51.47$\pm$0.73          & 44.70$\pm$1.00          & 60.38$\pm$2.15          \\
                            & ACC                      & 59.98$\pm$0.02 & 68.69$\pm$0.31          & 69.39$\pm$0.25 & 71.05$\pm$0.36 & 62.33$\pm$1.01 & 71.30$\pm$0.36          & 63.30$\pm$0.80          & 76.51$\pm$2.19          \\
                            & NMI                      & 58.86$\pm$0.01 & 71.42$\pm$0.97          & 72.91$\pm$0.39 & 74.19$\pm$0.39 & 55.06$\pm$1.39 & 62.95$\pm$0.36          & 57.10$\pm$1.40          & 69.10$\pm$2.28          \\
\hline\multirow{3}{*}{REUT}       & ARI                      & 27.95$\pm$0.38 & 49.55$\pm$0.37          & 48.44$\pm$0.14 & 51.26$\pm$0.21 & 19.61$\pm$0.22 & 26.18$\pm$0.36          & 24.50$\pm$0.40          & 31.12$\pm$0.18          \\
                            & ACC                      & 54.04$\pm$0.01 & 74.90$\pm$0.21          & 73.58$\pm$0.13 & 75.43$\pm$0.14 & 54.40$\pm$0.27 & 60.85$\pm$0.23          & 56.20$\pm$0.20          & 65.50$\pm$0.13          \\
                            & NMI                      & 41.54$\pm$0.51 & 49.69$\pm$0.29          & 47.50$\pm$0.34 & 50.28$\pm$0.17 & 25.92$\pm$0.41 & 25.51$\pm$0.22          & 28.70$\pm$0.30          & 30.55$\pm$0.29          \\
\hline\multirow{3}{*}{ACM}        & ARI                      & 30.60$\pm$0.69 & 54.64$\pm$0.16          & 60.64$\pm$1.87 & 62.16$\pm$1.50 & 59.46$\pm$3.10 & 57.72$\pm$0.67          & 62.90$\pm$2.10          & 59.35$\pm$3.89          \\
                            & ACC                      & 67.31$\pm$0.71 & 81.83$\pm$0.08          & 84.33$\pm$0.76 & 85.12$\pm$0.52 & 84.52$\pm$1.44 & 84.13$\pm$0.22          & 86.10$\pm$1.20          & 86.94$\pm$2.83          \\
                            & NMI                      & 32.44$\pm$0.46 & 49.30$\pm$0.16          & 54.54$\pm$1.51 & 56.61$\pm$1.16 & 55.38$\pm$1.92 & 53.20$\pm$0.52          & 55.70$\pm$1.40          & 56.18$\pm$4.15          \\
\hline\multirow{3}{*}{CITE}       & ARI                      & 13.43$\pm$3.02 & 29.31$\pm$0.14          & 28.12$\pm$0.36 & 25.70$\pm$2.65 & 33.55$\pm$1.18 & 33.13$\pm$0.53          & 33.40$\pm$1.50          & 37.78$\pm$1.24          \\
                            & ACC                      & 39.32$\pm$3.17 & 57.08$\pm$0.13          & 55.89$\pm$0.20 & 60.49$\pm$1.42 & 61.35$\pm$0.80 & 60.97$\pm$0.36          & 56.90$\pm$0.70          & 64.54$\pm$1.39          \\
                            & NMI                      & 16.94$\pm$3.22 & 27.64$\pm$0.08          & 28.34$\pm$0.30 & 27.17$\pm$2.40 & 34.63$\pm$0.65 & 32.69$\pm$0.27          & 34.50$\pm$0.80          & 36.41$\pm$0.86          \\
\hline\multirow{3}{*}{DBLP}       & ARI                      & 06.97$\pm$0.39 & 12.21$\pm$0.43          & 23.92$\pm$0.39 & 25.37$\pm$0.60 & 22.02$\pm$1.40 & 17.92$\pm$0.07          & 22.70$\pm$0.30          & 21.03$\pm$0.52          \\
                            & ACC                      & 38.65$\pm$0.65 & 51.43$\pm$0.35          & 58.16$\pm$0.56 & 60.31$\pm$0.62 & 61.21$\pm$1.22 & 58.59$\pm$0.06          & 61.60$\pm$1.00          & 62.05$\pm$0.48          \\
                            & NMI                      & 11.45$\pm$0.38 & 25.40$\pm$0.16          & 29.51$\pm$0.28 & 31.17$\pm$0.50 & 30.80$\pm$0.91 & 26.92$\pm$0.06          & 26.80$\pm$1.00          & 32.49$\pm$0.45          \\
\hline\multirow{3}{*}{PubMed}     & ARI                      & 28.10$\pm$0.01 & 23.86$\pm$0.67          & 19.55$\pm$0.13 & 20.58$\pm$0.39 & 20.62$\pm$1.39 & 30.15$\pm$1.23          & 24.35$\pm$0.17          & 29.84$\pm$0.04          \\
                            & ACC                      & 59.83$\pm$0.01 & 63.07$\pm$0.31          & 60.14$\pm$0.09 & 60.70$\pm$0.34 & 62.09$\pm$0.81 & 68.48$\pm$0.77          & 65.26$\pm$0.12          & 68.73$\pm$0.03          \\
                            & NMI                      & 31.05$\pm$0.02 & 26.32$\pm$0.57          & 22.44$\pm$0.14 & 23.67$\pm$0.29 & 23.84$\pm$3.54 & 30.61$\pm$1.71          & 24.80$\pm$0.17          & 28.26$\pm$0.03          \\
\hline\hline
\multirow{2}{*}{Datasets}   & \multirow{2}{*}{Metrics} & SDCN \cite{bo2020structural}                 & AGCN \cite{peng2021attention}                & FastGAE \cite{salha2021fastgae}      & RG-VGAE \cite{mrabah2022rethinking}      & MA-GAE  \cite{salha2022modularity}     & EGAE  \cite{zhang2022embedding}       & AGCC \cite{he2022parallelly}               & \multirow{2}{*}{Our}  \\
                            &                          & [WWW20]      & [MM21]                & [NN21]       & [TKDE22]     & [NN22]       & [NNLS22]              & [NNLS22]              &                       \\
\hline\hline
\multirow{3}{*}{USPS}       & ARI                      & 71.84$\pm$0.24 & \underline{73.61$\pm$0.43} & 00.36$\pm$0.05 & 04.82$\pm$0.07 & 03.09$\pm$0.17 & 64.13$\pm$2.22          & 68.50$\pm$3.83          & \textbf{75.61$\pm$1.92} \\
                            & ACC                      & 78.08$\pm$0.19 & \underline{80.98$\pm$0.28} & 13.34$\pm$0.28 & 22.42$\pm$0.11 & 31.33$\pm$0.69 & 76.22$\pm$2.48          & 77.14$\pm$1.21          & \textbf{83.41$\pm$4.40} \\
                            & NMI                      & 79.51$\pm$0.27 & \underline{79.64$\pm$0.32} & 00.93$\pm$0.06 & 12.22$\pm$0.12 & 26.72$\pm$1.03 & 70.83$\pm$2.25          & 75.93$\pm$3.83          & \textbf{80.94$\pm$0.92} \\
\hline\multirow{3}{*}{STL10}      & ARI                      & 84.09$\pm$0.72 & 84.59$\pm$0.26          & 10.80$\pm$1.25 & 01.48$\pm$0.02 & 04.07$\pm$0.89 & \underline{85.73$\pm$0.14} & OOM                   & \textbf{85.92$\pm$0.04} \\
                            & ACC                      & 92.27$\pm$0.38 & 92.57$\pm$0.14          & 42.45$\pm$3.08 & 17.08$\pm$0.08 & 26.45$\pm$1.41 & \underline{93.13$\pm$0.07} & OOM                   & \textbf{93.23$\pm$0.02} \\
                            & NMI                      & 85.03$\pm$0.42 & 85.27$\pm$0.15          & 35.68$\pm$3.76 & 05.93$\pm$0.05 & 22.74$\pm$0.74 & \underline{85.98$\pm$0.14} & OOM                   & \textbf{86.07$\pm$0.02} \\
\hline\multirow{3}{*}{ImageNet-10} & ARI                      & 72.41$\pm$0.34 & \underline{73.38$\pm$0.73} & 06.04$\pm$1.90 & 05.29$\pm$0.15 & 04.01$\pm$0.64 & 69.84$\pm$4.25          & OOM                   & \textbf{78.38$\pm$3.62} \\
                            & ACC                      & 78.37$\pm$0.25 & 79.30$\pm$0.29          & 25.91$\pm$2.52 & 22.14$\pm$0.35 & 31.57$\pm$2.10 & \underline{84.66$\pm$2.62} & OOM                   & \textbf{87.29$\pm$3.90} \\
                            & NMI                      & 79.83$\pm$0.26 & 80.58$\pm$0.73          & 10.26$\pm$2.59 & 10.04$\pm$0.06 & 22.28$\pm$2.89 & 72.18$\pm$3.26          & OOM                   & \textbf{82.28$\pm$1.80} \\
\hline\multirow{3}{*}{HHAR}       & ARI                      & 72.84$\pm$0.09 & \underline{77.07$\pm$0.66} & 00.07$\pm$0.05 & 16.29$\pm$0.06 & 15.68$\pm$1.55 & 59.37$\pm$2.76          & 75.58$\pm$1.85          & \textbf{78.22$\pm$0.30} \\
                            & ACC                      & 84.26$\pm$0.17 & \underline{88.11$\pm$0.43} & 19.39$\pm$0.30 & 40.11$\pm$0.07 & 49.50$\pm$1.22 & 75.97$\pm$1.05          & 86.54$\pm$1.79          & \textbf{88.95$\pm$0.21} \\
                            & NMI                      & 79.90$\pm$0.09 & \underline{82.44$\pm$0.62} & 00.45$\pm$0.11 & 27.51$\pm$0.10 & 37.59$\pm$2.41 & 67.26$\pm$2.30          & 82.21$\pm$1.78          & \textbf{82.48$\pm$0.30} \\
\hline\multirow{3}{*}{REUT}       & ARI                      & 55.36$\pm$0.37 & 60.55$\pm$1.78          & 01.12$\pm$0.66 & 05.86$\pm$0.31 & 02.84$\pm$4.21 & 48.12$\pm$3.90          & \underline{62.98$\pm$2.24} & \textbf{63.53$\pm$0.66} \\
                            & ACC                      & 77.15$\pm$0.21 & 79.30$\pm$1.07          & 40.11$\pm$0.64 & 43.06$\pm$0.38 & 40.69$\pm$5.13 & 72.12$\pm$2.67          & \underline{81.65$\pm$1.52} & \textbf{81.90$\pm$0.16} \\
                            & NMI                      & 50.82$\pm$0.21 & 57.83$\pm$1.01          & 00.39$\pm$0.13 & 04.39$\pm$0.08 & 07.08$\pm$4.42 & 49.17$\pm$3.75          & \underline{59.56$\pm$0.94} & \textbf{60.32$\pm$0.40} \\
\hline\multirow{3}{*}{ACM}        & ARI                      & 73.91$\pm$0.40 & \underline{74.20$\pm$0.38} & 03.08$\pm$0.55 & 18.48$\pm$0.30 & 63.89$\pm$1.33 & 73.97$\pm$1.01          & 73.73$\pm$0.90          & \textbf{76.04$\pm$0.39} \\
                            & ACC                      & 90.45$\pm$0.18 & \underline{90.59$\pm$0.15} & 43.25$\pm$1.45 & 63.85$\pm$0.23 & 86.45$\pm$0.56 & 90.51$\pm$0.41          & 90.38$\pm$0.38          & \textbf{91.30$\pm$0.17} \\
                            & NMI                      & 68.31$\pm$0.25 & \underline{68.38$\pm$0.45} & 13.16$\pm$2.34 & 32.58$\pm$0.25 & 57.31$\pm$1.31 & 67.71$\pm$0.92          & 68.34$\pm$0.89          & \textbf{70.40$\pm$0.30} \\
\hline\multirow{3}{*}{CITE}       & ARI                      & 40.17$\pm$0.43 & \underline{43.79$\pm$0.31} & 03.91$\pm$2.07 & 37.65$\pm$0.41 & 28.73$\pm$2.89 & 38.72$\pm$1.00          & 41.82$\pm$2.03          & \textbf{48.32$\pm$0.57} \\
                            & ACC                      & 65.96$\pm$0.31 & \underline{68.79$\pm$0.23} & 28.31$\pm$2.62 & 61.75$\pm$0.39 & 55.41$\pm$3.10 & 64.38$\pm$0.81          & 68.08$\pm$1.44          & \textbf{72.27$\pm$0.37} \\
                            & NMI                      & 38.71$\pm$0.32 & \underline{41.54$\pm$0.30} & 05.61$\pm$3.32 & 38.27$\pm$0.32 & 30.52$\pm$1.72 & 38.34$\pm$0.98          & 40.86$\pm$1.45          & \textbf{45.77$\pm$0.48} \\
\hline\multirow{3}{*}{DBLP}       & ARI                      & 39.15$\pm$2.01 & 42.49$\pm$0.31          & 02.42$\pm$0.46 & 08.03$\pm$0.40 & 06.17$\pm$1.64 & 31.64$\pm$2.34          & \underline{44.40$\pm$3.79} & \textbf{56.39$\pm$0.93} \\
                            & ACC                      & 68.05$\pm$1.81 & 73.26$\pm$0.37          & 33.85$\pm$1.04 & 42.61$\pm$0.38 & 40.43$\pm$1.31 & 63.24$\pm$2.17          & \underline{73.45$\pm$2.16} & \textbf{80.53$\pm$0.54} \\
                            & NMI                      & 39.50$\pm$1.34 & 39.68$\pm$0.42          & 03.33$\pm$1.05 & 11.76$\pm$0.32 & 11.36$\pm$1.37 & 31.26$\pm$2.42          & \underline{40.36$\pm$2.81} & \textbf{50.85$\pm$0.60} \\
\hline\multirow{3}{*}{PubMed}     & ARI                      & 22.30$\pm$2.07 & 31.39$\pm$0.67          & 12.08$\pm$6.58 & 07.30$\pm$0.13 & 30.55$\pm$2.82 & \underline{33.00$\pm$1.74} & OOM                   & \textbf{34.45$\pm$0.58} \\
                            & ACC                      & 64.20$\pm$1.30 & 69.67$\pm$0.42          & 53.01$\pm$5.45 & 52.51$\pm$0.58 & 68.82$\pm$1.57 & \underline{70.69$\pm$1.25} & OOM                   & \textbf{71.91$\pm$0.22} \\
                            & NMI                      & 22.87$\pm$2.04 & 30.96$\pm$0.99          & 18.09$\pm$4.34 & 07.17$\pm$0.33 & 29.86$\pm$2.76 & \underline{31.54$\pm$1.95} & OOM                   & \textbf{32.15$\pm$1.81} \\
\hline\hline
\end{tabular}
}
\end{table*}

\subsection{Evaluation Metrics} \label{sec: metrics}
\textcolor{black}{
The performance of the clustering was quantitatively evaluated using three commonly used metrics: adjusted rand index (ARI), accuracy (ACC), and normalized mutual information (NMI). The larger values of these metrics indicate a higher quality of clustering. 
}

\subsection{Implementation Details} \label{sec: details}
\textcolor{black}{
For fair comparisons, the dimensions of the vanilla DAE and GCN layers were set to $500$-$500$-$2000$-$10$, as in previous studies \cite{xie2016unsupervised,bo2020structural,peng2021attention,he2022parallelly}. The DAE module was first pre-trained for $30$ epochs using a learning rate of $0.001$. The entire network was then fine-tuned with $\emph{i}_{\it all}=200$. The learning rate was set to $0.001$ for the USPS, HHAR, ACM, DBLP, and PubMed datasets and $0.0001$ for the REUT and CITE datasets. \textcolor{black}{For the approaches FastGAE, RG-VGAE, MA-GAE, EGAE, and AGCC, we used their publicly available codes and parameter settings given by the original papers \cite{salha2021fastgae,mrabah2022rethinking,salha2022modularity,zhang2022embedding,he2022parallelly}. For other methods under comparison, we directly cited the results in \cite{peng2022deep}. The results were reported as the mean values and their corresponding standard deviations, obtained by repeating the experiments ten times, represented as mean $\pm$ std in the experimental results. 
The training was implemented using PyTorch on a GeForce RTX 2080 Ti, a GeForce RTX 3090, and a Quadro RTX 8000 GPU.} } 

\subsection{Quantitative Results} \label{sec: results}
\textcolor{black}{
Table \ref{tab: all_results} provides the quantitative comparisons of the proposed method EGRC-Net and fifteen compared approaches on nine benchmark datasets w.r.t. three metrics, where we can draw the following conclusions. 
\begin{itemize}
    \item EGRC-Net obtains the best experimental results in all metrics on nine benchmark datasets. For instance, in \textbf{non-graph} dataset USPS, our EGRC-Net enhances the ARI, ACC, and NMI metrics of the second-best approach AGCN \cite{peng2021attention} by 2.00\%, 2.43\%, and 1.30\% respectively. In the \textbf{graph} dataset CITE, our EGRC-Net improves 4.53\% over AGCN on ARI, 3.48\% on ACC, and 4.23\% on NMI.
    \item DAEGC \cite{wang2019attributed} outperforms GAE \cite{kipf2016variational} in almost all metrics on clustering performance, validating the effectiveness of exploiting an adaptive fusion mechanism. In this paper, we develop the embedding learning module and the graph refinement architecture in an adaptive fusion manner to flexibly explore the multiple off-the-shelf information, boosting graph clustering performance.
    \item AGCC \cite{he2022parallelly} obtains the second-best results on the commonly used graph dataset DBLP; however, it suffers from the out-of-memory case on the larger graph PubMed. Differently, our method EGRC-Net can get the best results on PubMed at the cost of acceptable resource consumption. Such a phenomenon also emphasizes the importance of network scalability.
    \item EGRC-Net obtains a significant improvement on DBLP, e.g., we enhance the ARI, ACC, and NMI metrics of the best-compared approach AGCC \cite{he2022parallelly} by 11.99\%, 7.08\%, and 10.49\% respectively. The reason may be attributed to the fact that DBLP has low feature dimensions, i.e., little learnable information, meaning that sufficiently utilizing the available off-the-shelf embedding and graph information greatly improves the clustering performance.
\end{itemize}
}

\begin{figure*}[]
	\centering
    \subfigure[$\lambda_3$=0.001 (ACC)]{
	\includegraphics [width=0.2800\columnwidth]{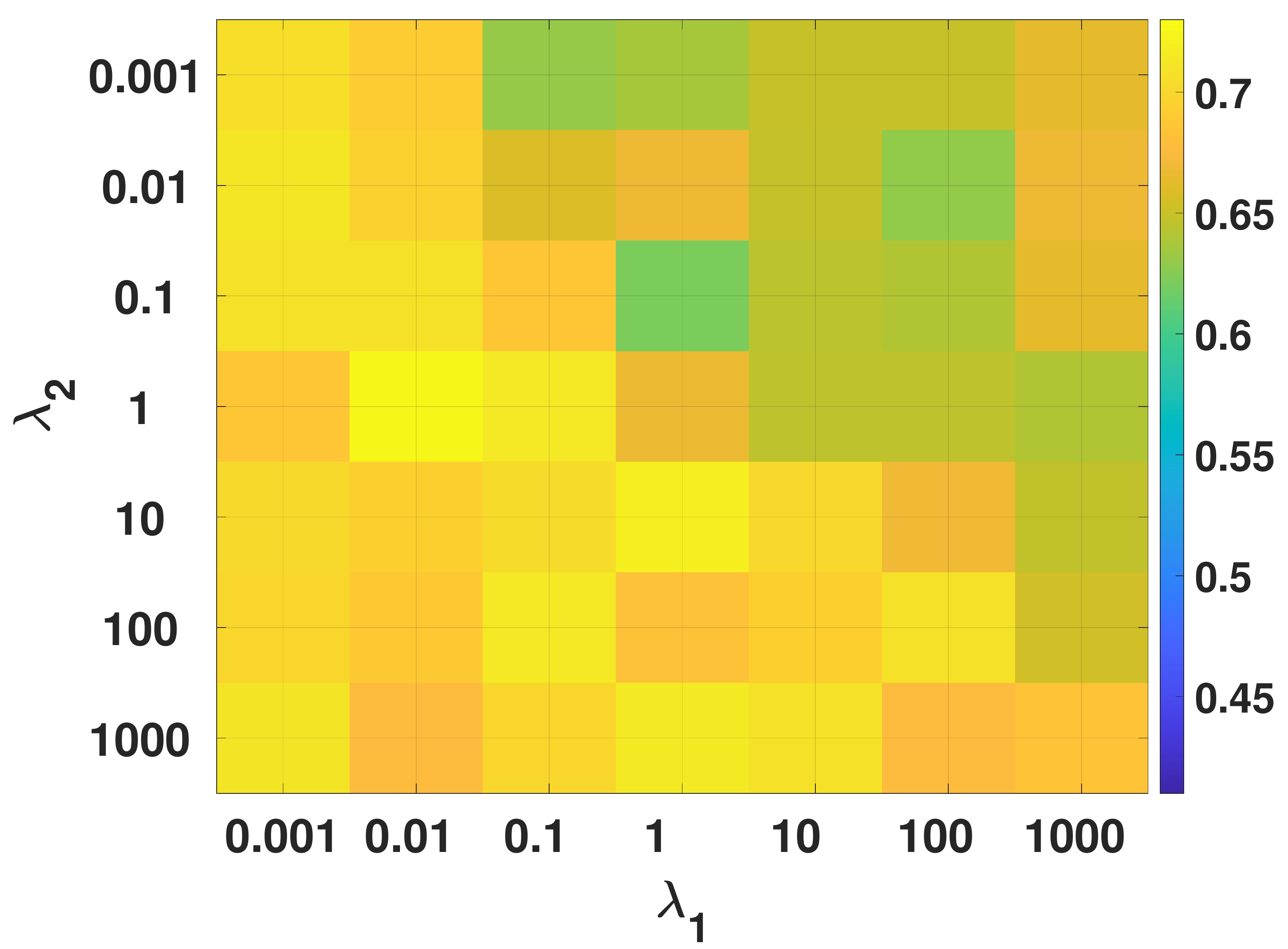}
     }
    \subfigure[$\lambda_3$=0.1 (ACC)]{
	\includegraphics [width=0.2800\columnwidth]{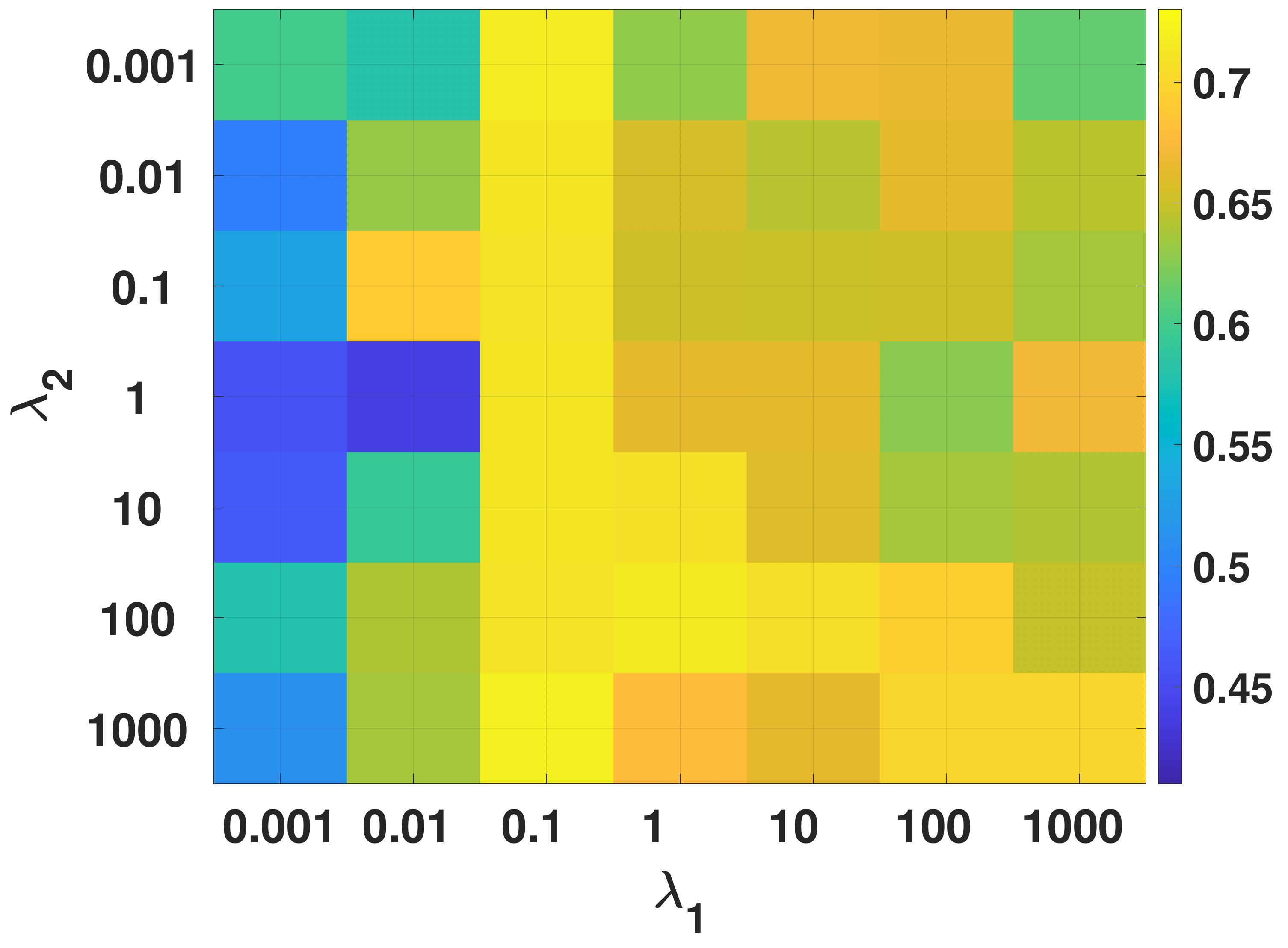}
     }
    \subfigure[$\lambda_3$=10 (ACC)]{
	\includegraphics [width=0.2800\columnwidth]{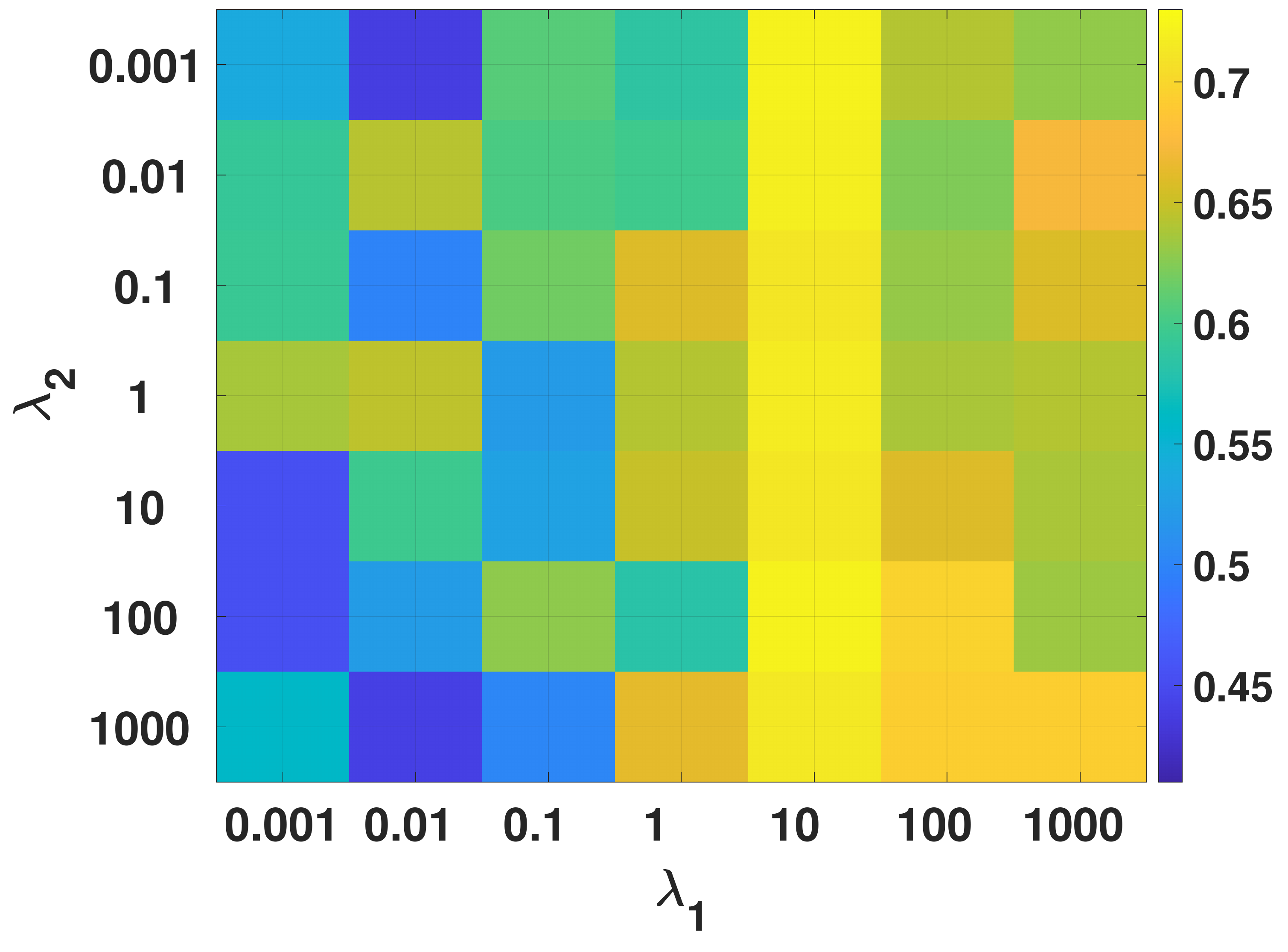}
     }
     \subfigure[$\lambda_3$=1000 (ACC)]{
	\includegraphics [width=0.2800\columnwidth]{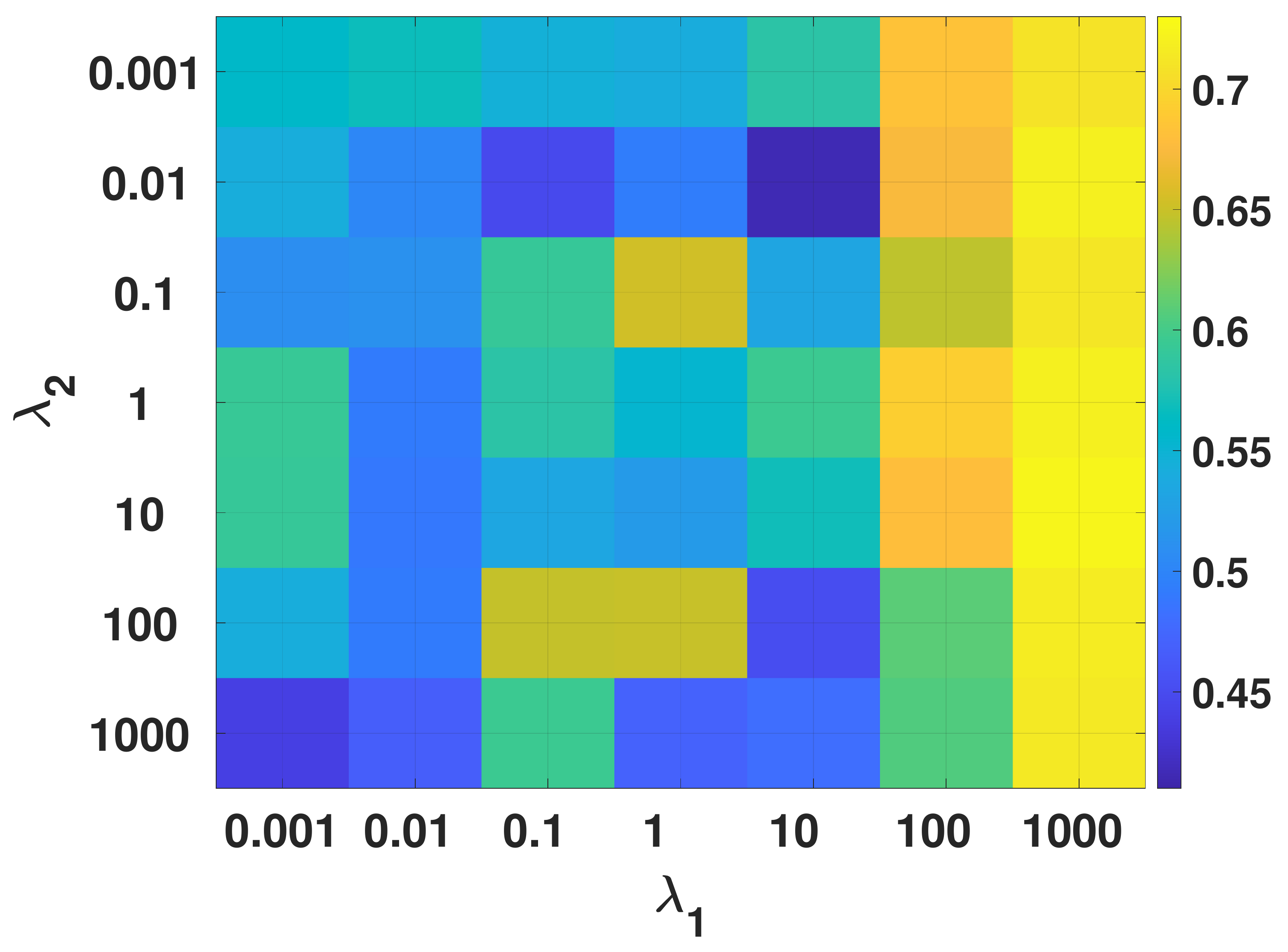}
     }
    \subfigure[$\lambda_3$=0.001 (NMI)]{
	\includegraphics [width=0.2800\columnwidth]{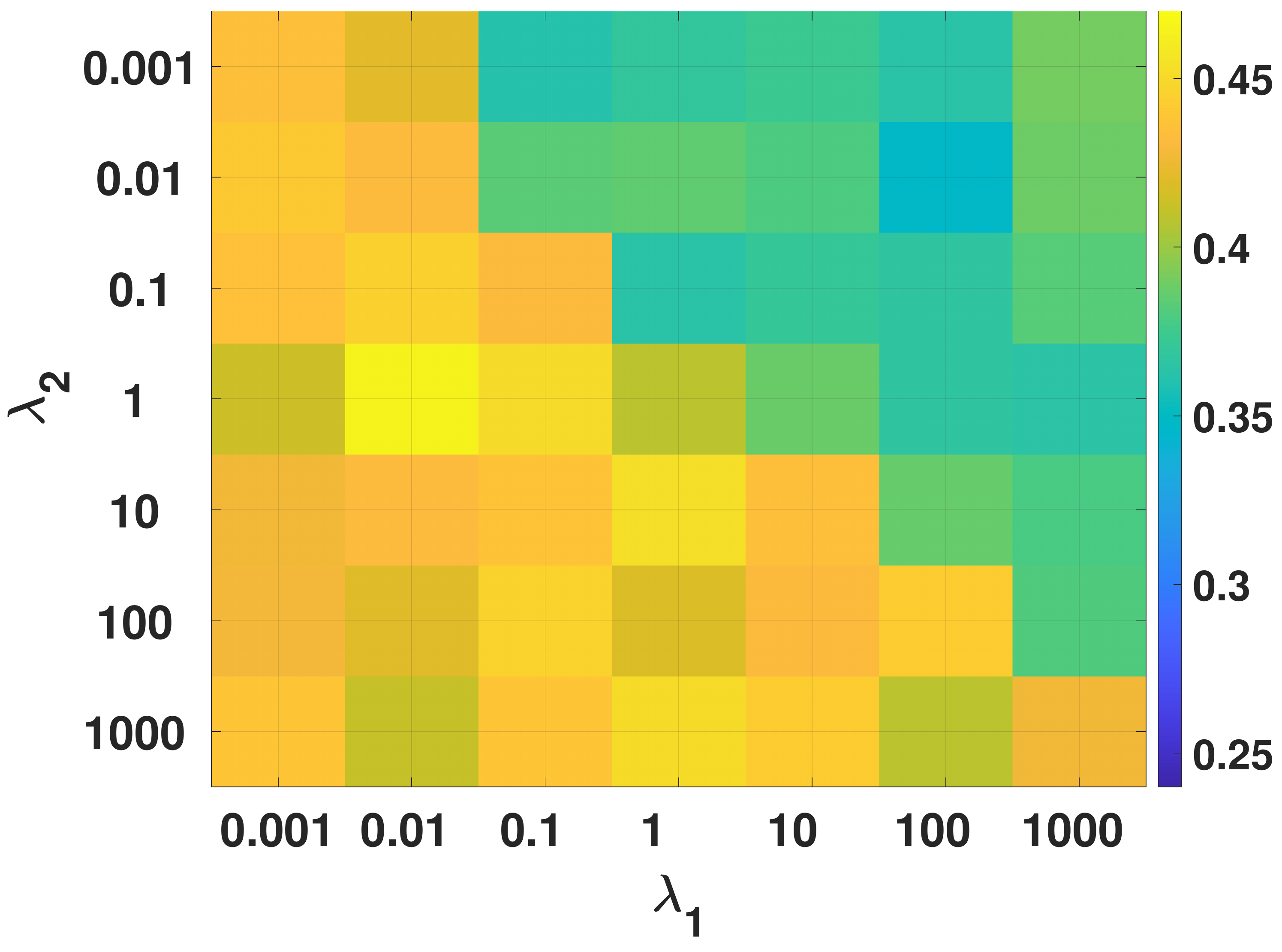}
     }
    \subfigure[$\lambda_3$=0.1 (NMI)]{
	\includegraphics [width=0.2800\columnwidth]{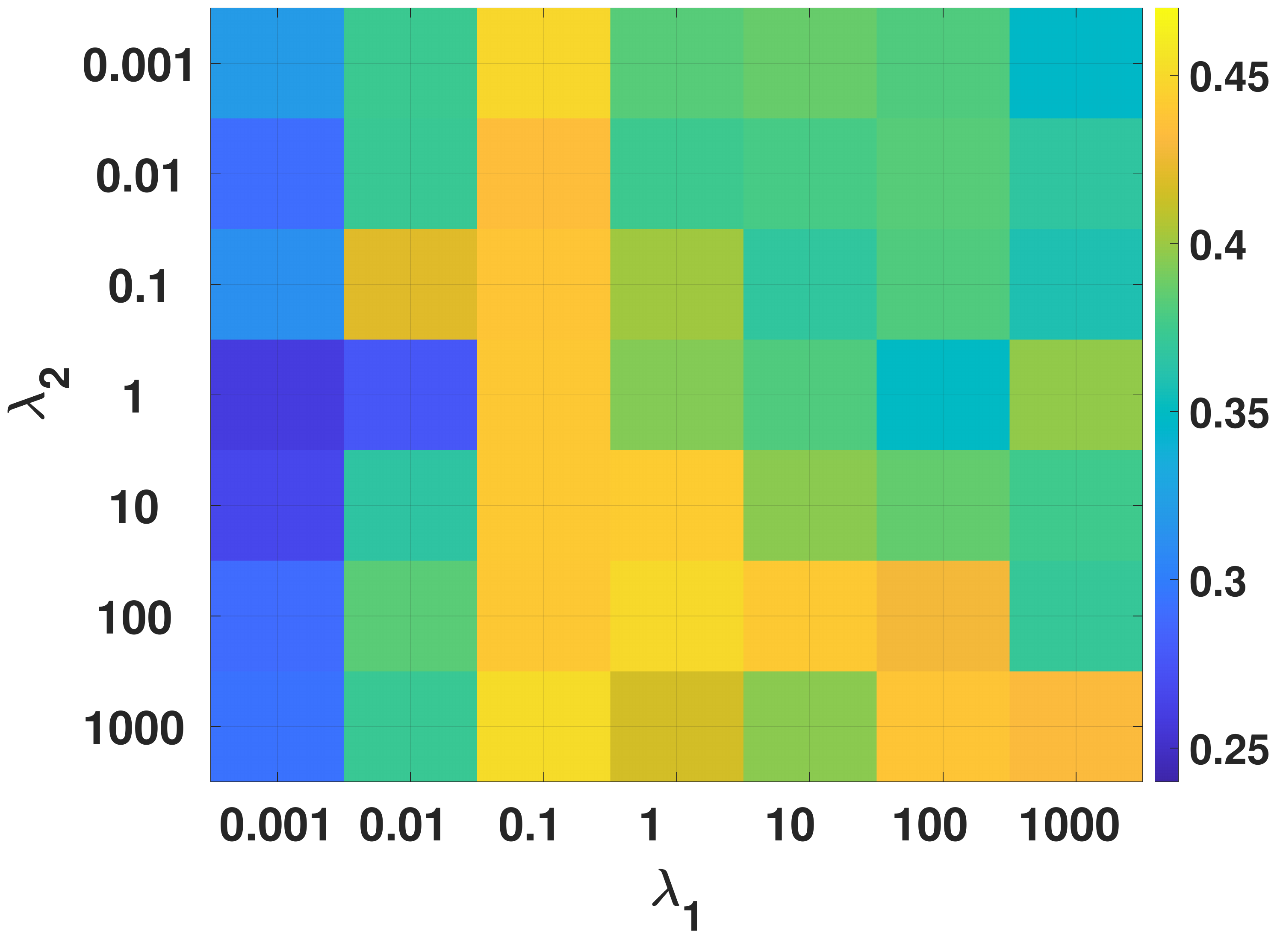}
     }
    \subfigure[$\lambda_3$=10 (NMI)]{
	\includegraphics [width=0.2800\columnwidth]{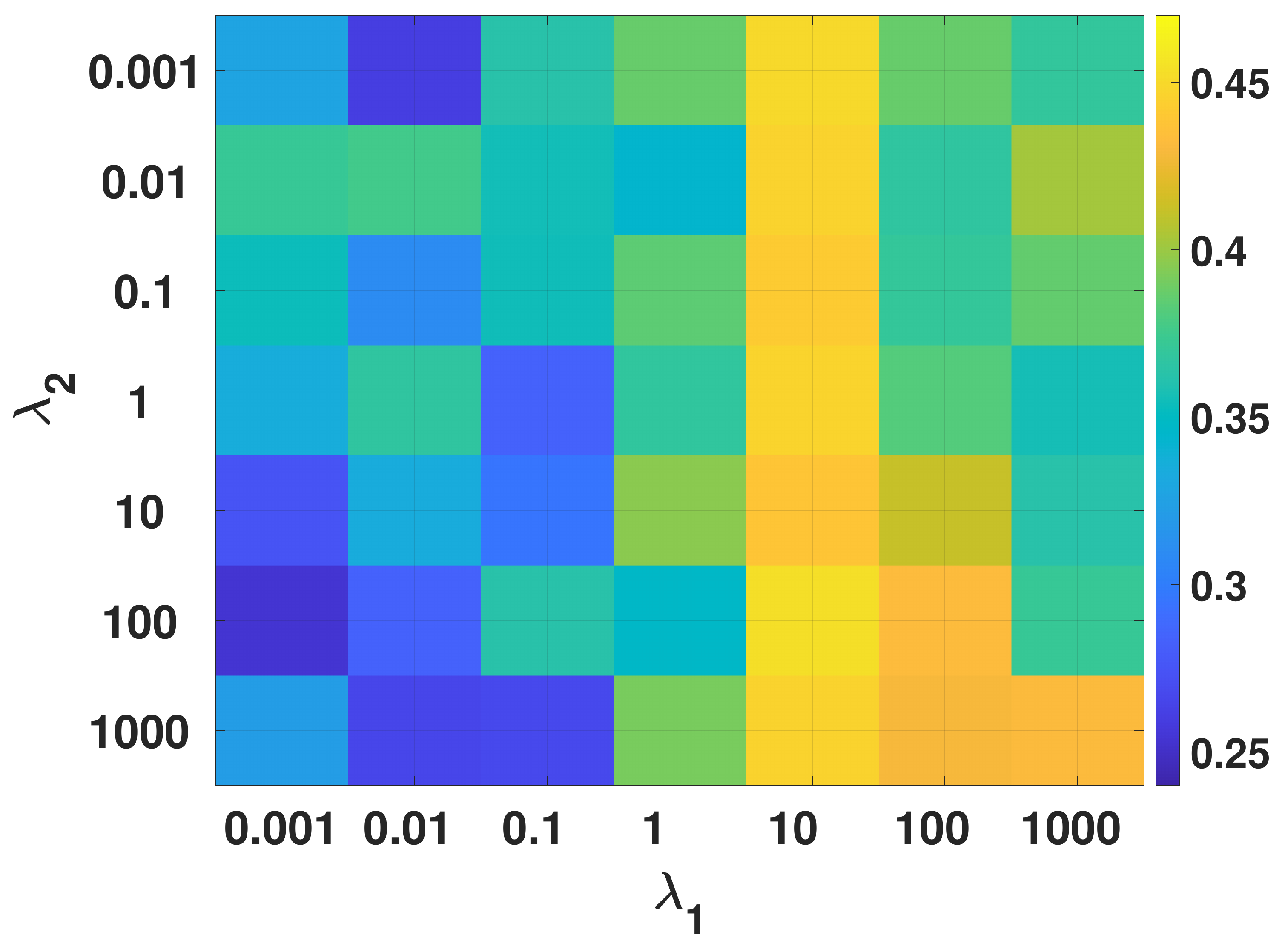}
     }
     \subfigure[$\lambda_3$=1000 (NMI)]{
	\includegraphics [width=0.2800\columnwidth]{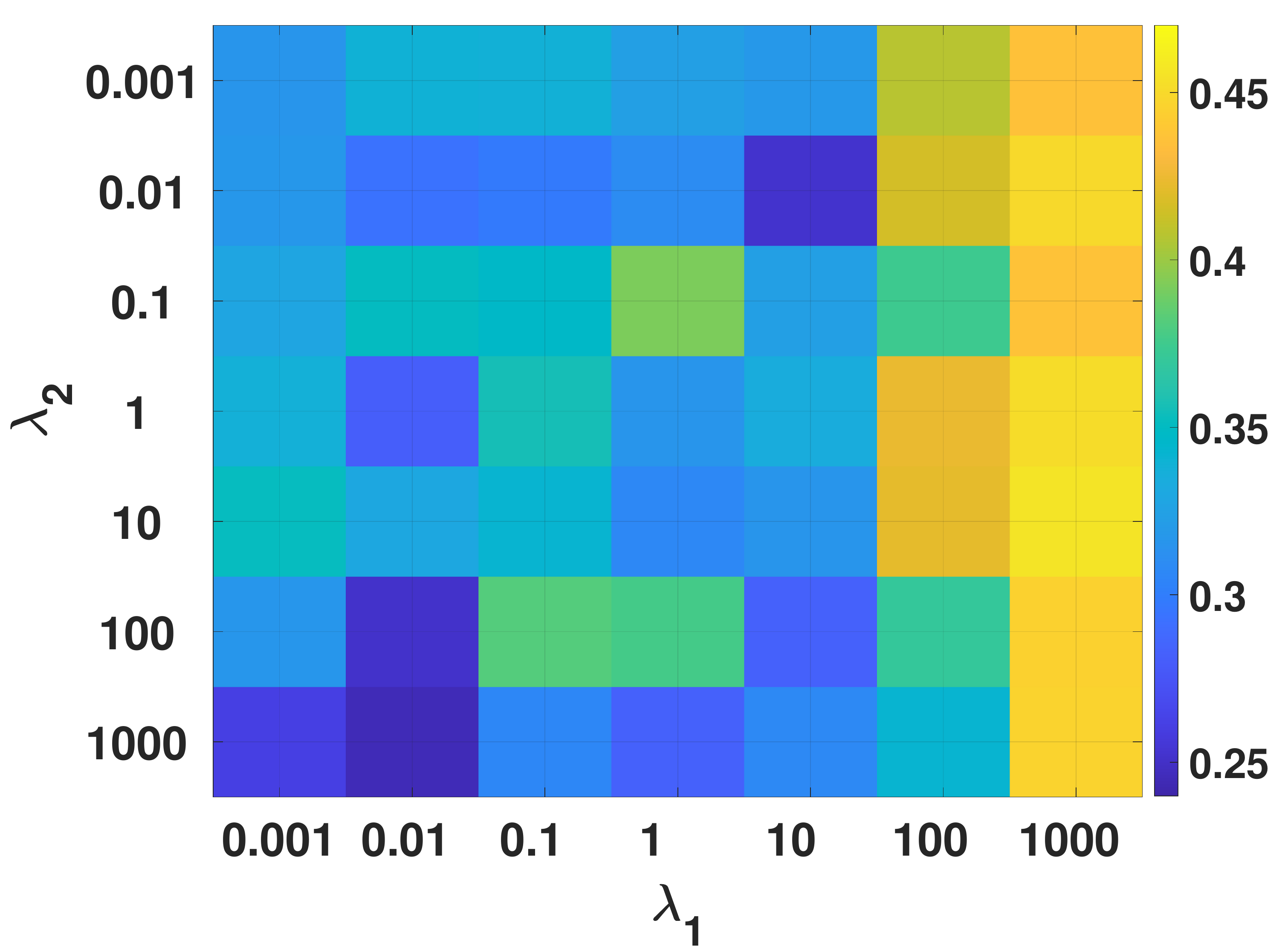}
     }
    \subfigure[$\lambda_3$=0.001 (ARI)]{
	\includegraphics [width=0.2800\columnwidth]{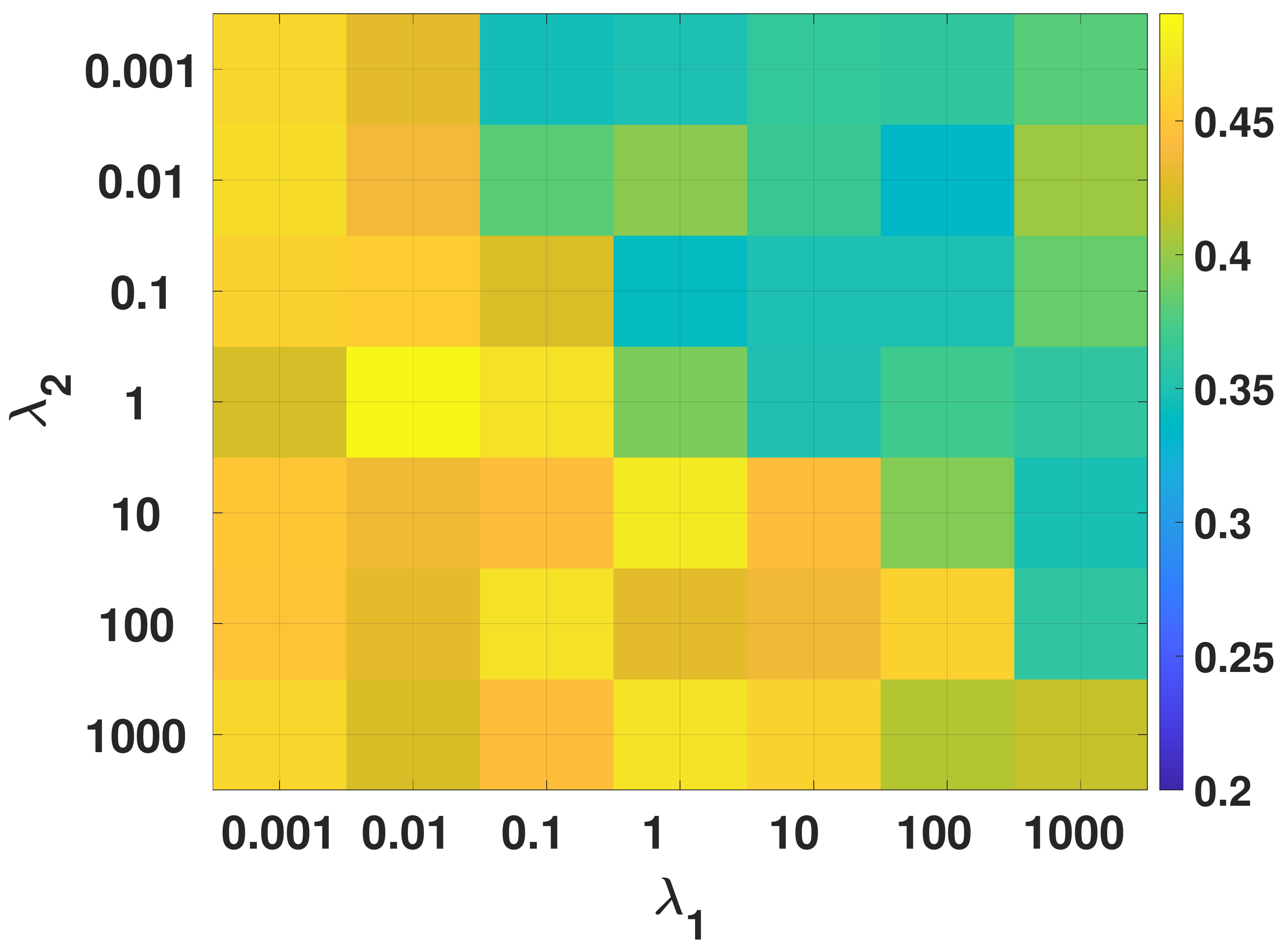}
     }
    \subfigure[$\lambda_3$=0.1 (ARI)]{
	\includegraphics [width=0.2800\columnwidth]{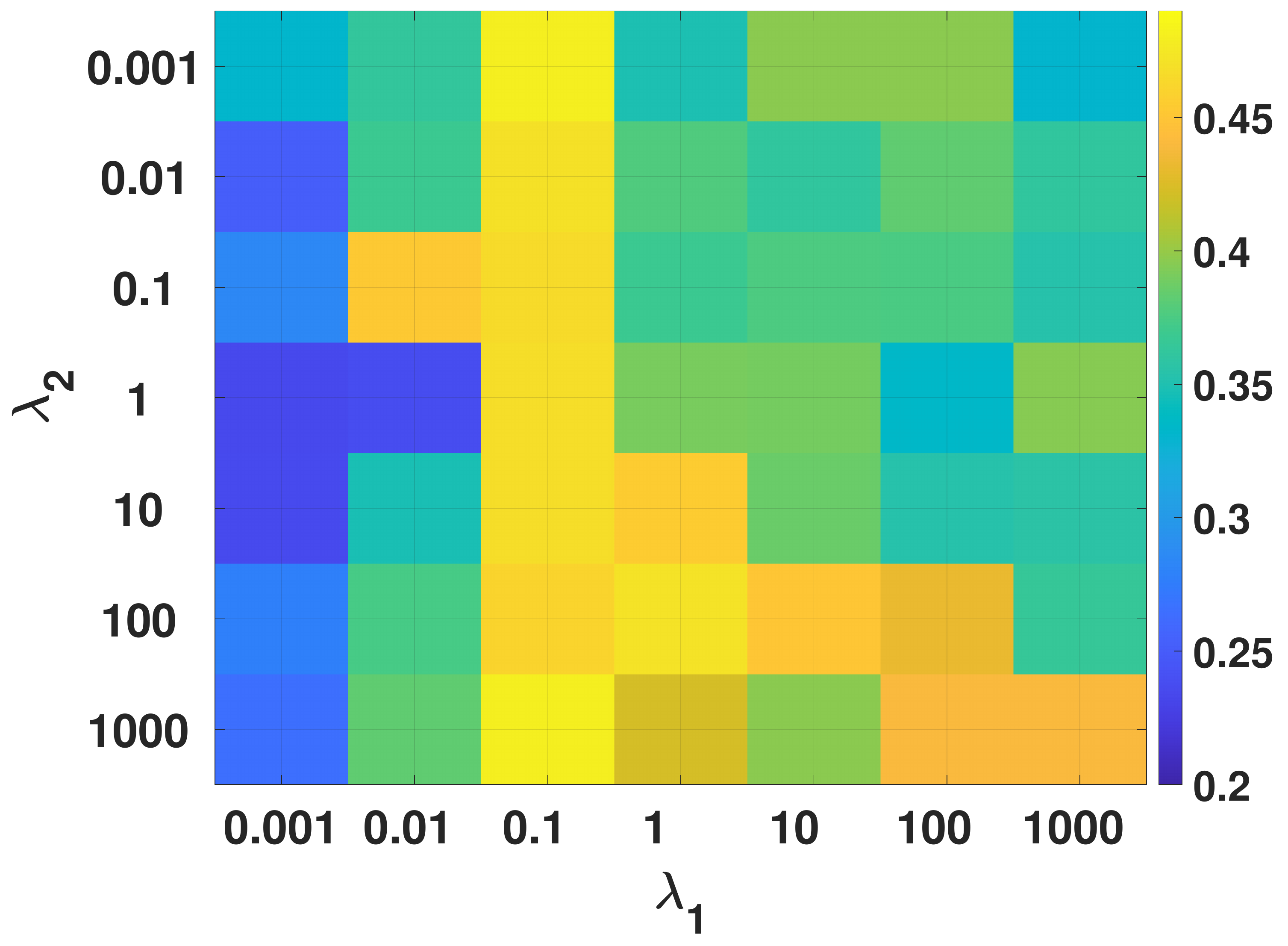}
     }
    \subfigure[$\lambda_3$=10 (ARI)]{
	\includegraphics [width=0.2800\columnwidth]{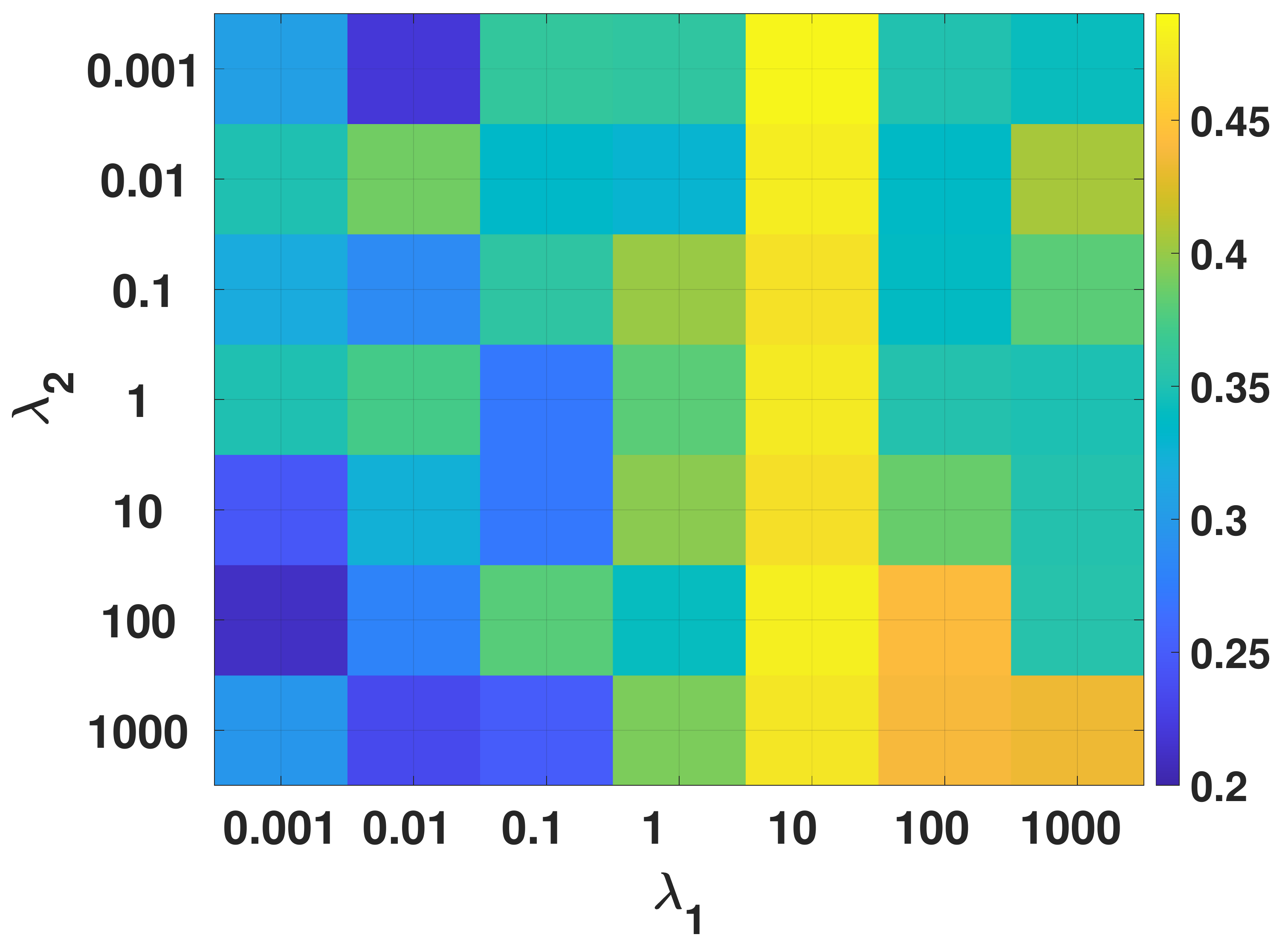}
     }
     \subfigure[$\lambda_3$=1000 (ARI)]{
	\includegraphics [width=0.2800\columnwidth]{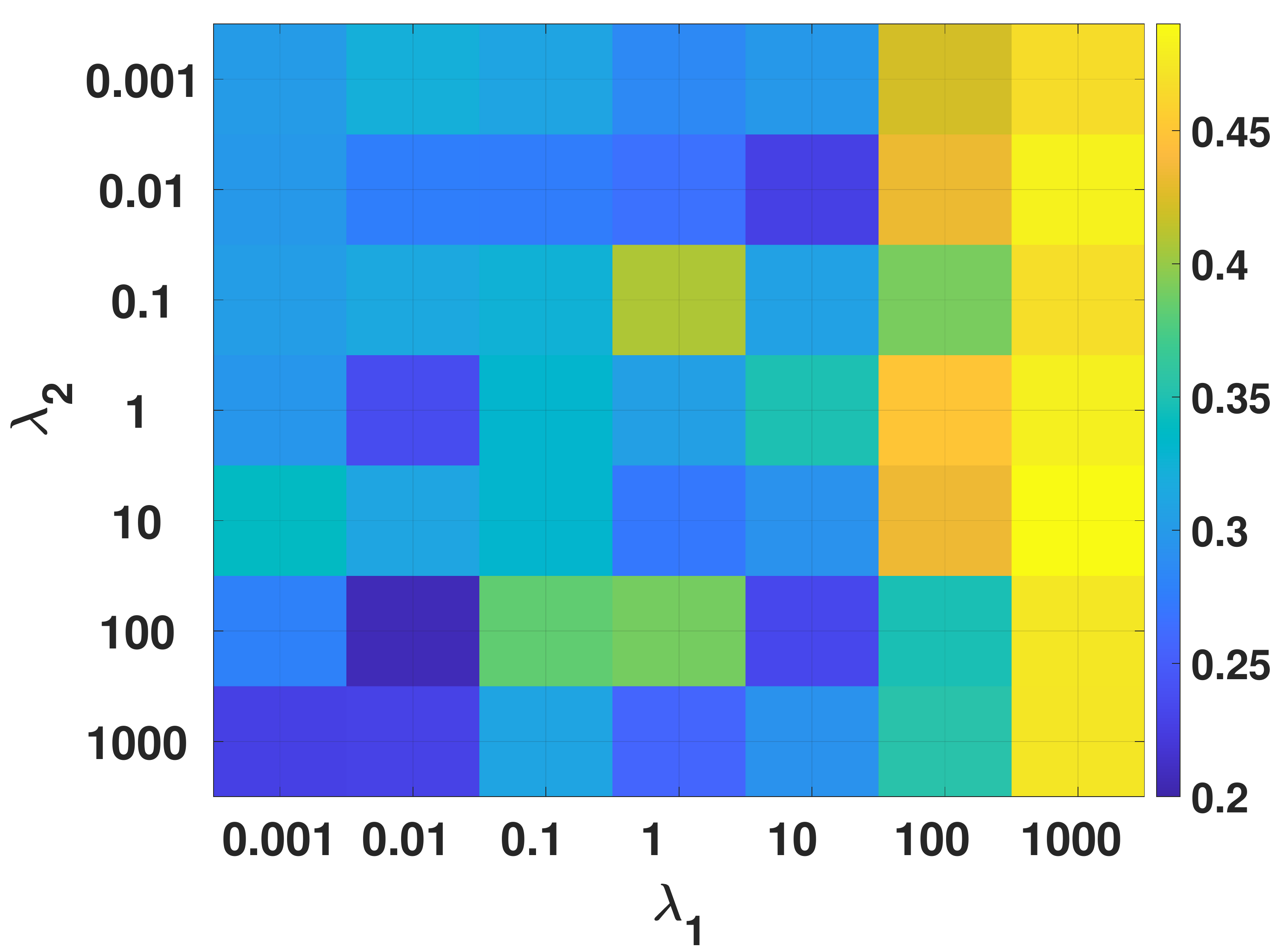}
     }
    \subfigure[$\lambda_3$=0.001 (ACC)]{
	\includegraphics [width=0.2800\columnwidth]{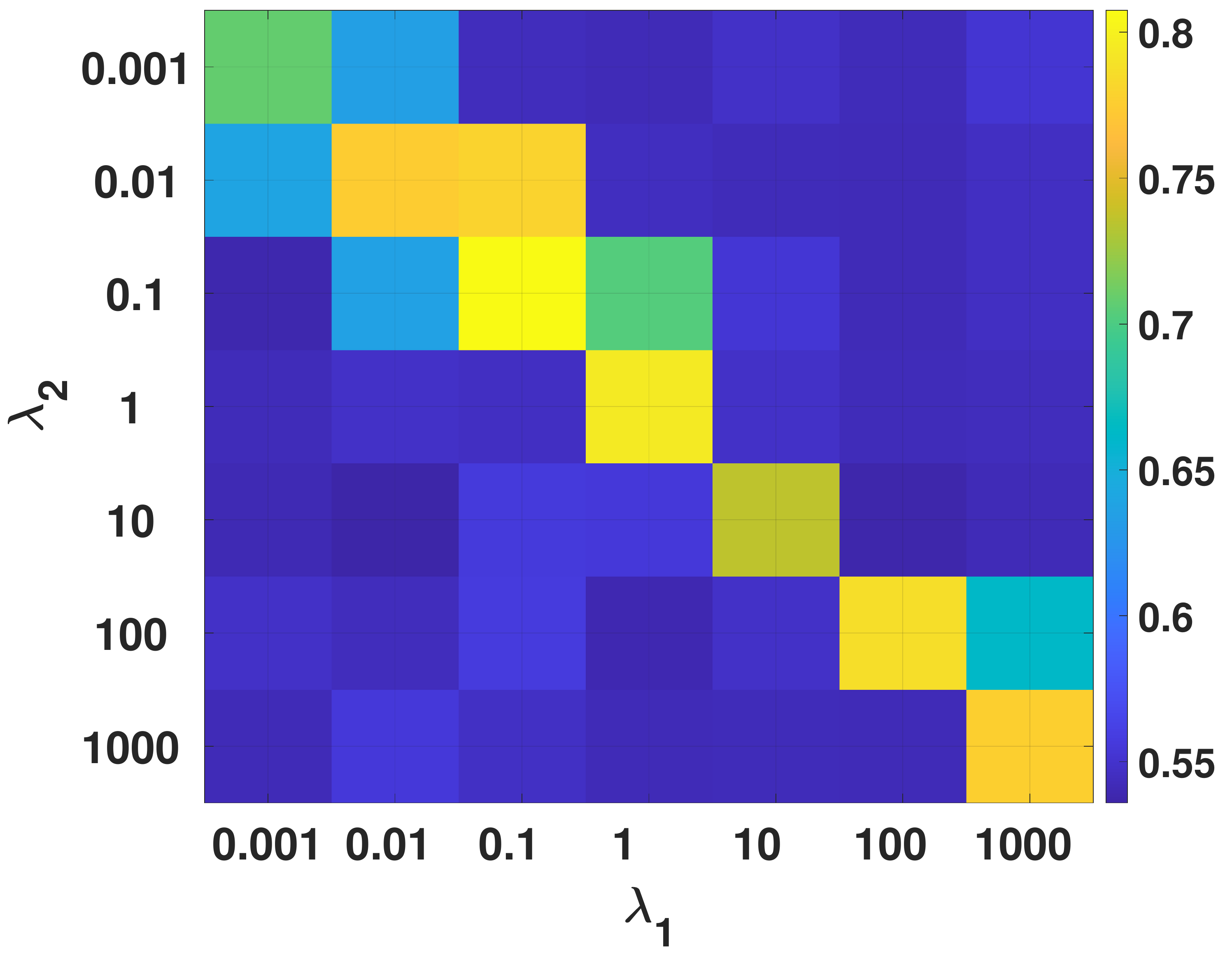}
     }
    \subfigure[$\lambda_3$=0.1 (ACC)]{
	\includegraphics [width=0.2800\columnwidth]{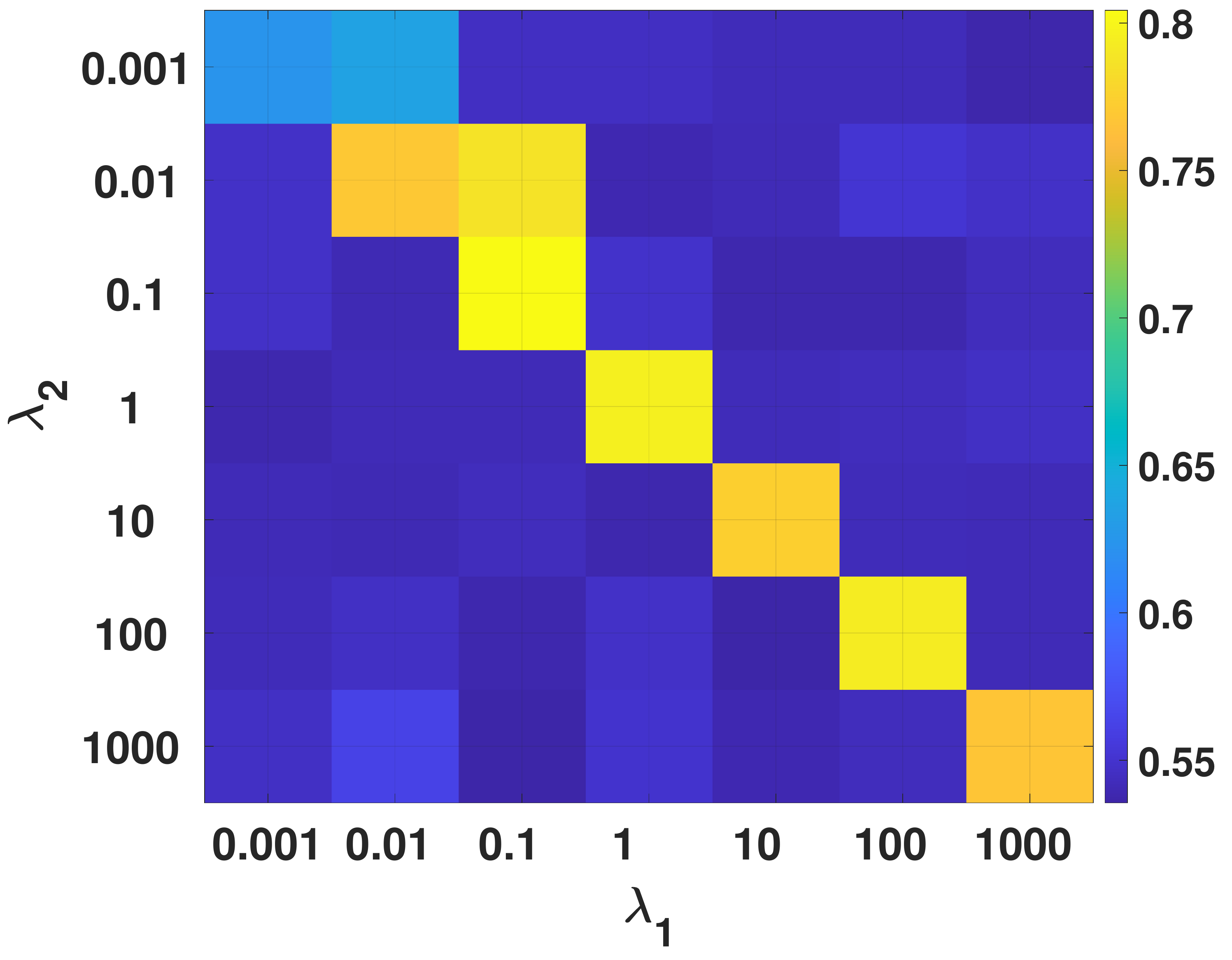}
     }
    \subfigure[$\lambda_3$=10 (ACC)]{
	\includegraphics [width=0.2800\columnwidth]{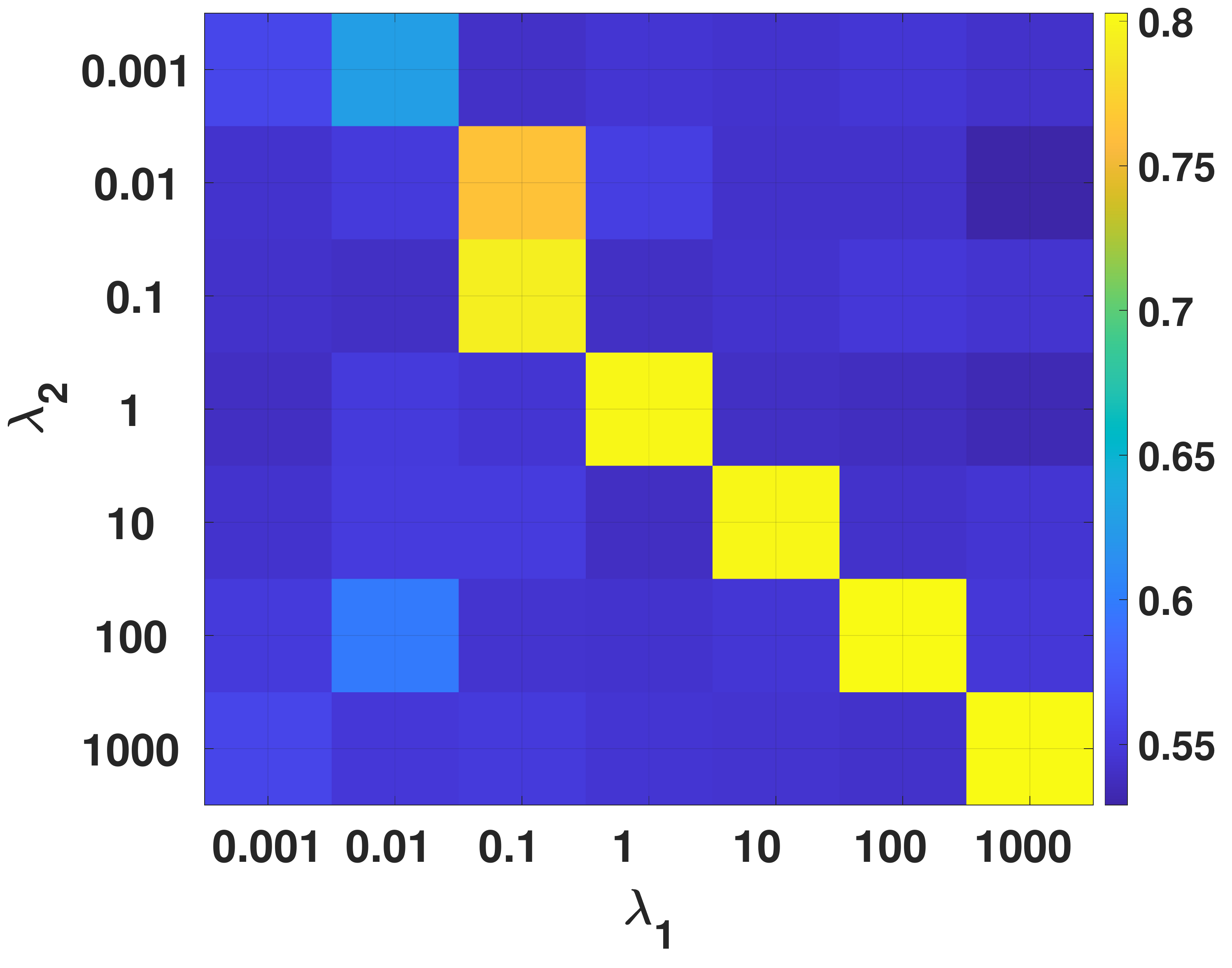}
     }
     \subfigure[$\lambda_3$=1000 (ACC)]{
	\includegraphics [width=0.2800\columnwidth]{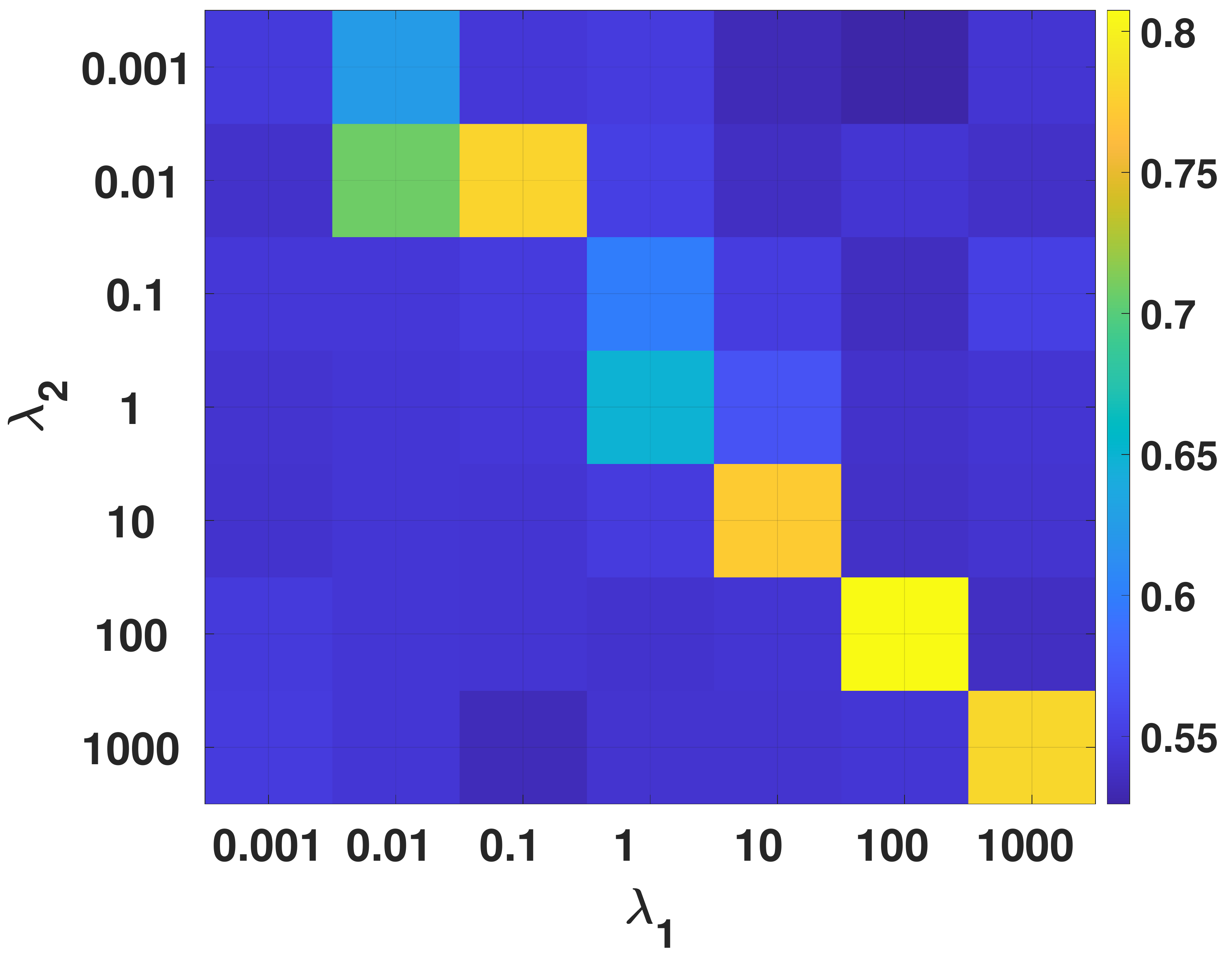}
     }
    \subfigure[$\lambda_3$=0.001 (NMI)]{
	\includegraphics [width=0.2800\columnwidth]{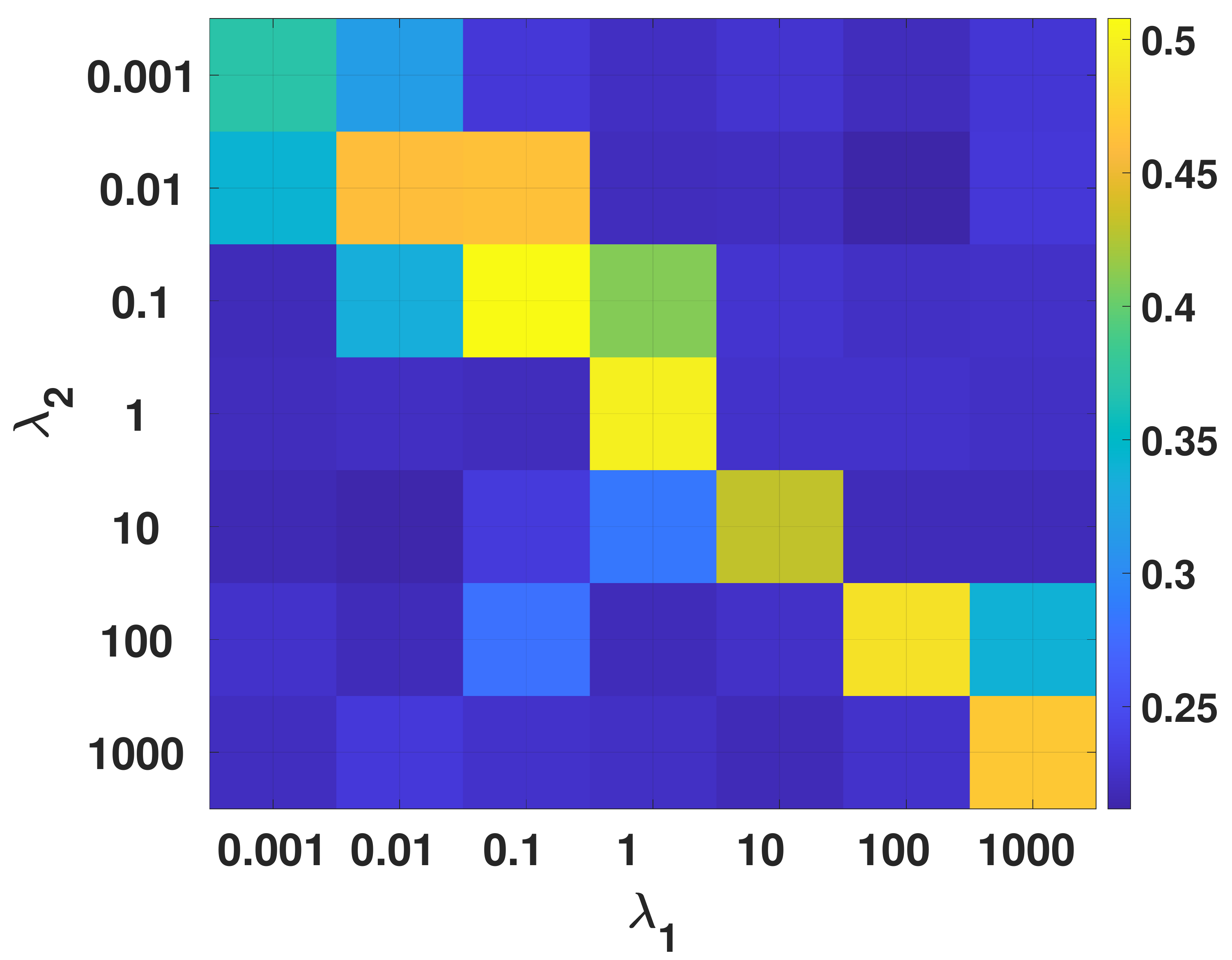}
     }
    \subfigure[$\lambda_3$=0.1 (NMI)]{
	\includegraphics [width=0.2800\columnwidth]{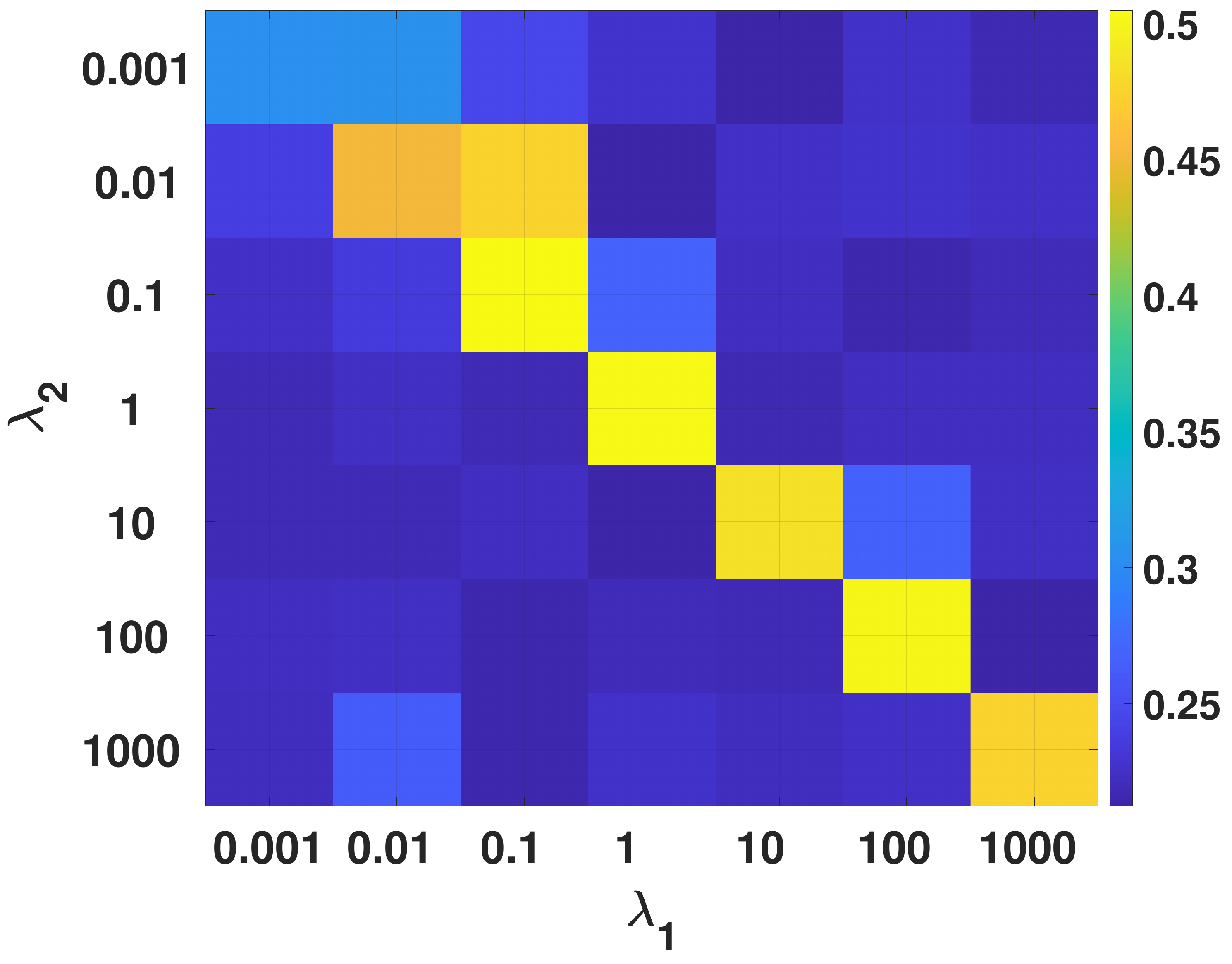}
     }
    \subfigure[$\lambda_3$=10 (NMI)]{
	\includegraphics [width=0.2800\columnwidth]{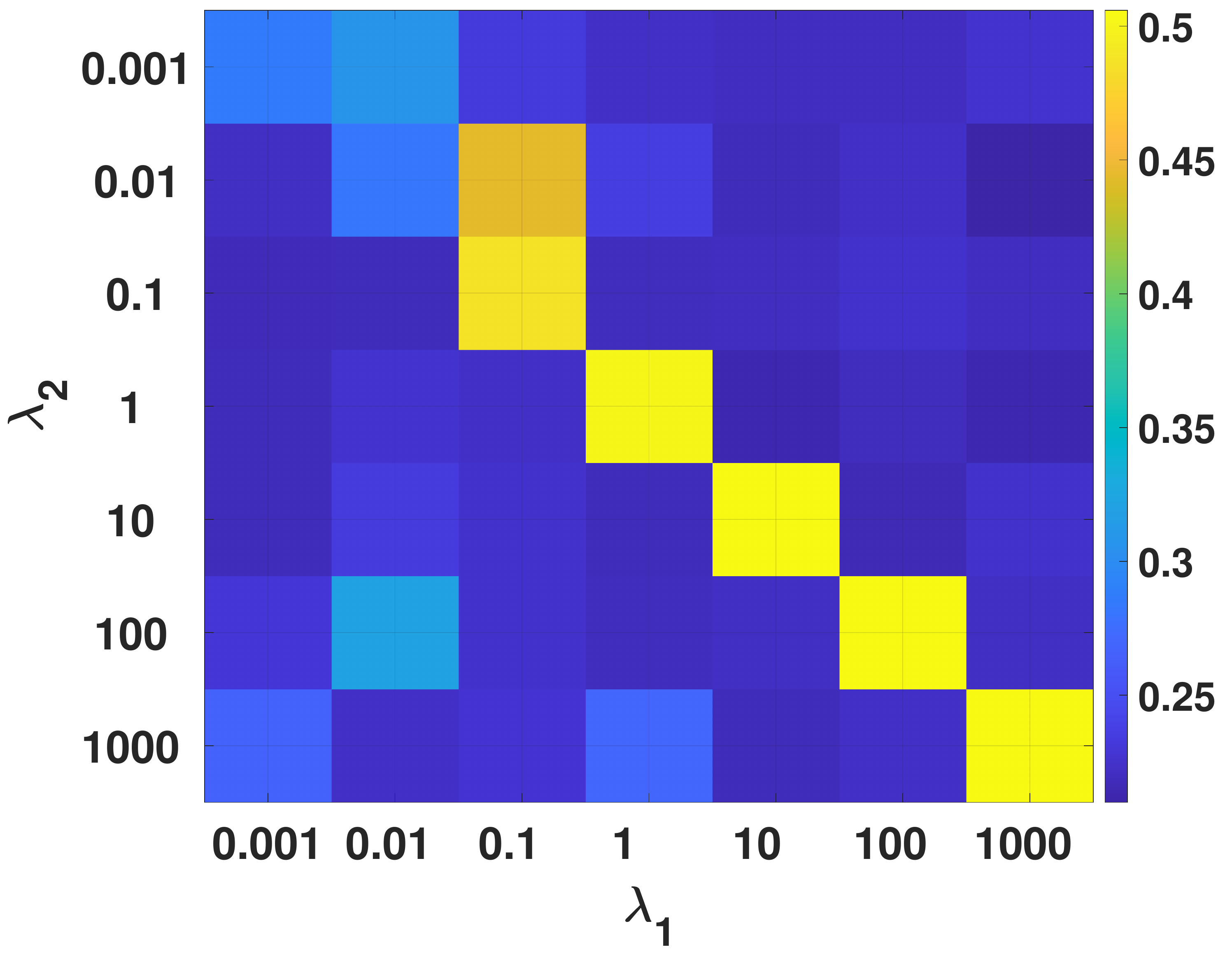}
     }
     \subfigure[$\lambda_3$=1000 (NMI)]{
	\includegraphics [width=0.2800\columnwidth]{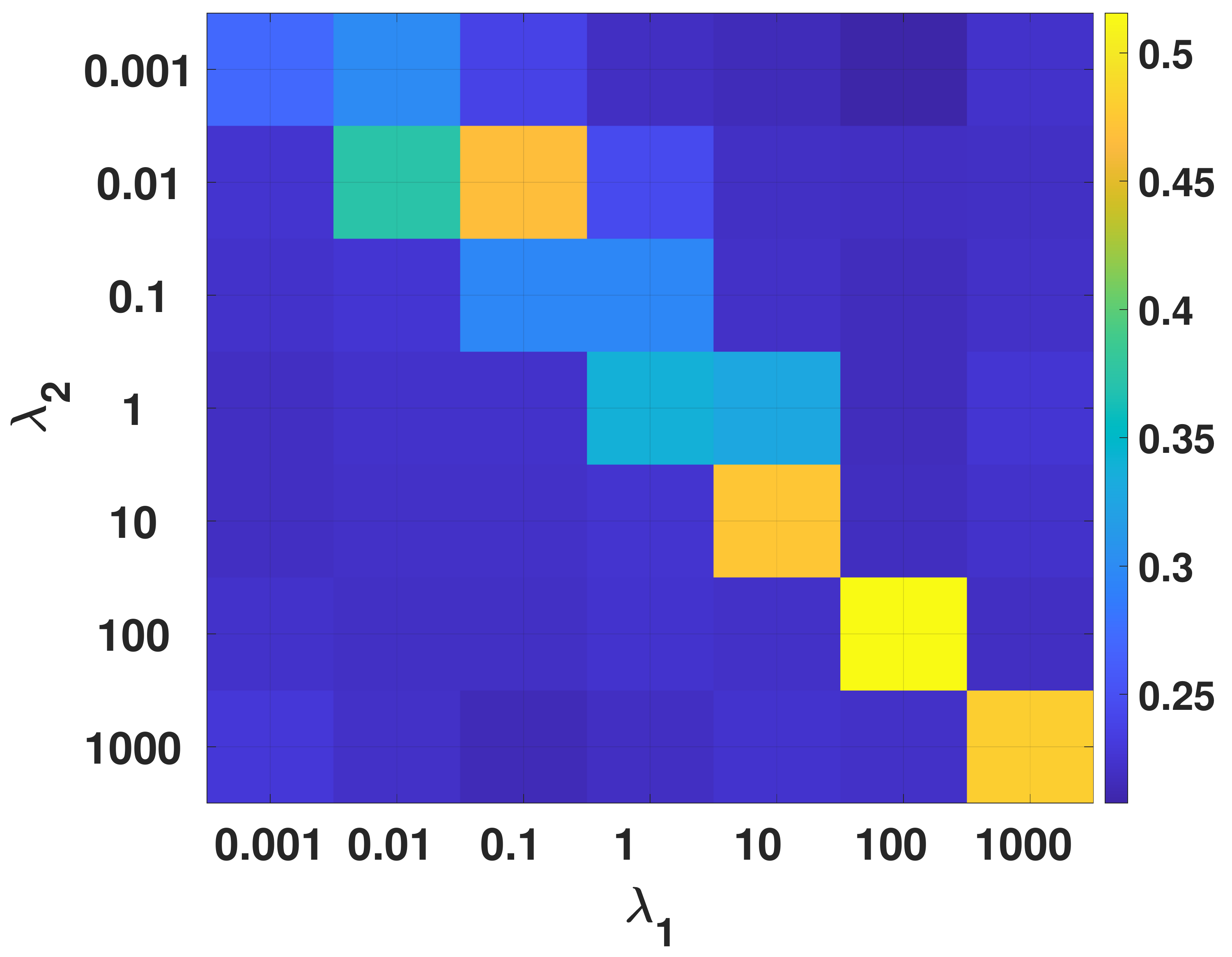}
     }
    \subfigure[$\lambda_3$=0.001 (ARI)]{
	\includegraphics [width=0.2800\columnwidth]{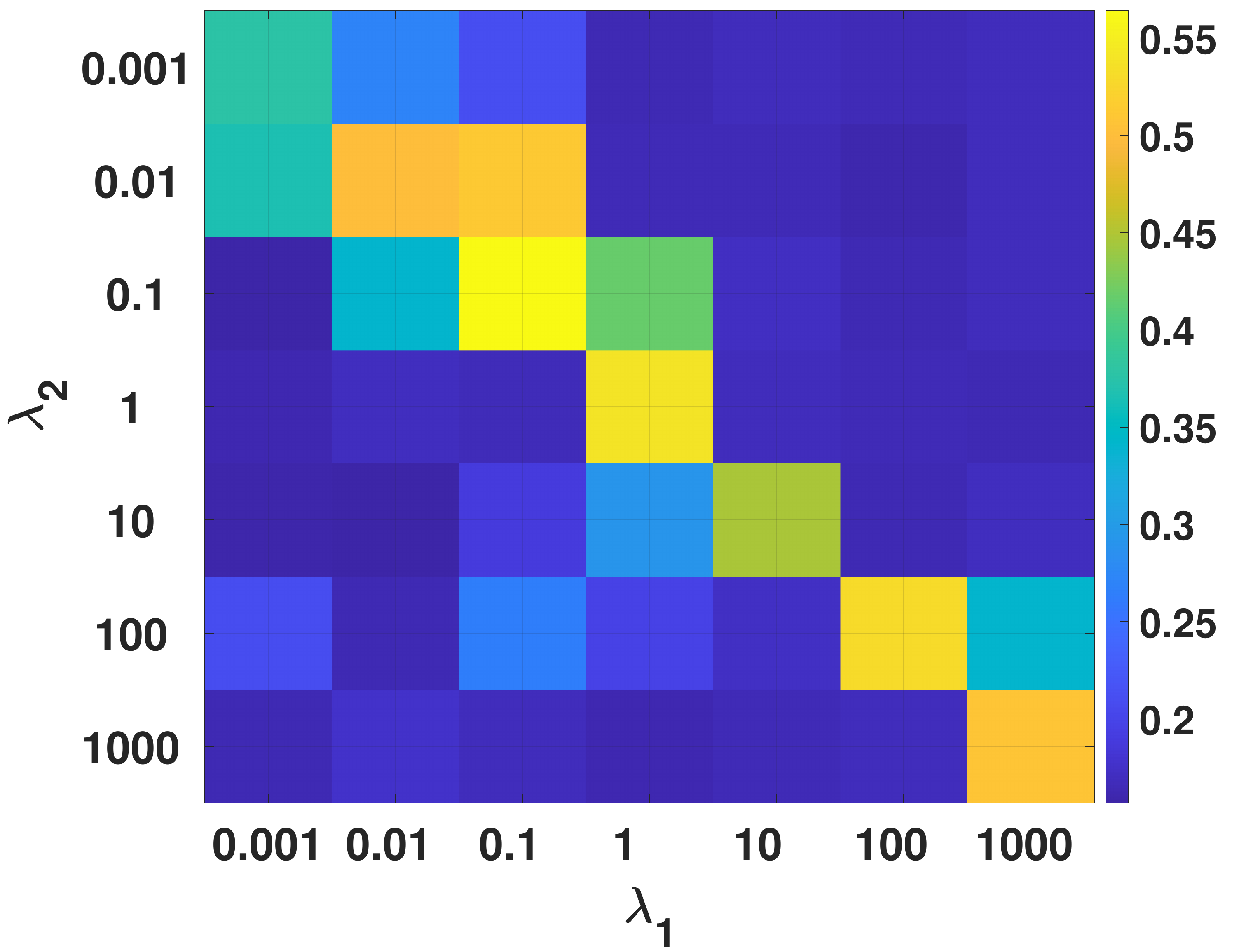}
     }
    \subfigure[$\lambda_3$=0.1 (ARI)]{
	\includegraphics [width=0.2800\columnwidth]{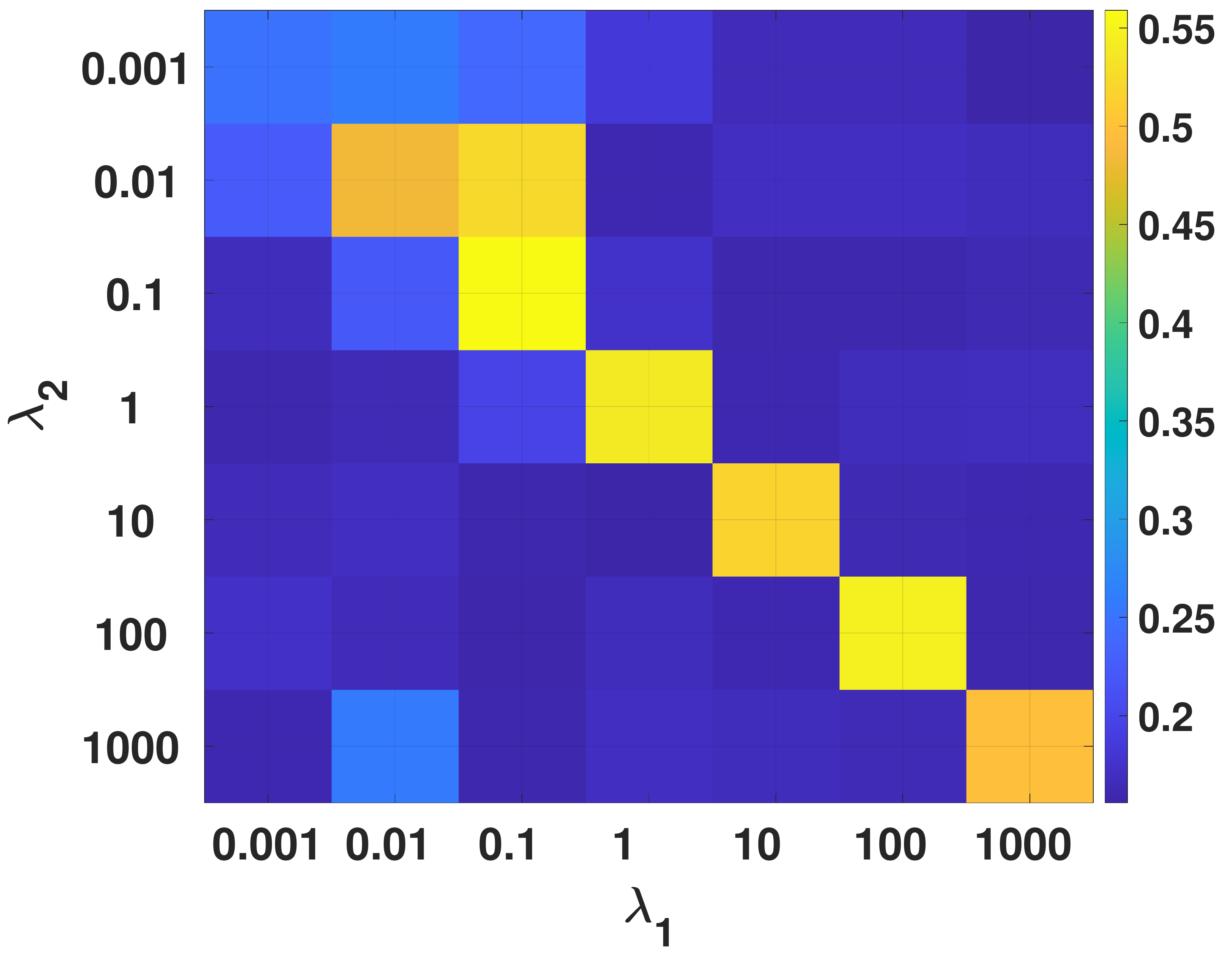}
     }
    \subfigure[$\lambda_3$=10 (ARI)]{
	\includegraphics [width=0.2800\columnwidth]{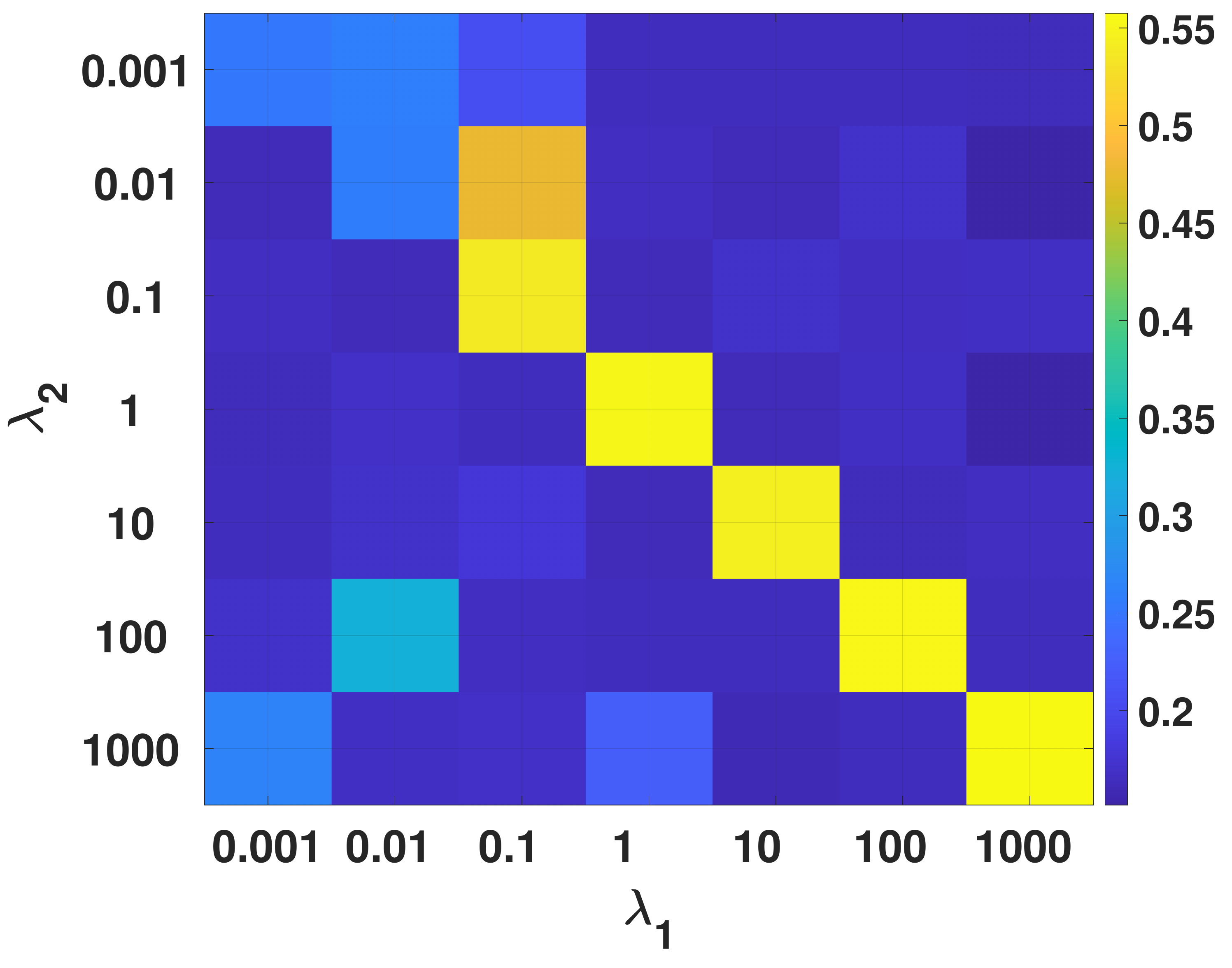}
     }
     \subfigure[$\lambda_3$=1000 (ARI)]{
	\includegraphics [width=0.2800\columnwidth]{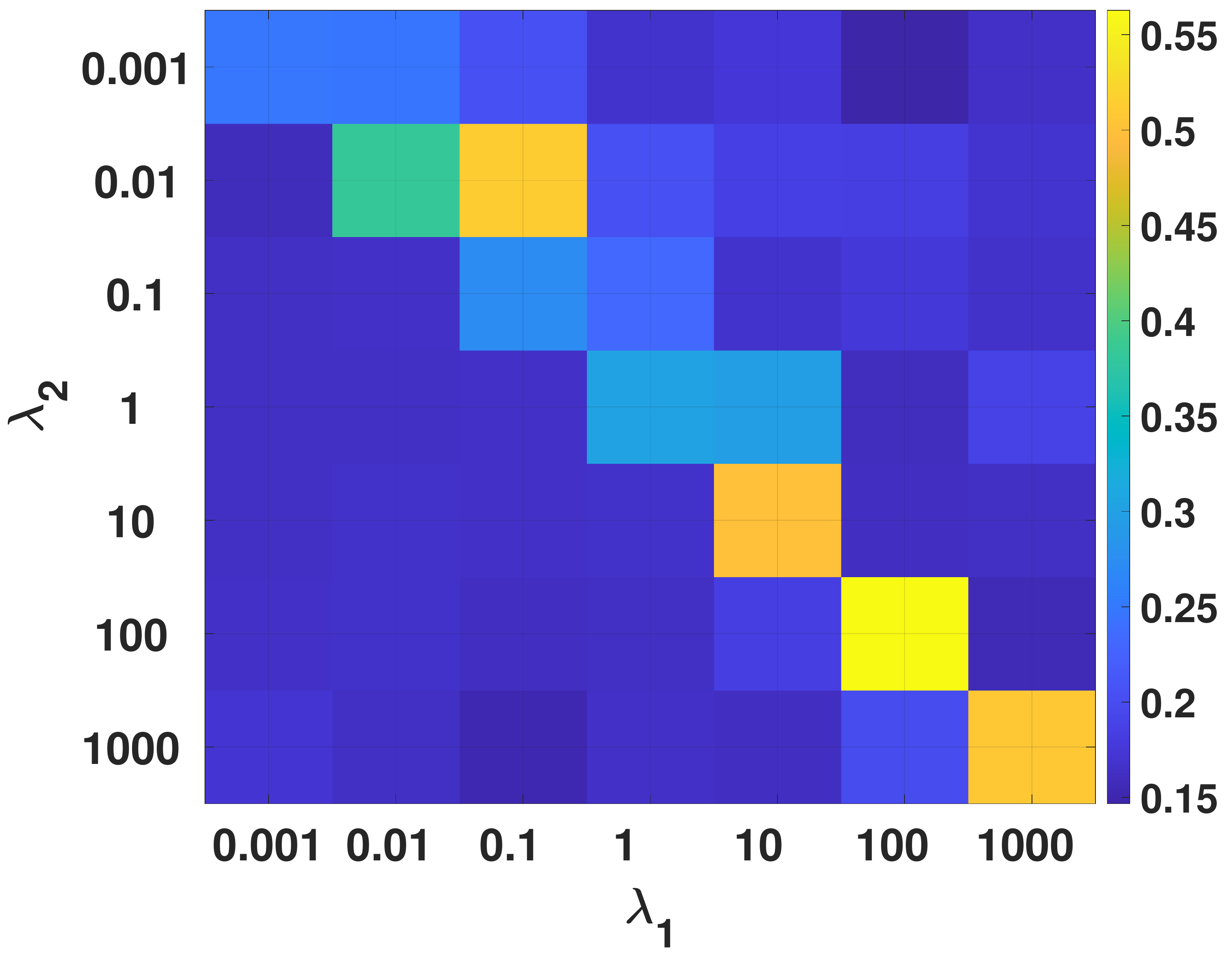}
     }
    \caption{Analysis of different hyperparameters ($\lambda_1$, $\lambda_2$, and $\lambda_3$) with three metrics on CITE (i.e., (a) - (l)) and DBLP (i.e., (m) - (x)). The results of these hyperparameters are visually depicted in a 3D figure, where the color represents the third dimension, i.e., the experimental outcomes.}
	\label{fig: PA-ld123s}
\end{figure*}

\begin{figure}[!t]
	\centering
	\subfigure[DBLP]{
	\includegraphics [width=0.46\columnwidth, height=0.20\columnwidth]{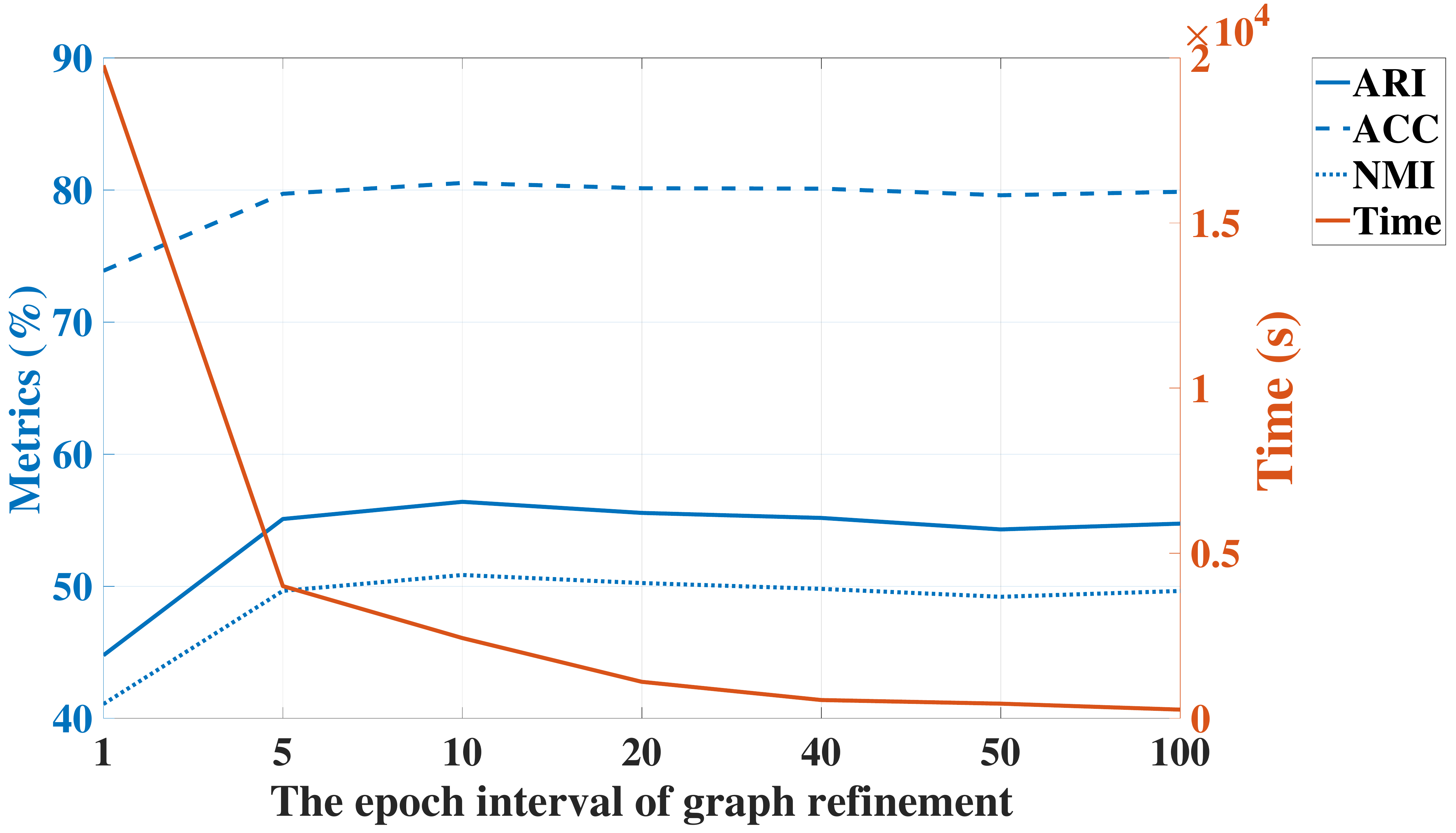}
	}
	\subfigure[ACM]{
	\includegraphics [width=0.46\columnwidth, height=0.20\columnwidth]{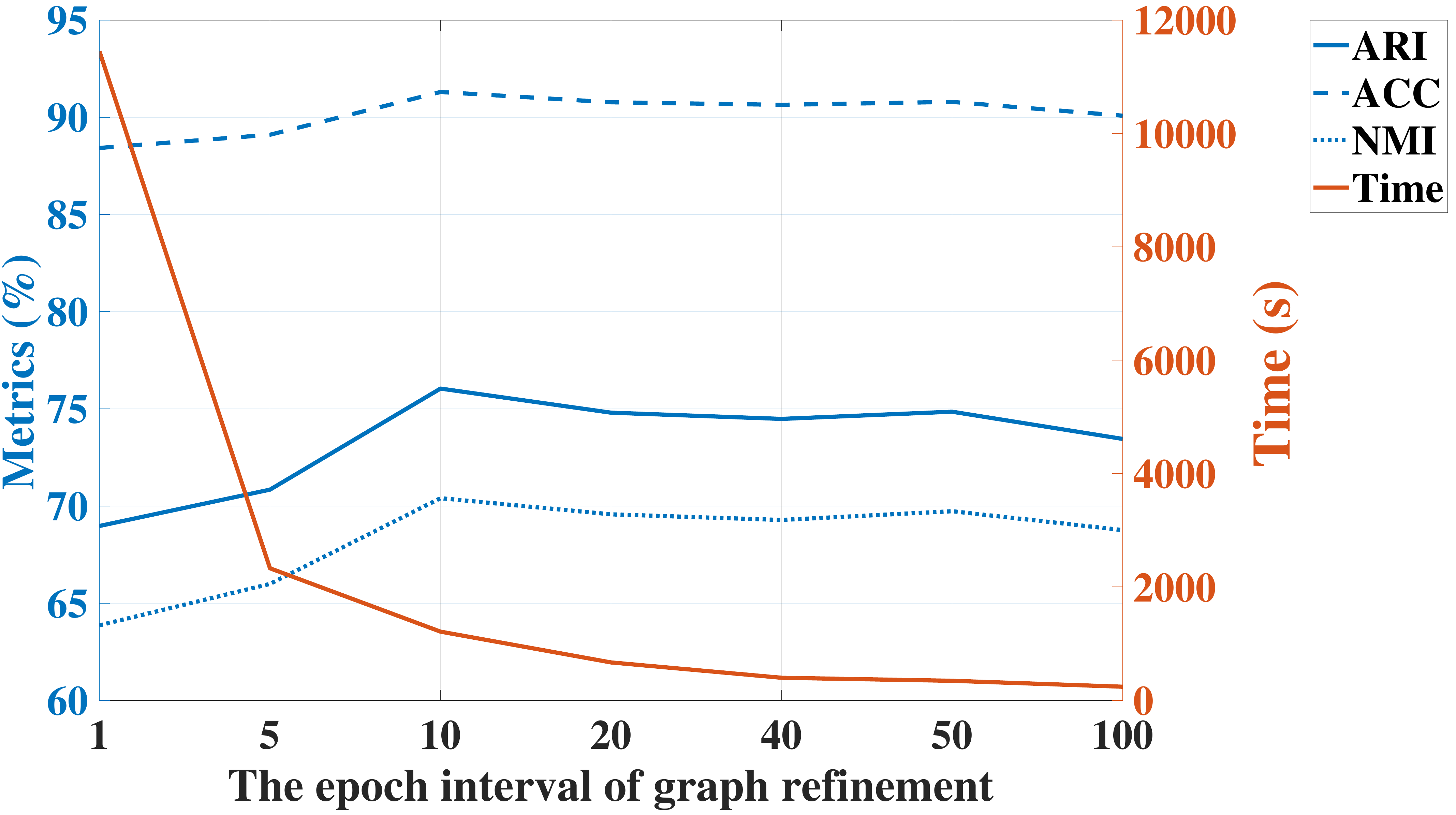}
	}
    \caption{Parameter analysis w.r.t. the epoch interval of graph refinement on (a) DBLP and (b) ACM.}
	\label{fig: PA_ip}
\end{figure}

\begin{figure}[!t]
	\centering
	\subfigure[Overall Loss]{
	\includegraphics [width=0.46\columnwidth]{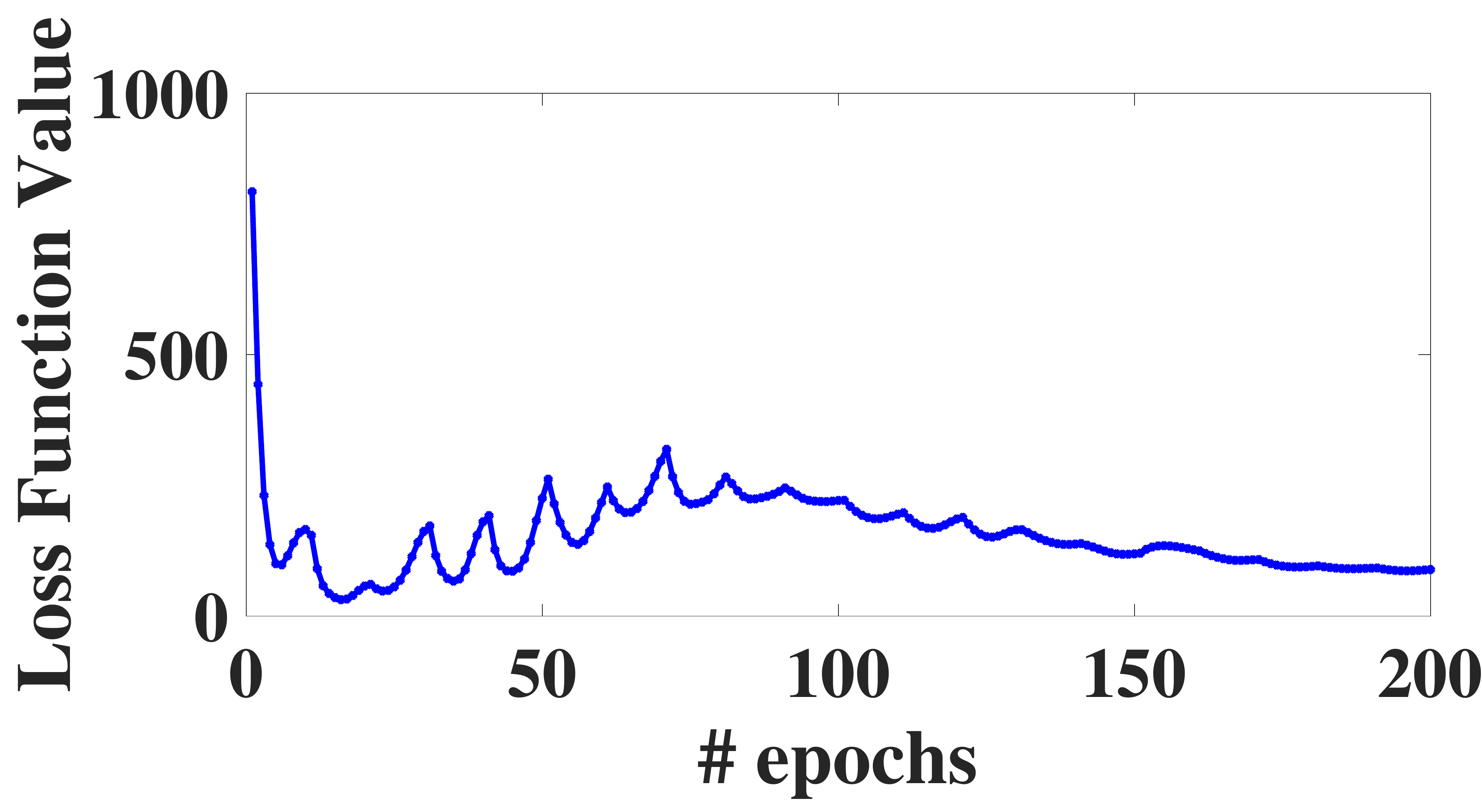}
	}
	\subfigure[Accuracy]{
	\includegraphics [width=0.46\columnwidth]{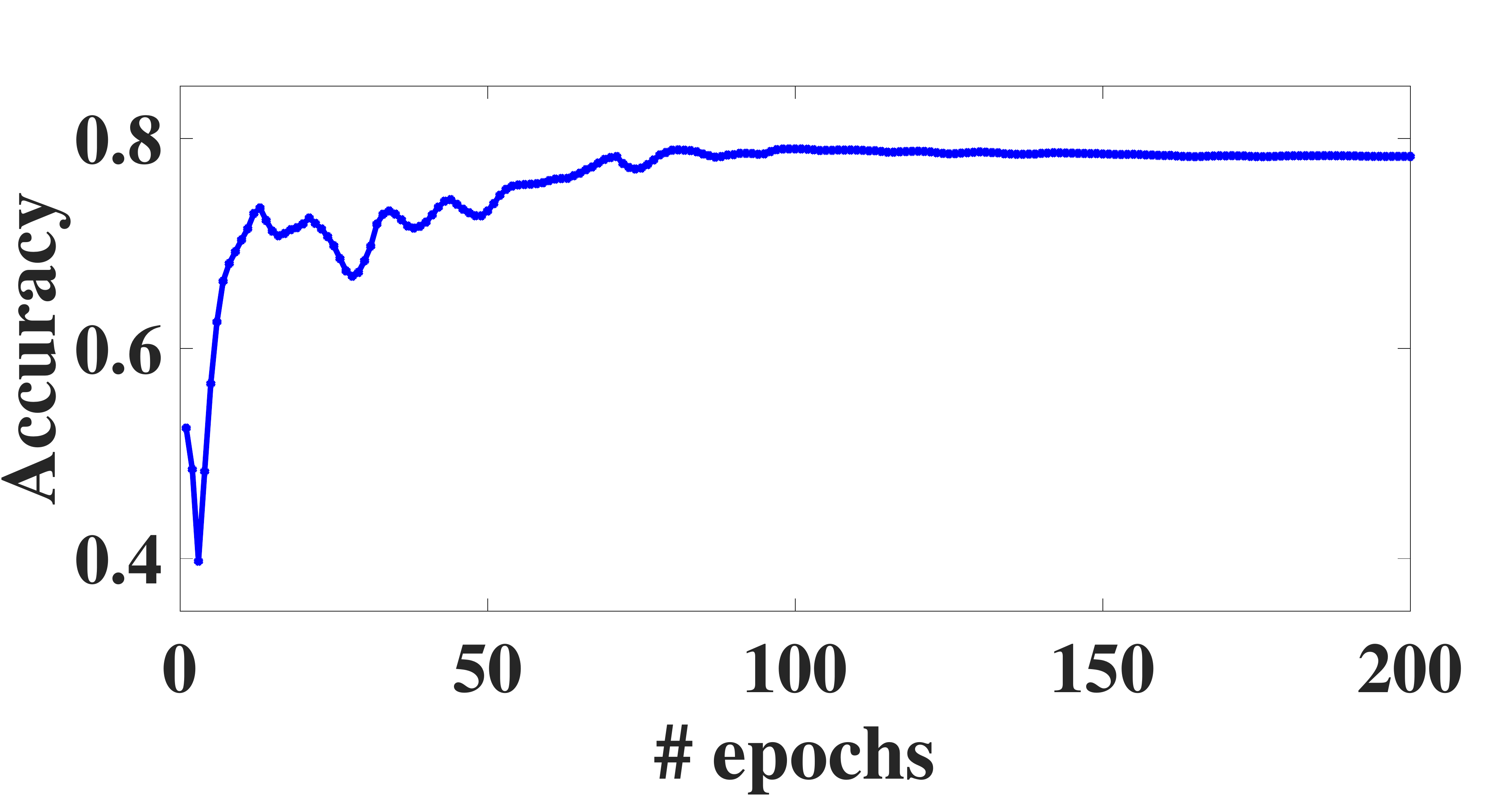}
	}
    \caption{Training stability analysis w.r.t. (a) the loss value and (b) accuracy with different epochs on REUT.}
	\label{fig: loss-convergence}
\end{figure}

\begin{figure*}[!htb]
	\centering
	\subfigure[epoch = 1]{
	\includegraphics [width=0.37000\columnwidth]{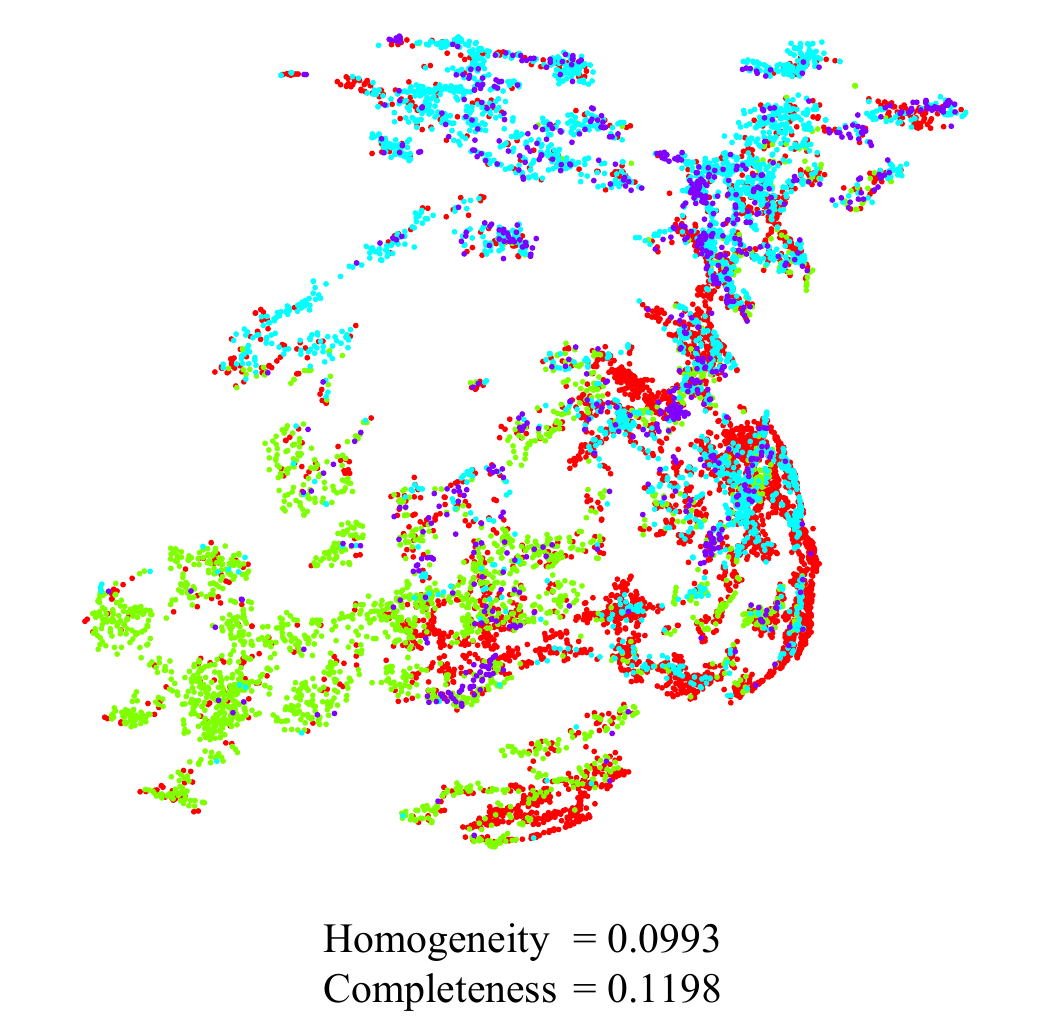}
	}
	\subfigure[epoch = 50]{
	\includegraphics [width=0.37000\columnwidth]{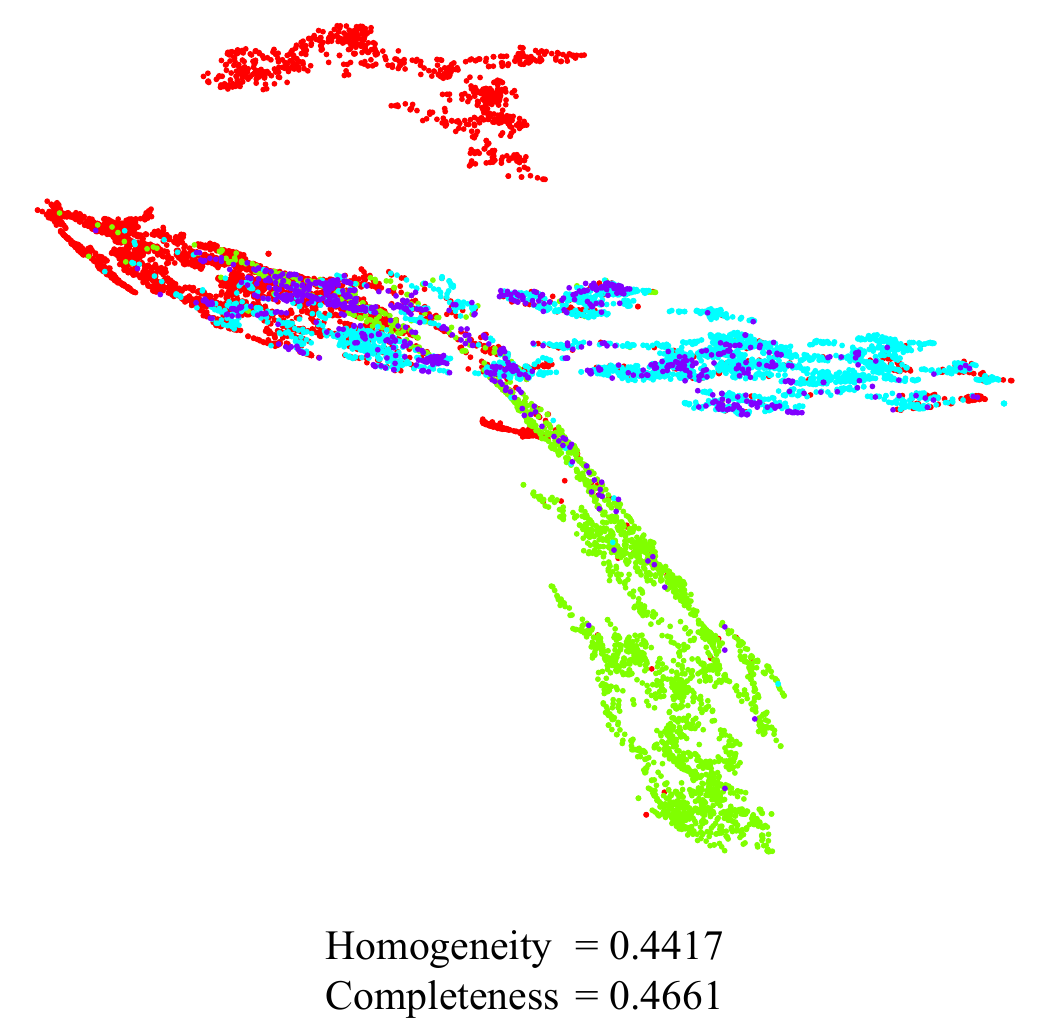}
	}
	\subfigure[epoch = 100]{
	\includegraphics [width=0.37000\columnwidth]{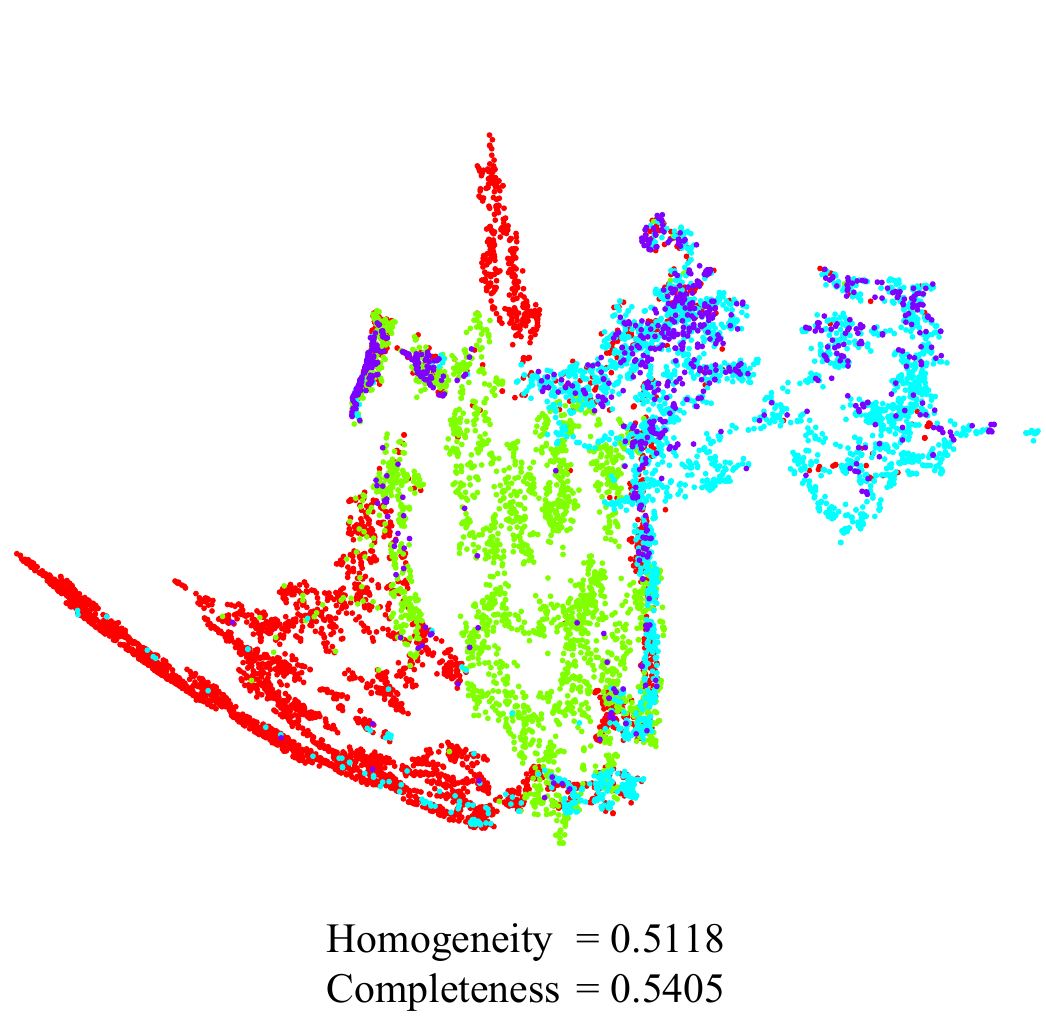}
	}
	\subfigure[epoch = 150]{
	\includegraphics [width=0.37000\columnwidth]{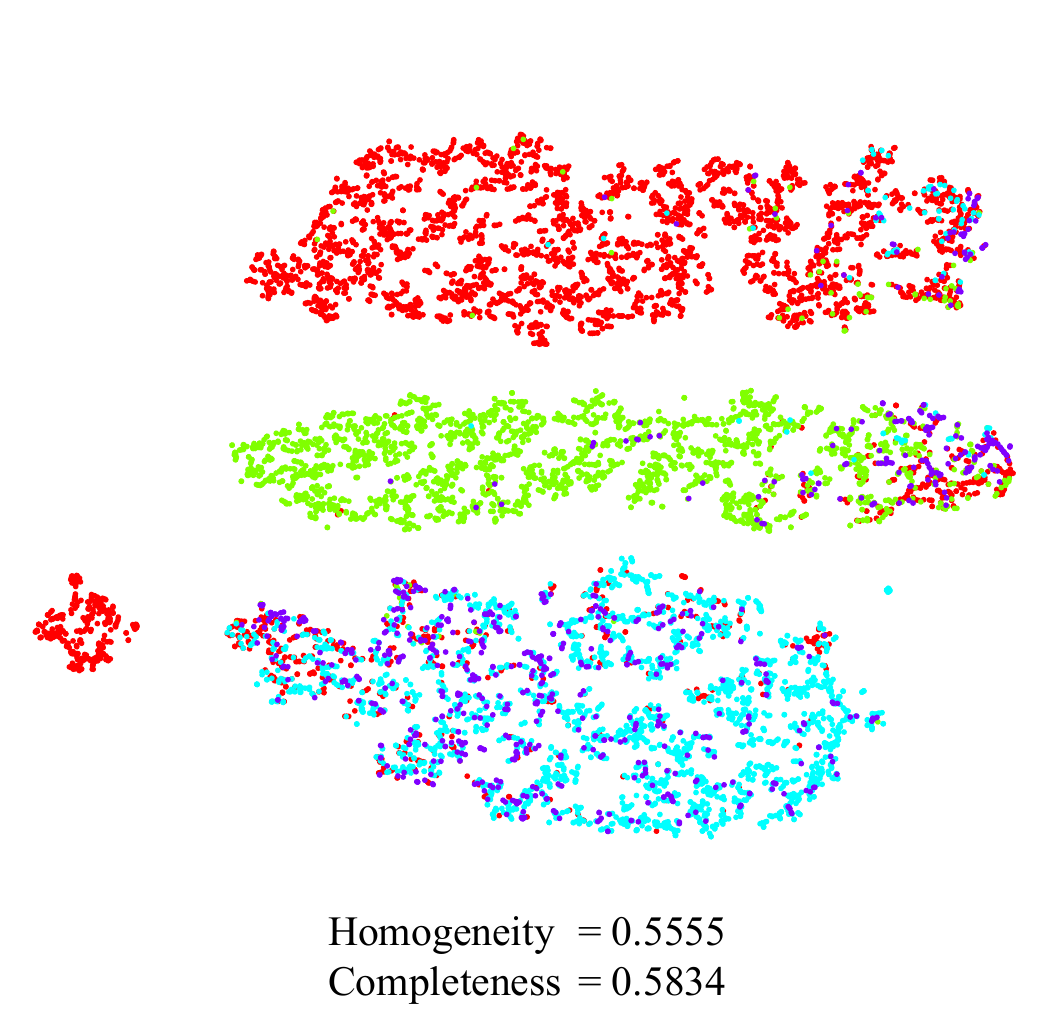}
	}
        \subfigure[epoch = 200]{
	\includegraphics [width=0.37000\columnwidth]{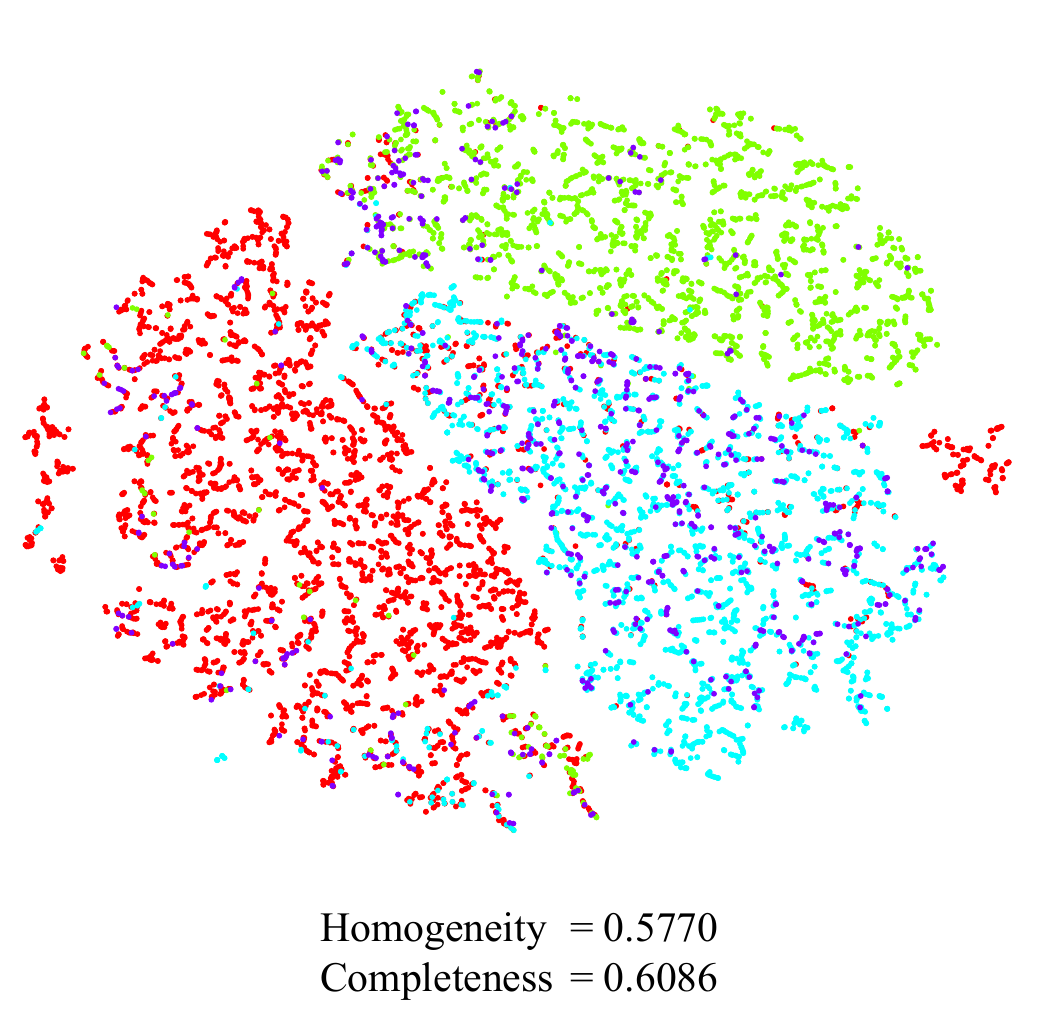}
	}
    \caption{Visualization of the learned representations w.r.t. the epoch number on REUT, where different colors represent different clusters, the homogeneity and completeness measure the visualized quality of the embeddings with a higher score indicating a better representation.}
	\label{fig: tsne_epoch}
\end{figure*}

\begin{table}[]
\centering
\caption{\textcolor{black}{The ablation study of the proposed graph refinement (GR) architecture and the Jeffreys divergence (JD) minimization term, where \XSolidBrush and \Checkmark in each row indicate the non-use or use of the corresponding component, respectively. We highlighted the best results with \textbf{bold}.}}
\label{tab: as_results}
\resizebox{0.58\columnwidth}{!}{
\begin{tabular}{c|cc|cccc}
\hline\hline
                            & GR     & JD     & ARI                   & ACC                   & NMI                   \\
\hline\hline
\multirow{4}{*}{USPS}       & \XSolidBrush & \XSolidBrush & 70.81$\pm$1.03          & 78.02$\pm$2.22          & 79.19$\pm$0.76          \\
                            & \XSolidBrush & \Checkmark & 69.82$\pm$2.78          & 78.83$\pm$2.23          & 77.58$\pm$2.30          \\
                            & \Checkmark & \XSolidBrush & 73.05$\pm$1.87          & 79.56$\pm$1.97          & 79.26$\pm$1.27          \\
                            & \Checkmark & \Checkmark & \textbf{75.61$\pm$1.92} & \textbf{83.41$\pm$4.40} & \textbf{80.94$\pm$0.92} \\
\hline\multirow{4}{*}{STL10}      & \XSolidBrush & \XSolidBrush & 83.23$\pm$0.49          & 91.80$\pm$0.27          & 84.15$\pm$0.72          \\
                            & \XSolidBrush & \Checkmark & 82.71$\pm$1.19          & 91.50$\pm$0.65          & 84.04$\pm$0.33          \\
                            & \Checkmark & \XSolidBrush & 84.70$\pm$0.99          & 92.57$\pm$0.53          & 85.09$\pm$0.85          \\
                            & \Checkmark & \Checkmark & \textbf{85.92$\pm$0.04} & \textbf{93.23$\pm$0.02} & \textbf{86.07$\pm$0.02} \\
\hline\multirow{4}{*}{ImageNet-10} & \XSolidBrush & \XSolidBrush & 75.79$\pm$2.19          & 81.88$\pm$2.08          & 82.51$\pm$1.37          \\
                            & \XSolidBrush & \Checkmark & 75.54$\pm$2.17          & 80.86$\pm$1.97          & \textbf{82.73$\pm$1.39} \\
                            & \Checkmark & \XSolidBrush & 74.69$\pm$1.24          & 79.75$\pm$1.54          & 82.09$\pm$0.77          \\
                            & \Checkmark & \Checkmark & \textbf{77.25$\pm$3.75} & \textbf{86.21$\pm$4.03} & 81.68$\pm$1.99          \\
\hline\multirow{4}{*}{HHAR}       & \XSolidBrush & \XSolidBrush & 73.84$\pm$1.85          & 85.12$\pm$1.14          & 80.70$\pm$1.01          \\
                            & \XSolidBrush & \Checkmark & 75.90$\pm$1.55          & 87.06$\pm$1.30          & 81.13$\pm$1.01          \\
                            & \Checkmark & \XSolidBrush & 77.12$\pm$0.84          & 88.07$\pm$0.62          & 82.40$\pm$0.68          \\
                            & \Checkmark & \Checkmark & \textbf{78.20$\pm$0.33} & \textbf{88.95$\pm$0.21} & \textbf{82.42$\pm$0.36} \\
\hline\multirow{4}{*}{REUT}       & \XSolidBrush & \XSolidBrush & 53.37$\pm$8.33          & 77.22$\pm$4.58          & 50.96$\pm$6.06          \\
                            & \XSolidBrush & \Checkmark & 61.26$\pm$1.34          & 80.71$\pm$0.69          & 57.92$\pm$1.62          \\
                            & \Checkmark & \XSolidBrush & 63.08$\pm$0.48          & 81.84$\pm$0.48          & 59.11$\pm$0.19          \\
                            & \Checkmark & \Checkmark & \textbf{63.53$\pm$0.66} & \textbf{81.90$\pm$0.16} & \textbf{60.32$\pm$0.40} \\
\hline\multirow{4}{*}{ACM}        & \XSolidBrush & \XSolidBrush & 63.87$\pm$5.98          & 86.07$\pm$2.81          & 59.64$\pm$4.46          \\
                            & \XSolidBrush & \Checkmark & 73.58$\pm$1.55          & 90.31$\pm$0.62          & 68.42$\pm$1.30          \\
                            & \Checkmark & \XSolidBrush & 72.60$\pm$2.51          & 89.79$\pm$1.10          & 68.29$\pm$2.07          \\
                            & \Checkmark & \Checkmark & \textbf{76.04$\pm$0.39} & \textbf{91.30$\pm$0.17} & \textbf{70.40$\pm$0.30} \\
\hline\multirow{4}{*}{CITE}       & \XSolidBrush & \XSolidBrush & 43.10$\pm$2.49          & 69.06$\pm$1.65          & 42.79$\pm$1.30          \\
                            & \XSolidBrush & \Checkmark & 47.21$\pm$0.50          & 71.36$\pm$0.32          & 44.09$\pm$0.49          \\
                            & \Checkmark & \XSolidBrush & 44.68$\pm$1.23          & 70.15$\pm$0.71          & 43.60$\pm$0.77          \\
                            & \Checkmark & \Checkmark & \textbf{48.32$\pm$0.57} & \textbf{72.27$\pm$0.37} & \textbf{45.77$\pm$0.48} \\
\hline\multirow{4}{*}{DBLP}       & \XSolidBrush & \XSolidBrush & 51.66$\pm$3.06          & 77.62$\pm$2.05          & 47.04$\pm$1.87          \\
                            & \XSolidBrush & \Checkmark & 54.64$\pm$0.59          & 79.79$\pm$0.38          & 49.57$\pm$0.33          \\
                            & \Checkmark & \XSolidBrush & 48.20$\pm$1.14          & 75.81$\pm$0.73          & 43.67$\pm$0.82          \\
                            & \Checkmark & \Checkmark & \textbf{56.39$\pm$0.93} & \textbf{80.53$\pm$0.54} & \textbf{50.85$\pm$0.60} \\
\hline
\multirow{4}{*}{PubMed}     & \XSolidBrush & \XSolidBrush & 28.47$\pm$1.67          & 67.95$\pm$0.99          & 28.31$\pm$1.64          \\
                            & \XSolidBrush & \Checkmark & 30.39$\pm$1.15          & 68.84$\pm$0.94          & 30.08$\pm$1.25          \\
                            & \Checkmark & \XSolidBrush & 27.20$\pm$1.69          & 66.85$\pm$0.78          & 27.87$\pm$1.84          \\
                            & \Checkmark & \Checkmark & \textbf{34.45$\pm$0.58} & \textbf{71.91$\pm$0.22} & \textbf{32.15$\pm$1.81} \\
\hline\hline
\end{tabular}
}
\end{table}

\subsection{Parameters Analysis} \label{sec: hyper}
\subsubsection{Analysis of hyperparameters}
\textcolor{black}{Our loss function (i.e., Eq. (\ref{eq: DGAC_loss})) has three hyperparameters ($\lambda_1$, $\lambda_2$, and $\lambda_3$) to balance the contributions of multiple loss terms. We conducted the parameter analysis in Figure \ref{fig: PA-ld123s}, where we can find that $\lambda_1$ and $\lambda_2$ (w.r.t. the distribution $\mathbf{P}$) should be larger than $\lambda_3$, reflecting the importance of $\mathbf{P}$ in guiding the network training. In addition, we observe that $\lambda_1$ (w.r.t. $\mathbf{Z}_{a}$) should be larger than $\lambda_2$ because $\mathbf{Z}_{a}$ holds the node attribute and topology structure information by integrating the deep auto-encoder and graph convolution network features, resulting in a stronger guidance ability.}

\subsubsection{Analysis of $\emph{i}_{\it p}$}
 We investigated the effect of the epoch interval of graph refinement $\emph{i}_{\it p}$ on DBLP and ACM in Figure \ref{fig: PA_ip}, where we have the following conclusions.
 \begin{itemize}
     \item Overmuch frequency of graph refinement (e.g., $\emph{i}_{\it p}=1$) leads to high time cost and can lead to performance degradation. The possible reason is that the early training does not learn a useful embedding, and the subsequent graph refinement will cause the model to fall into a suboptimal solution.
     \item Setting $\emph{i}_{\it p}$ to an appropriate value is capable of reducing numerous time consumption with a good clustering performance, e.g., $\emph{i}_{\it p}=10$. Therefore, we empirically set $\emph{i}_{\it p}=10$ in the experiments, which obtains a significant clustering improvement at the cost of acceptable time consumption.
 \end{itemize}

\subsection{Model Stability Analysis} \label{sec: conver}
\textcolor{black}{We conducted experiments on REUT to verify the training stability of the proposed EGRC-Net model. The overall loss and accuracy values concerning different epochs are shown in Figure \ref{fig: loss-convergence}, where we first find that the loss function value gradually becomes stable after 150 iterations, where the proposed network can effectively converge within 200 iterations. 
In addition, we plotted 2D t-distributed stochastic neighbor embedding (t-SNE) visualizations \cite{maaten2008visualizing} of the learned embeddings w.r.t. the epoch number in Figure \ref{fig: tsne_epoch}. Since there is no globally-accepted metric in the existing literature for the t-SNE visualization, we used the homogeneity and completeness scores \cite{wang2018unified,peng2021maximum} to quantitatively measure the visualization performance. A visualization result satisfies the high homogeneity if all of its groups contain only data points that are members of a single class, and that one satisfies the high completeness if all the data points that are members of a given class are elements of the same cluster. These metrics quantitatively measure the visualized quality of the embeddings with a higher score indicating a better representation. From Figures \ref{fig: loss-convergence} (b) and \ref{fig: tsne_epoch}, we observe that, as the epoch increases, the learned embeddings gradually become better, and the corresponding clustering accuracy also becomes higher, demonstrating that the network training is stable and effective.}

\begin{figure*}[]
	\centering
	\subfigure[SDCN]{
	\includegraphics [width=0.37\columnwidth]{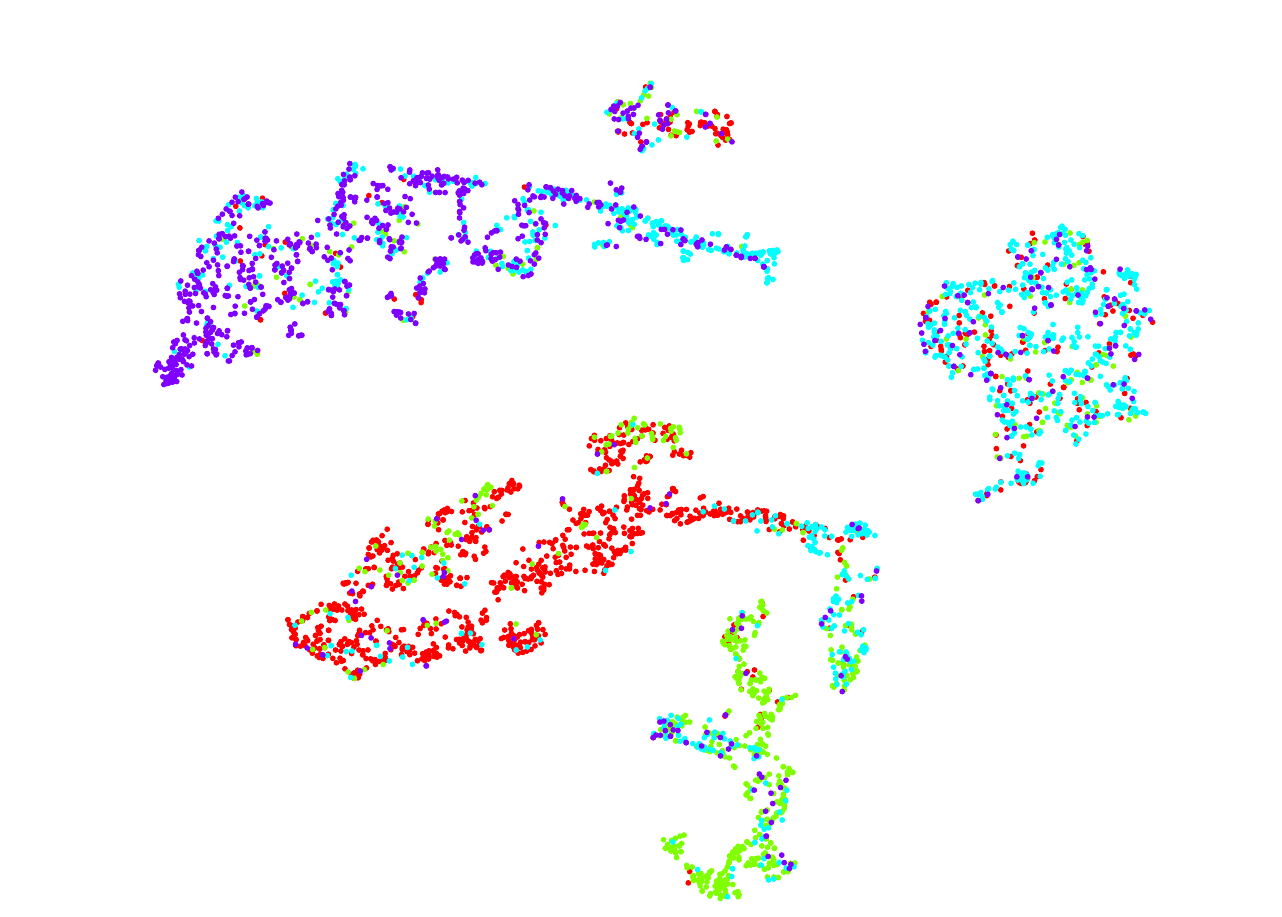}
	}
	\subfigure[AGCN]{
	\includegraphics [width=0.37\columnwidth]{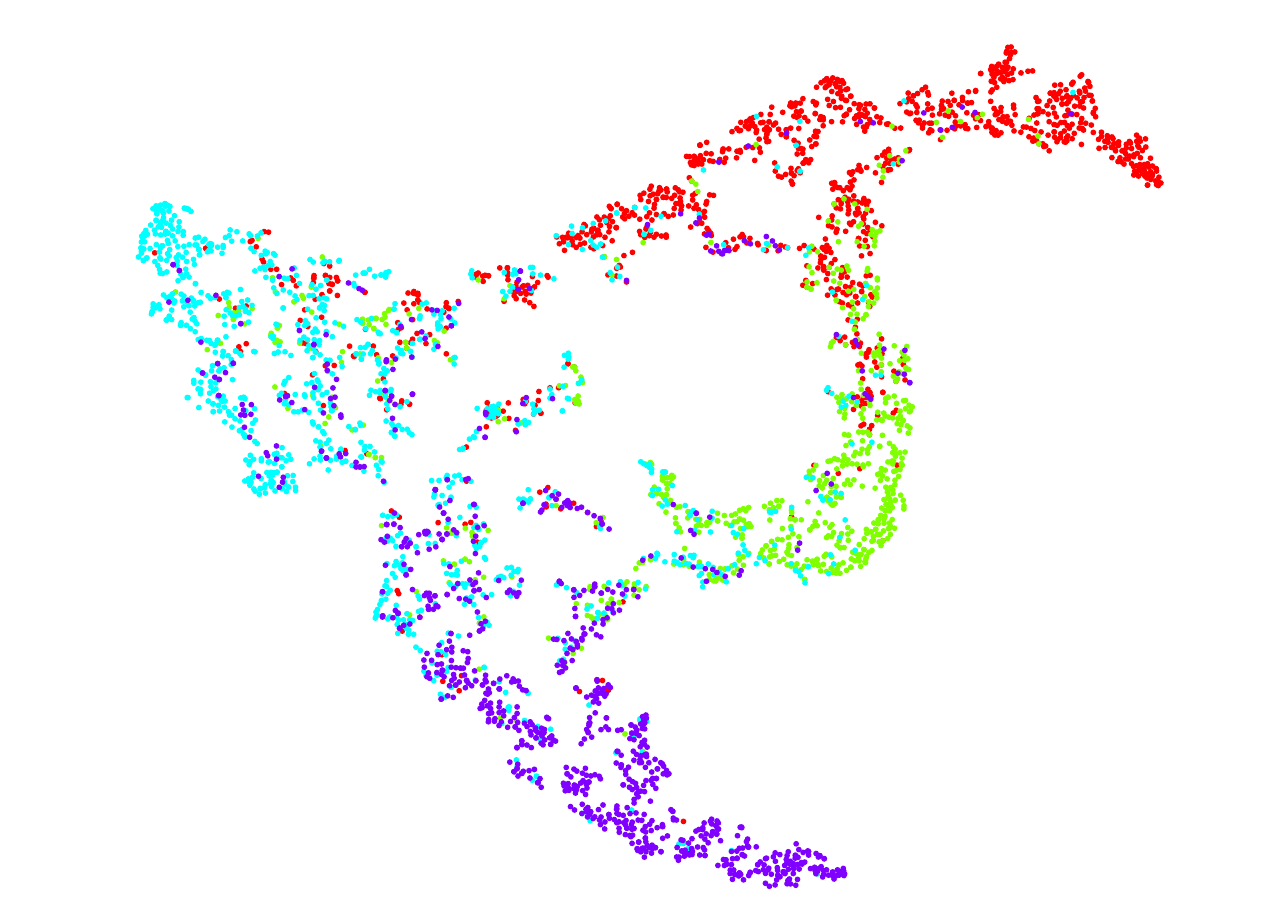}
	}
    \subfigure[EGAE]{
	\includegraphics [width=0.37\columnwidth, height=0.23\columnwidth]{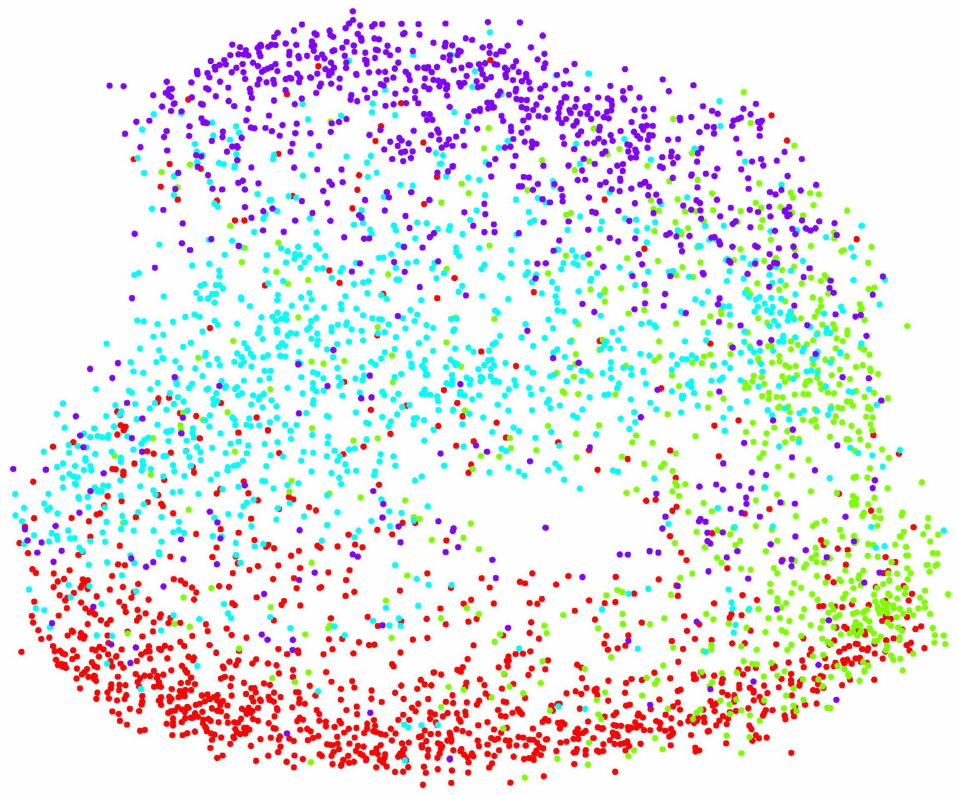}
	}
	\subfigure[AGCC]{
	\includegraphics [width=0.37\columnwidth]{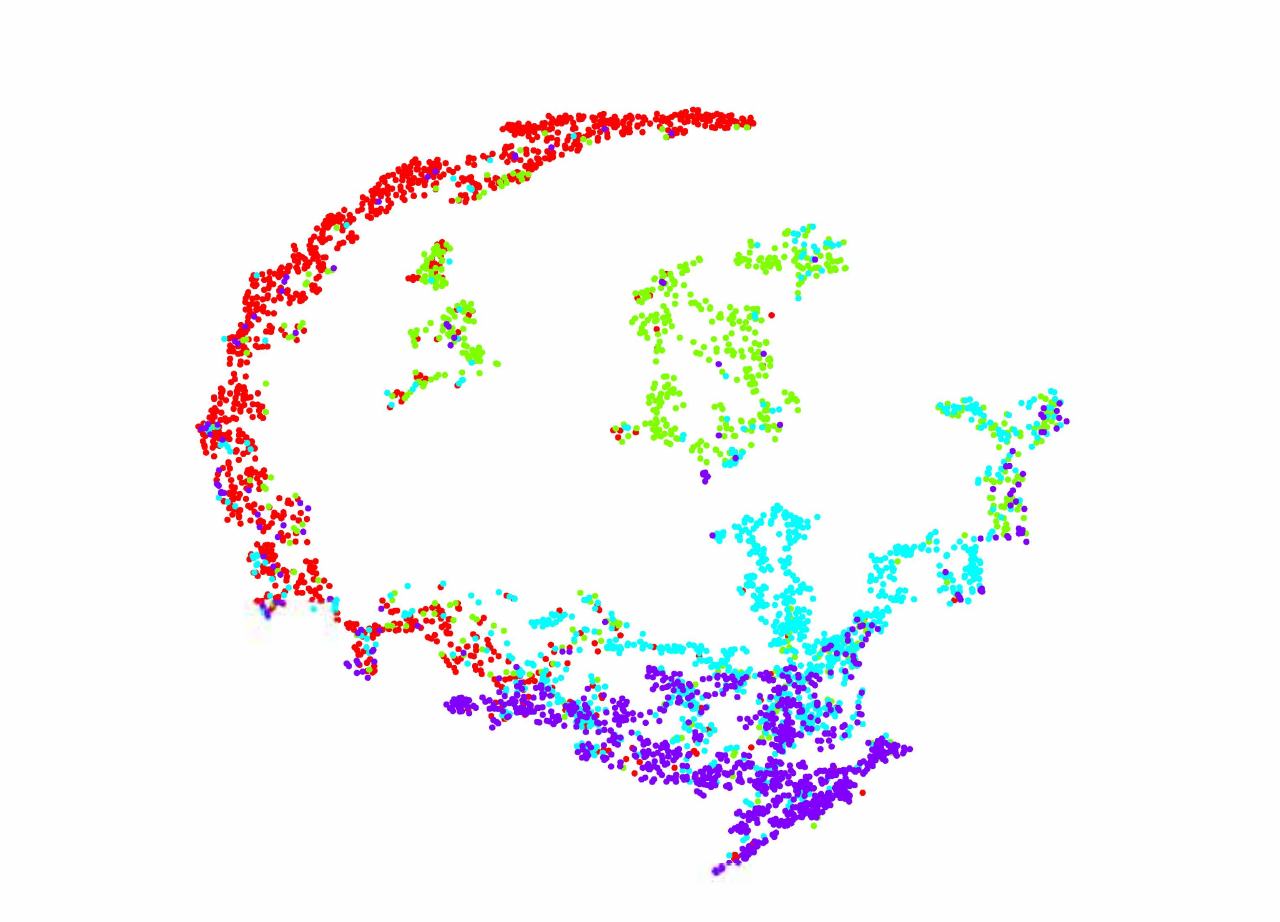}
	}
	\subfigure[Our]{
	\includegraphics [width=0.37\columnwidth]{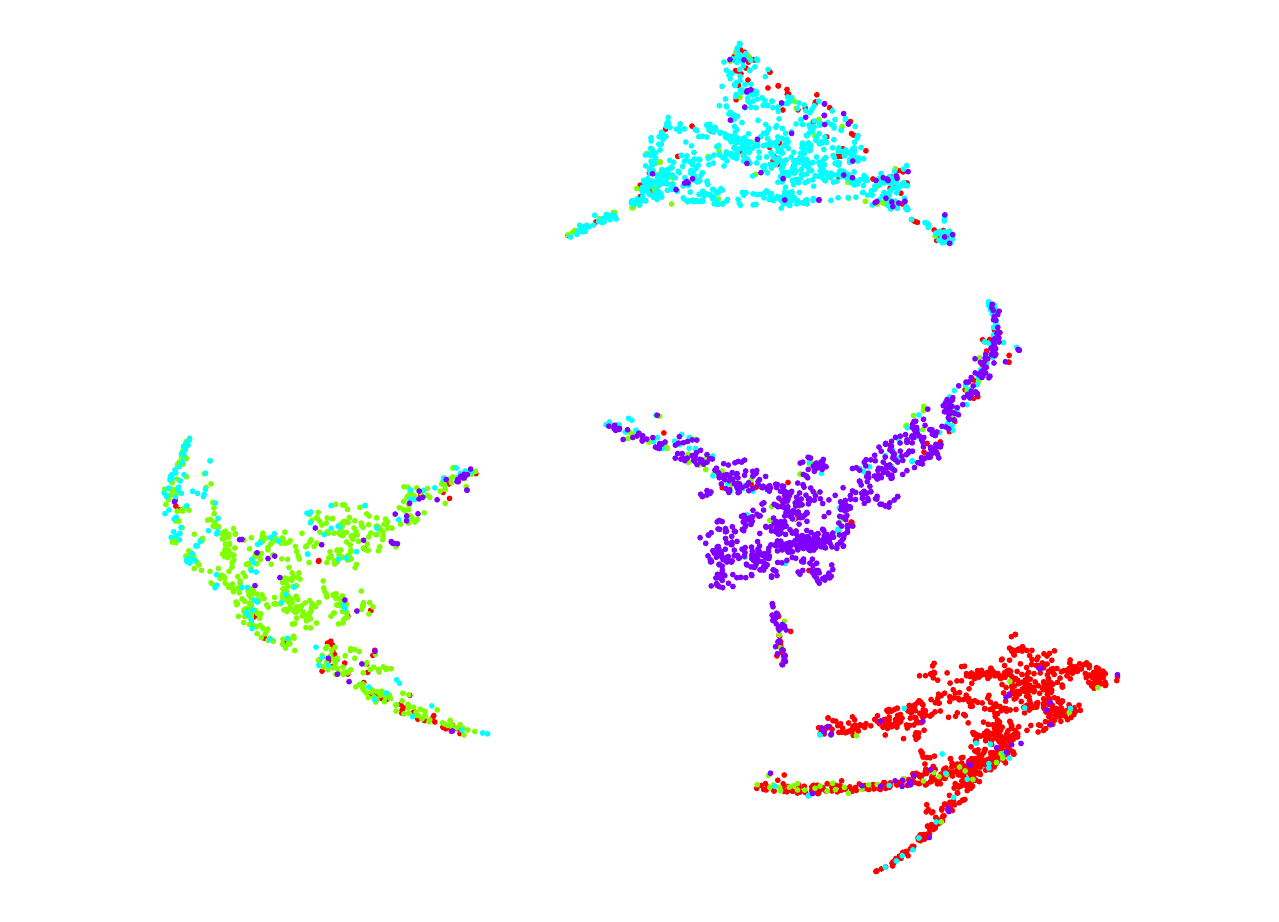}
	}
    \caption{Visualization of the learned representations by (a) SDCN \cite{bo2020structural}, (b) AGCN \cite{peng2021attention}, (c) EGAE \cite{zhang2022embedding}, (d) AGCC \cite{he2022parallelly}, and (e) our EGRC-Net on DBLP, where different colors represent different clusters.}
	\label{fig: tsne2}
\end{figure*}

\begin{figure}[ht]
	\centering
        \subfigure[DBLP]{
	\includegraphics [width=0.42\columnwidth, height=0.22\columnwidth]{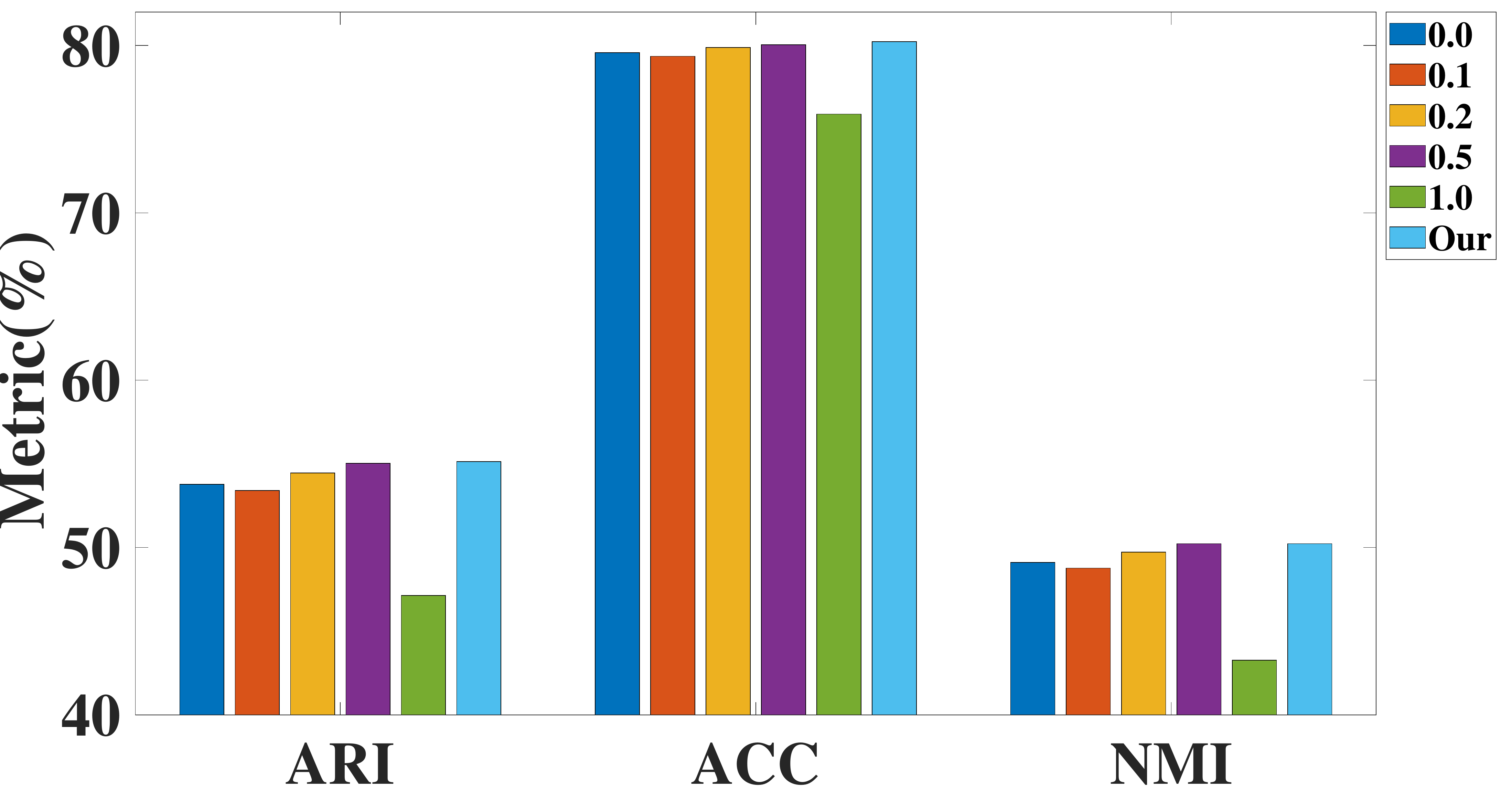}
         }
        \subfigure[ACM]{
        \includegraphics [width=0.42\columnwidth, height=0.220\columnwidth]{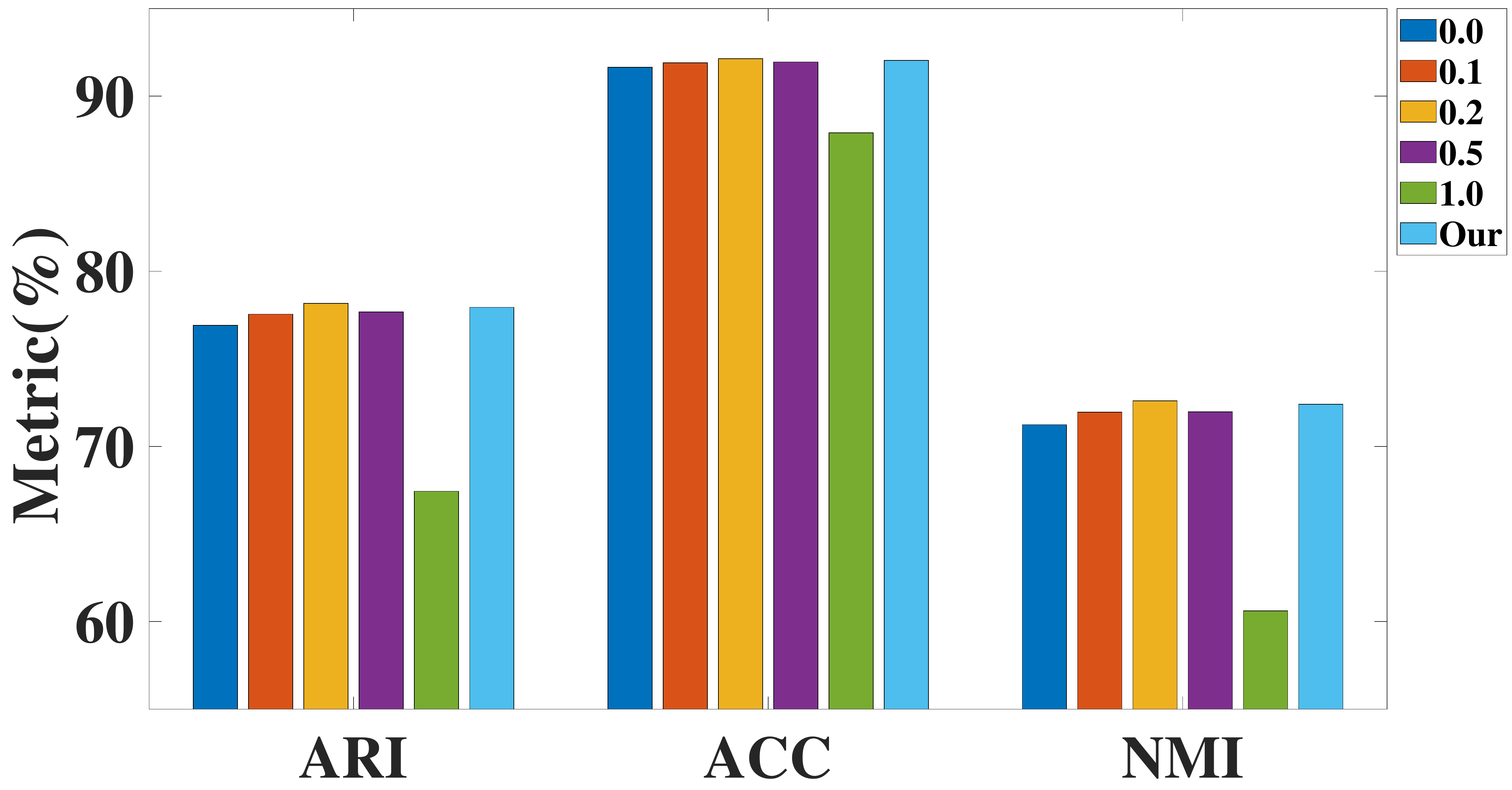}
        }
    \caption{The clustering results on (a) DBLP and (b) ACM corresponding to the fixed teleport probability values (i.e., $\rho={0.0, 0.1, 0.2, 0.5, 1.0}$ in Eq. (\ref{eq: APPNP})) and our learned one (i.e., $\Theta$ in Eq. (\ref{eq: z-APPNP})), where our learned one always obtains the best clustering performance among all metrics.}
	\label{fig: PA_rho}
\end{figure}

\begin{figure}[ht]
	\centering
        \subfigure[DBLP]{
	\includegraphics [width=0.46\columnwidth, height=0.20\columnwidth]{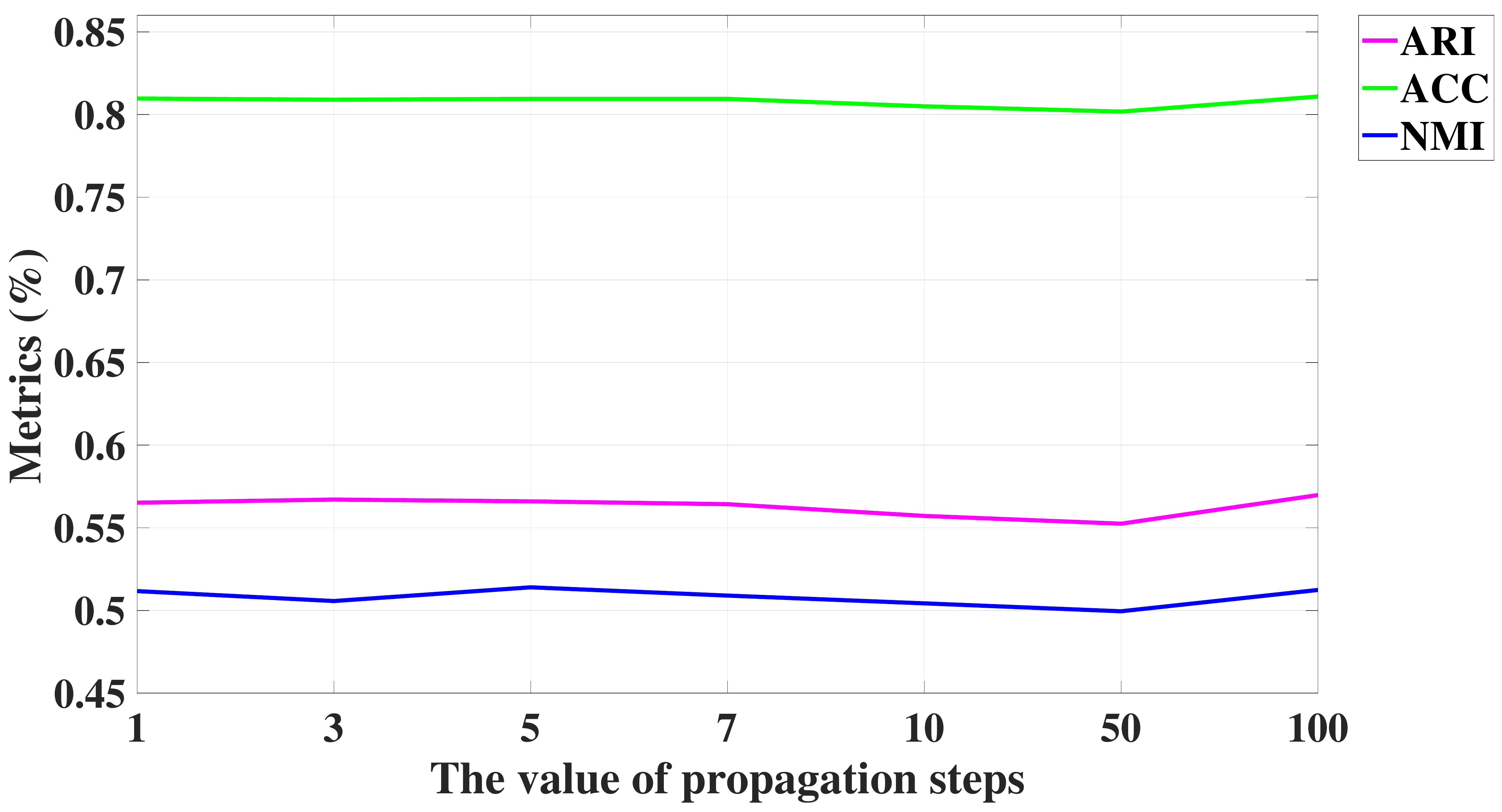}
         }
        \subfigure[ACM]{
        \includegraphics [width=0.46\columnwidth, height=0.20\columnwidth]{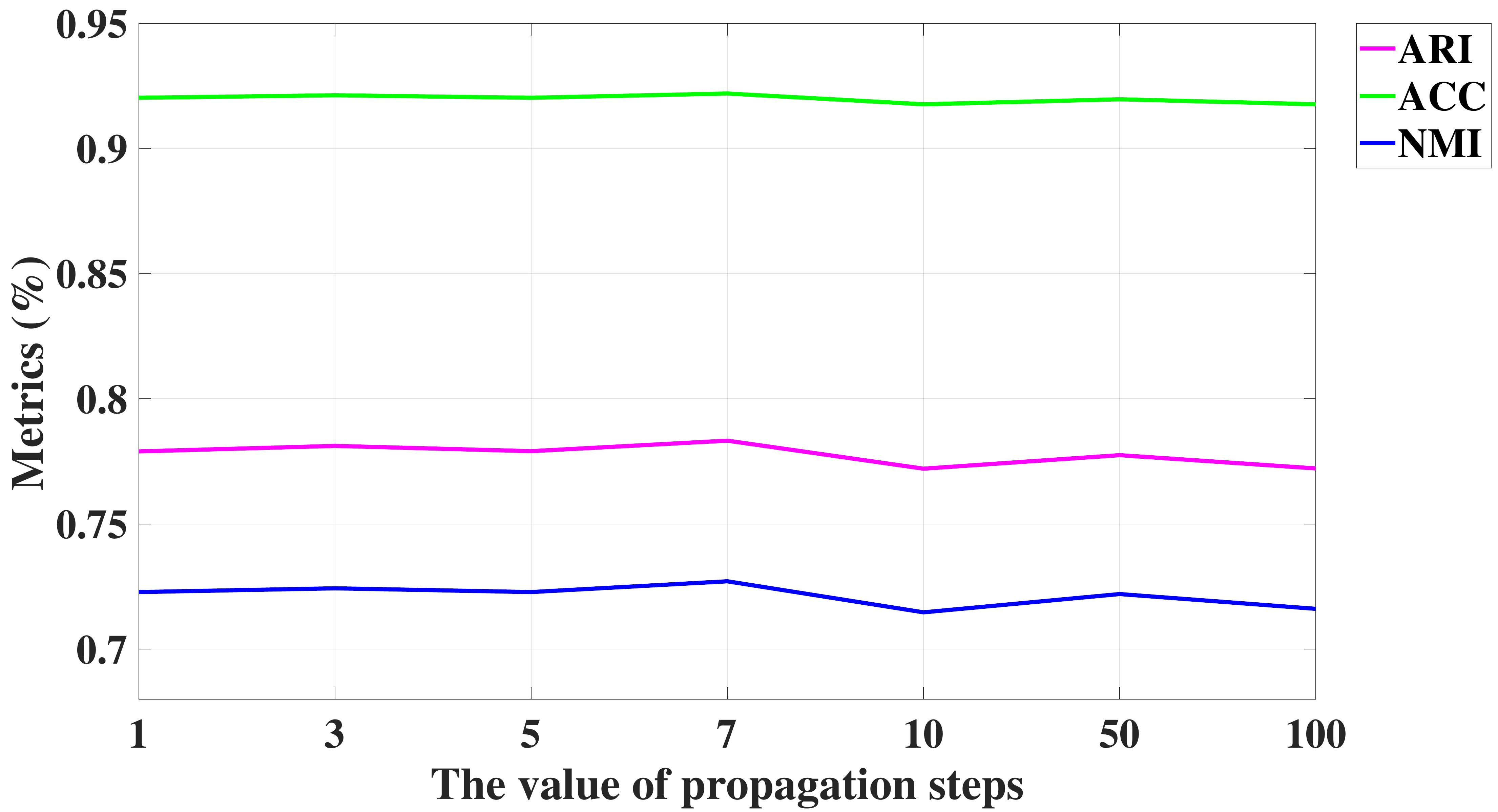}
        }
    \caption{The clustering results on (a) DBLP and (b) ACM with three metrics w.r.t. the number of propagation steps.}
	\label{fig: PA_tau}
\end{figure}

\vspace{-10 mm}

\subsection{Ablation Study} \label{sec: AS}
We conducted a series of ablation experiments to evaluate the effectiveness of the proposed graph refinement (GR) architecture and the Jeffreys divergence (JD) minimization term, where the experimental results are shown in Table \ref{tab: as_results}, \XSolidBrush and \Checkmark in each row indicate the non-use and use of the corresponding component, respectively. Specifically, non-use of the GR term means that we use a fixed graph to conduct graph clustering, and non-use of the JD term means that we minimize an asymmetric divergence between $\mathbf{Q}$, $\mathbf{Z}_{a}$, and $\mathbf{P}$ (i.e., $\lambda_1 \sum_\emph{i}^\emph{n}\sum_\emph{j}^\kappa{\emph{p}_{\emph{i},\emph{j}}log{\frac{\emph{p}_{\emph{i},\emph{j}}}{\emph{z}_{i,\emph{j}}}}}
+ \lambda_2 \sum_\emph{i}^\emph{n}\sum_\emph{j}^\kappa\emph{p}_{\emph{i},\emph{j}}log{\frac{\emph{p}_{\emph{i},\emph{j}}}{\emph{q}_{i,\emph{j}}}}
+ \lambda_3 \sum_\emph{i}^\emph{n}\sum_\emph{j}^\kappa
\emph{z}_{\emph{i},\emph{j}}log{\frac{\emph{z}_{\emph{i},\emph{j}}}{\emph{q}_{i,\emph{j}}}}$).
\textcolor{black}{As shown in Table \ref{tab: as_results}, we can find that the GR architecture or the JD minimization term improves the clustering performance on most benchmark datasets, verified by comparing the first and second-row results and the first and third-row results in all metrics. In addition, comparing the second (using the fixed graph based on JD) and fourth (using our refined graph based on JD) row results, we can find that the GR strategy can achieve the best clustering performance among all metrics based on the JD minimization, validating the effectiveness of the proposed method.}

\begin{table}[ht]
\centering
\caption{\textcolor{black}{The comparisons of state-of-the-art approaches and the proposed methods on training time and network parameters as well as the clustering performance, where the $\uparrow$ shows the clustering improvements over the best comparisons with three metrics. We highlighted the best results with \textbf{bold}.}}
\label{tab: TIME_results}
\resizebox{0.99\columnwidth}{!}{
\begin{tabular}{c|c|ccc|cc}
\hline\hline
Dataset               &                  & ARI(\%)                & ACC(\%)                & NMI(\%)                & Time(s)          & Parameter(M)      \\
\hline\hline
\multirow{6}{*}{ACM}  & SDCN             & 73.91$\pm$0.40          & 90.45$\pm$0.18          & 68.31$\pm$0.25          & \textbf{265.930} & 6.622940           \\
                      & AGCN             & 74.20$\pm$0.38          & 90.59$\pm$0.15          & 68.38$\pm$0.45          & 502.224          & 6.659081           \\
                      & AGCC             & 73.73$\pm$0.90          & 90.38$\pm$0.38          & 68.34$\pm$0.89          & 2204.842         & 14.934554          \\
\cline{2-7}
                      & Our              & 76.04$\pm$0.39          & 91.30$\pm$0.17          & 70.40$\pm$0.30          & 1307.355         & 6.671183           \\
                      & Our   (Scalable) & \textbf{77.94$\pm$0.49} & \textbf{92.04$\pm$0.19} & \textbf{72.41$\pm$0.47} & 1224.115         & \textbf{4.435626}  \\
\hline
                      & boost   $\uparrow$ & $\uparrow$  3.74        & $\uparrow$ 1.45         & $\uparrow$ 4.03         &                  &                   \\
\hline\hline
\multirow{6}{*}{CITE} & SDCN             & 40.17$\pm$0.43          & 65.96$\pm$0.31          & 38.71$\pm$0.32          & \textbf{296.461} & 9.374333           \\
                      & AGCN             & 43.79$\pm$0.31          & 68.79$\pm$0.23          & 41.54$\pm$0.30          & 512.913          & 9.419504           \\
                      & AGCC             & 41.82$\pm$2.03          & 68.08$\pm$1.44          & 40.86$\pm$1.45          & 3230.083         & 18.608447          \\
\cline{2-7}
                      & Our              & \textbf{48.32$\pm$0.57} & \textbf{72.27$\pm$0.37} & \textbf{45.77$\pm$0.48} & 1726.946         & 9.432814           \\
                      & Our   (Scalable) & 34.38$\pm$0.93          & 62.71$\pm$0.74          & 35.15$\pm$0.65          & 1524.338         & \textbf{6.288308}  \\
\hline
                      & boost   $\uparrow$ & $\uparrow$4.53          & $\uparrow$ 3.48         & $\uparrow$ 4.23         &                  &\\
\hline\hline
\multirow{6}{*}{DBLP} & SDCN             & 39.15$\pm$2.01          & 68.05$\pm$1.81          & 39.50$\pm$1.34          & \textbf{299.241} & 4.317424          \\
                      & AGCN             & 42.49$\pm$0.31          & 73.26$\pm$0.37          & 39.68$\pm$0.42          & 648.020          & 4.356575          \\
                      & AGCC             & 44.40$\pm$3.79          & 73.45$\pm$2.16          & 40.36$\pm$2.81          & 2539.032         & 11.863038         \\
\cline{2-7}
                      & Our              & \textbf{56.39$\pm$0.93} & \textbf{80.53$\pm$0.54} & \textbf{50.85$\pm$0.60} & 2430.525         & 4.372805          \\
                      & Our   (Scalable) & 55.13$\pm$1.67          & 80.23$\pm$0.86          & 50.23$\pm$1.12          & 1951.387         & \textbf{2.897955} \\
\hline
                      & boost   $\uparrow$ & $\uparrow$ 11.99        & $\uparrow$ 7.08         & $\uparrow$ 10.49        &                  &                   \\
\hline\hline
\end{tabular}
}
\end{table}

\subsection{Visual Comparison} \label{sec: visua}
\textcolor{black}{
To qualitatively validate the significant performance of the proposed method EGRC-Net, we plotted t-SNE visualizations \cite{maaten2008visualizing} of the learned embeddings by SDCN \cite{bo2020structural}, AGCN \cite{peng2021attention}, EGAE \cite{zhang2022embedding}, AGCC \cite{he2022parallelly}, and our EGRC-Net on DBLP. The visualized results are shown in Figure \ref{fig: tsne2}, where we can observe that the representation resulting from our method shows the best separability within different clusters. That is to say, the intra-cluster samples gather together, and gaps among inter-clusters are obvious, illustrating that our proposed method EGRC-Net can provide a better discriminative representation than state-of-the-art methods.}

\subsection{Analysis of scalable EGRC-Net} \label{sec: ana_scalable}  
\textcolor{black}{We investigated the effect of the teleport probability value corresponding to the typical fixed value (i.e., $\rho$ in Eq. (\ref{eq: APPNP})) and our learned one (i.e., $\Theta$ in Eq. (\ref{eq: z-APPNP})). Figure \ref{fig: PA_rho} gets the clustering results on DBLP and ACM with a series of threshold settings (i.e., $\rho={0.0, 0.1, 0.2, 0.5, 1.0}$ and our learned one), where different fixed heuristic threshold values show different clustering performances while our learned teleport probability value always obtains the best clustering performance among all metrics. In addition, we can find that the case without the topology structure information (i.e., $\rho=1.0$) makes a great performance degradation, illustrating the importance of considering graph information in the clustering task.}

\textcolor{black}{
Additionally, we investigated different numbers of propagation steps $\tau$ on DBLP and ACM in Figure \ref{fig: PA_tau}, where the scalable EGRC-Net performs stable clustering results in three metrics, which has the same effect as APPNP \cite{gasteigerpredict} of utilizing far more propagation steps without leading to over smoothing. As also shown in that figure, it is enough to use one power iteration to effectively complete the transmission and aggregation of node information on the graph.}

\subsection{Running Time and Parameters Analysis} \label{sec: time-para}
\textcolor{black}{We trained SDCN \cite{bo2020structural}, AGCN \cite{peng2021attention}, AGCC \cite{he2022parallelly}, our EGRC-Net, and the scalable variant for 200 epochs on ACM, CITE, and DBLP. The experiments were repeated ten times. The compared results regarding the training time are shown in Table \ref{tab: TIME_results}, where the comparisons show that our EGRC-Net achieves a noteworthy improvement in clustering performance while maintaining acceptable resource consumption. Furthermore, in the case of DBLP, the scalable variant demonstrates a 10.73\% increase in ARI while reducing running time by 19.71\% and memory usage by 33.73\%.  Particularly, the performance of the scalable one in CITE is not significant. The reason is possible that in the graph of CITE, many nodes are not well-connected, making a limited graph embedding learning capability with the IAPPNP module. Notably, how to ensure the scalability of a graph neural network while ensuring its effectiveness is still an open and challenging research topic.}

\section{Conclusion}\label{sec: con}
We presented a novel graph refinement clustering network to address the problem that existing GCN-based graph clustering networks heavily rely on a predefined graph and may fail if the initial graph cannot truly and precisely reflect their topology structures on the embedding space. Specifically, we leveraged the vanilla DAE and GCN modules, a series of MLPs, and a graph fusion module to conduct embedding-induced graph refinement. 
Afterward, we minimized the Jeffreys divergence between multiple derived distributions to jointly optimize the embedding learning, the graph refinement, and the clustering assignments to achieve the clustering performance.
In addition, we also designed a simple yet effective graph embedding learning module (i.e., improved APPNP) to replace the vanilla GCN, resulting in a scalable variant of EGRC-Net. Extensive experiments and analyses on nine benchmark datasets with fifteen compared methods demonstrated that our EGRC-Net could consistently outperform state-of-the-art approaches. A series of experiments and analyses were conducted to understand the effectiveness of the proposed methods. In future work, we will focus on the interoperability of this method, contributing to the graph neural network community.

\balance
\bibliographystyle{IEEEtran}
\bibliography{gacn}

\begin{IEEEbiography}[{\includegraphics[width=1in,height=1.75in,clip,keepaspectratio]{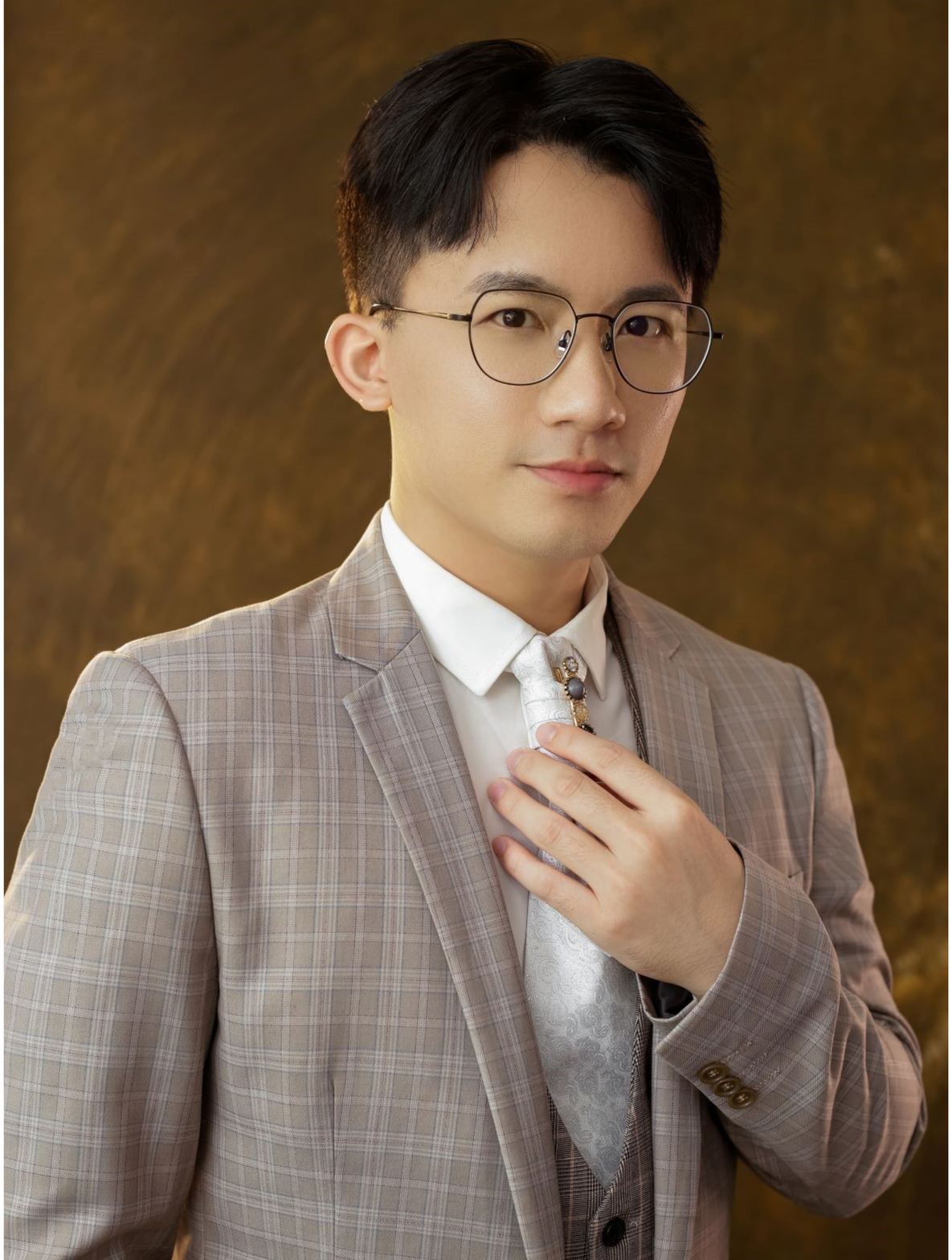}}]{Zhihao Peng}
received the B.S. and M.S. degrees in computer science and technology from Guangdong University of Technology, Guangzhou, China, in 2016 and 2019, respectively, and the Ph.D. degree in Computer Science from the City University of Hong Kong, SAR, China, in 2023. 

He is currently a postdoctoral fellow in Electrical Engineering at The Chinese University of Hong Kong. His current research interests include spectral clustering, subspace learning, and domain adaptation in image processing with unsupervised learning.
\end{IEEEbiography}

\vspace{-10 mm}
\begin{IEEEbiography}[
 {
  \includegraphics[width=1in,height=1.75in,clip,keepaspectratio]{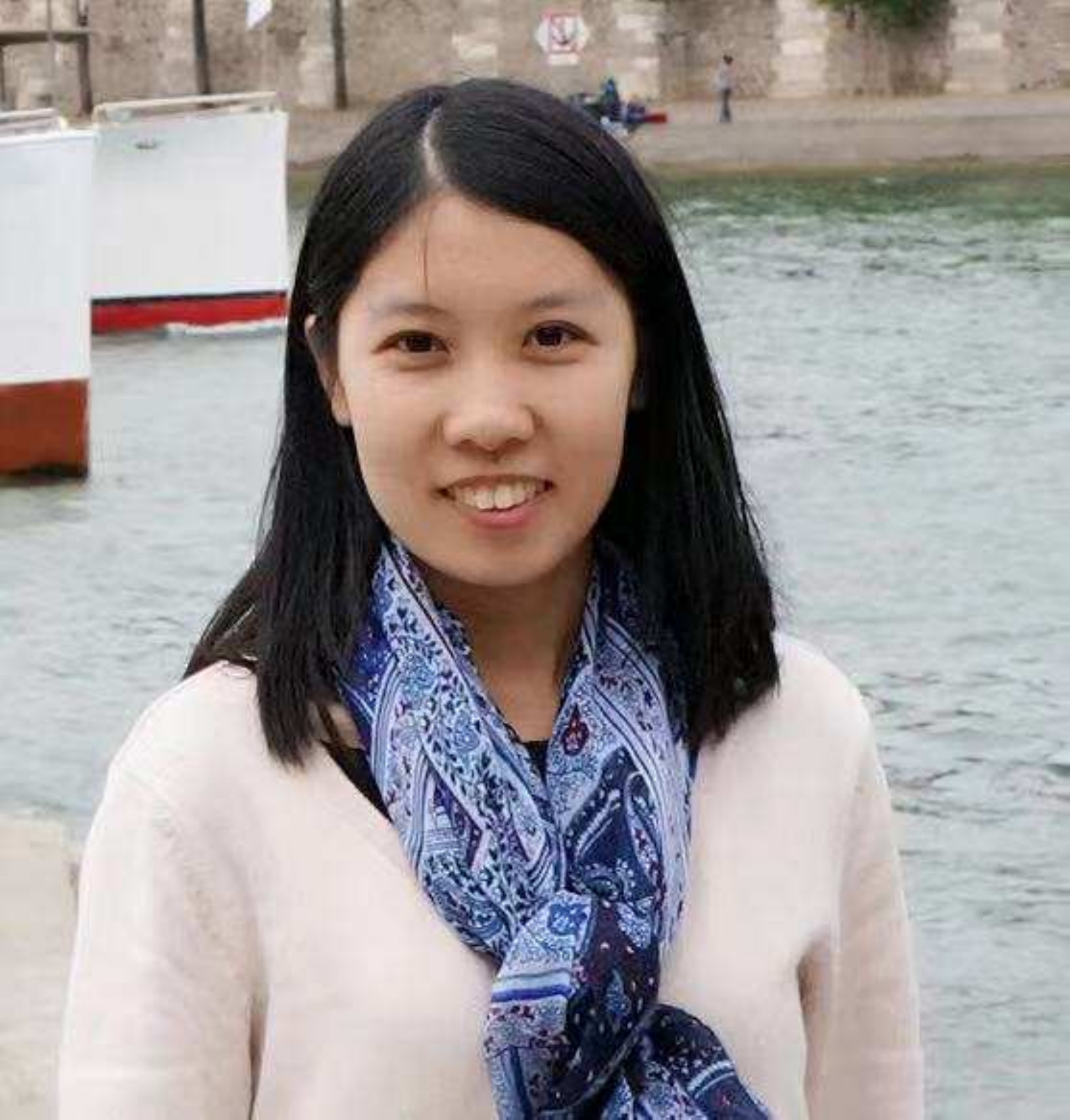}
 }]{Hui Liu}
received the B.Sc. degree in Communication Engineering from Central South University, Changsha, China, the M.Eng. degree in Computer Science from Nanyang Technological University, Singapore, and the Ph.D. degree in Computer Science from the City University of Hong Kong, Hong Kong. From 2014 to 2017, she was a Research Associate at the Maritime Institute, Nanyang Technological University. She is currently an Assistant Professor with the School of Computing Information Sciences, Caritas Institute of Higher Education, Hong Kong. Her research interests include image processing and machine learning. 
\end{IEEEbiography}
\vspace{-10 mm}

\begin{IEEEbiography}[{\includegraphics[width=1in,height=1.75in,clip,keepaspectratio]{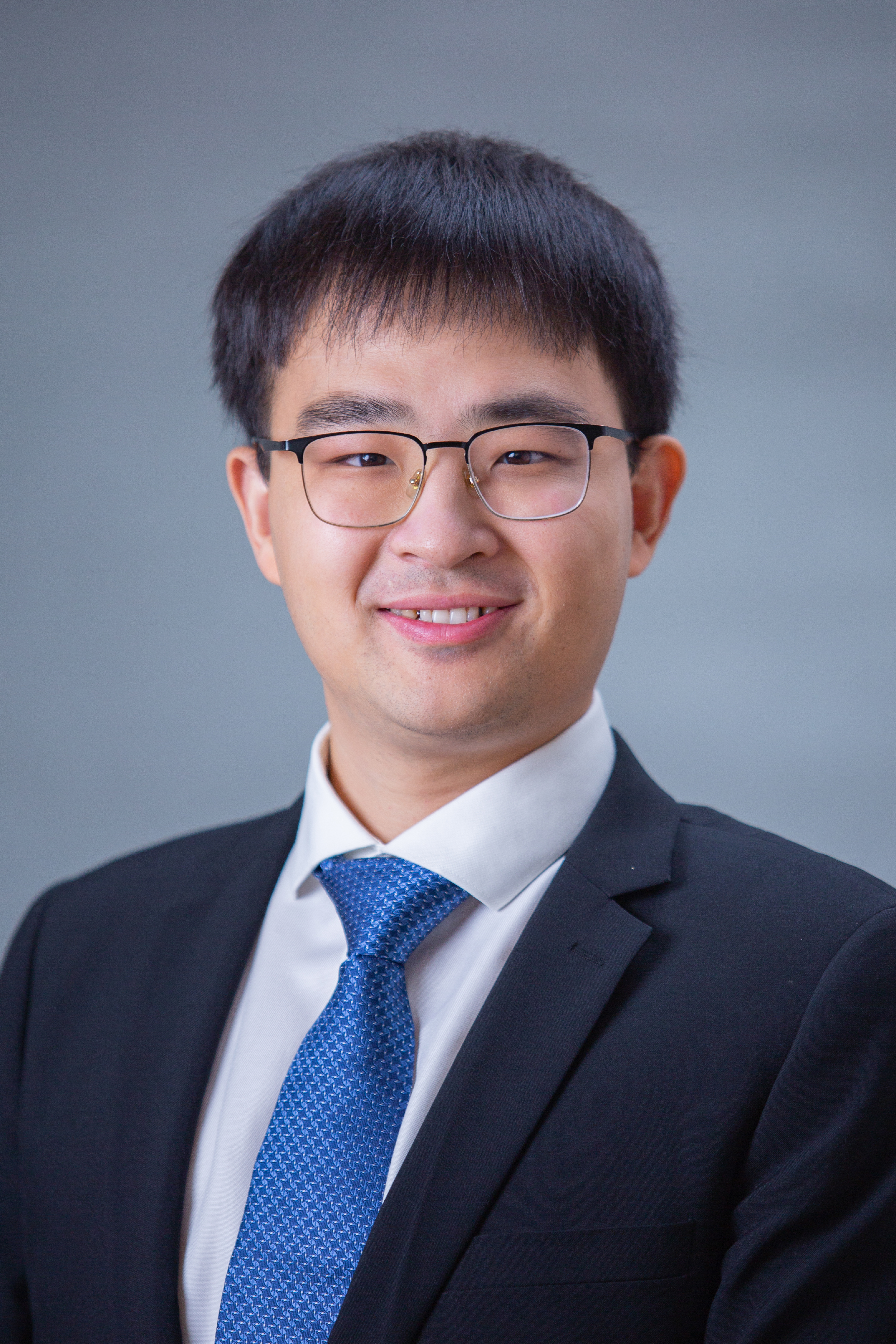}}]{Yuheng Jia}
(Member, IEEE) received the B.S. degree in automation and the M.S. degree in control theory and engineering from Zhengzhou University, Zhengzhou, China, in 2012 and 2015, respectively, and the Ph.D. degree in Computer Science from the City University of Hong Kong, Hong Kong, China, in 2019. 

He is currently an Associate Professor with the School of Computer Science and Engineering, Southeast University, Nanjing, China. His research interests broadly include topics in machine learning and data representation, such as weakly-supervised learning, high-dimensional data modeling and analysis, and low-rank tensor/matrix approximation and factorization.
\end{IEEEbiography}

\vspace{-10 mm}

\begin{IEEEbiography}[{\includegraphics[width=1in,height=1.75in,clip,keepaspectratio]{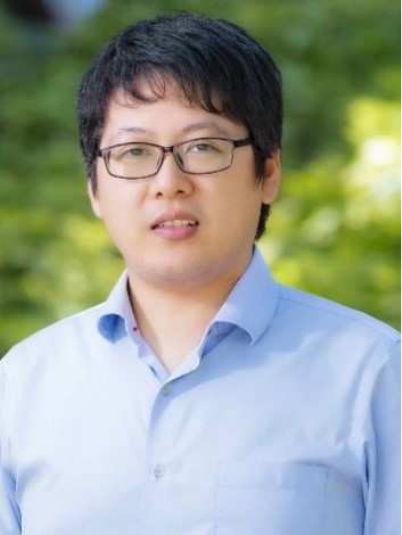}}]{Junhui Hou}
(Senior Member, IEEE) received the B.Eng. degree in information engineering (talented students program) from the South China University of Technology, Guangzhou, China, in 2009, the M.Eng. degree in signal and information processing from Northwestern Polytechnical University,  Xi’an, China, in 2012, and the Ph.D. degree from the School of Electrical and Electronic Engineering, Nanyang Technological University, Singapore,  in 2016.  

He is currently an Associate Professor with the Department of Computer Science, City University of Hong Kong. His current research interests include multi-dimensional visual computing. He is an elected member of IEEE MSA-TC, VSPC-TC, and MMSP-TC. He received the Early Career Award (3/381) from the Hong Kong Research Grants Council in 2018. He is currently serving as an Associate Editor for IEEE TRANSACTIONS ON VISUALIZATION AND COMPUTER GRAPHICS, IEEE  TRANSACTIONS ON CIRCUITS AND SYSTEMS FOR VIDEO TECHNOLOGY, IEEE TRANSACTIONS ON IMAGE PROCESSING, Signal Processing: Image Communication, and The Visual Computer.
\end{IEEEbiography}

\end{document}